\documentclass{article}

\PassOptionsToPackage{numbers, compress}{natbib}
  \usepackage[preprint]{neurips_2026}

\usepackage[utf8]{inputenc} 
\usepackage[T1]{fontenc}    
\usepackage{hyperref}       
\usepackage{url}            
\usepackage{booktabs}       
\usepackage{amsfonts}       
\usepackage{nicefrac}       
\usepackage{microtype}      
\usepackage{xcolor}         
\usepackage{enumitem}
\usepackage{amsmath}
\usepackage{graphicx}
\usepackage{subcaption}
\usepackage{cleveref}

\title{Topological Signatures of Grokking}

\author{%
  Yifan Tang \\
  Imperial College London \\
  \texttt{yifan.tang23@imperial.ac.uk} \\
   \And
   Qiquan Wang \\
   Queen Mary University of London \\
   \texttt{qiquan.wang@qmul.ac.uk} \\
   \And
   In\'{e}s Garc\'{i}a-Redondo \\
   University of Fribourg \\
   \texttt{ines.garciaredondo@unifr.ch} \\
   \And
   Anthea Monod \\
   Imperial College London \\
   \texttt{a.monod@imperial.ac.uk} \\
}

\begin{document}

\maketitle

\begin{abstract}
We study the grokking phenomenon through the lens of topology. Using persistent homology on point clouds derived from the embedding matrices of a range of models trained on modular arithmetic with varying primes, we identify a clear and consistent topological signature of grokking: a sharp increase in both the maximum and total persistence of first homology ($H_1$). Persistence diagrams reveal the emergence of a dominant long-lived topological feature together with increasingly structured secondary features, reflecting the underlying cyclic structure of the task. Compared to existing spectral and geometric diagnostics---specifically, Fourier analysis and local intrinsic dimension---persistent homology provides a unified geometric and topological characterization of representation learning, capturing both local and global multi-scale structure. Ablations across data regimes and control settings show that these topological transitions are tied to generalization rather than memorization. Our results suggest that persistent homology offers a principled and interpretable framework for analyzing how neural networks internalize latent structure during training.

\end{abstract}

\section{Introduction}


Deep neural networks can exhibit a striking training dynamic known as \emph{grokking}, where models initially memorize training data before abruptly transitioning to near-perfect generalization after prolonged optimization \citep{power2022grokking}. Despite growing interest in grokking and a range of proposed diagnostics, including Fourier analysis and geometric statistics \citep[e.g.,][]{humayun2024deep,mohamadi2024you}, there remains limited understanding of the global structure that emerges in learned representations during generalization.


In this work, we study grokking through the lens of topology. We apply \emph{persistent homology} (PH) to point clouds derived from neural representations during training, enabling a multi-scale analysis of their geometric and topological organization.


In the classical modular arithmetic setting, we identify a clear and reproducible topological signature of grokking. Across a range of models and primes, the maximum and total persistence of first homology ($H_1$) increase sharply during generalization, while persistence diagrams reveal the emergence of a dominant long-lived 1-cycle together with increasingly structured secondary features. These patterns are consistent with the cyclic structure underlying modular arithmetic and suggest that grokking coincides with the emergence of coherent topological organization in representation space.


We further investigate the robustness and limitations of this phenomenon. Similar topological transitions arise across different model classes, including transformers and multilayer perceptrons (MLPs), indicating that the effect is not architecture-specific. In contrast, experiments on MNIST \citep{lecun-mnisthandwrittendigit-2010} produce more diffuse and less interpretable topological behavior, suggesting that the observed signatures are tied to the presence of underlying global structure rather than arising universally during training.


We also compare PH with existing specral and geometric grokking diagnostics. While these methods capture aspects of the grokking transition, PH provides a unified geometric and topological perspective that captures both local and global multi-scale structure. Finally, through ablations varying training fraction and label randomization, we show that the observed PH signatures track generalization and disappear when grokking fails to occur.


\textbf{Our contributions are as follows.}
\begin{itemize}[noitemsep]
\item We identify a robust topological signature of grokking using persistent homology.
\item We provide a geometric and topological interpretation of grokking in terms of emergent structure in representation space.
\item We demonstrate consistent behavior across model classes and analyze both synthetic and real-data settings.
\item We show through ablations and controls that the observed PH signatures are associated with generalization rather than memorization.
\end{itemize}

Overall, our results suggest that persistent homology provides a principled and interpretable framework for studying representation learning, offering new insight into how neural networks internalize latent structure during training.

\section{Background}

In this section we introduce the relevant background on PH to facilitate the interpretation of our results, and position our work with respect to the existing literature.

\subsection{Persistent Homology (PH)}


PH is a tool from topological data analysis that quantifies the multi-scale \emph{shape} of data. Given a point cloud equipped with a distance metric, PH constructs a nested sequence of simplicial complexes, called a \emph{filtration}, by progressively increasing a distance threshold $\epsilon \geq 0$. In this work we use the Vietoris--Rips filtration: at scale $\epsilon$, points within distance $\epsilon$ are connected by edges, and higher-dimensional simplices are included whenever all pairwise distances among their vertices are below $\epsilon$. As $\epsilon$ increases, topological features such as connected components and loops appear and disappear.


These features are summarized in a \emph{persistence diagram}, a multiset of points $(b,d)$ recording the birth and death scales of each feature. Degree-0 points correspond to connected components, while degree-1 points correspond to loops or cycles. The persistence $d-b$ measures how prominent a feature is across scales: points close to the diagonal are typically interpreted as noise, while long-lived features indicate meaningful geometric structure.


In our setting, we apply PH to point clouds extracted from neural representations throughout training. The emergence of persistent degree-1 ($H_1$) features indicates the formation of loop-like organization in representation space, which we interpret as a geometric signature of grokking.

\subsection{Related Work}

Grokking was first observed by \cite{power2022grokking} in transformers trained on algorithmic tasks, and was later extended to non-algorithmic settings and broader architectures \cite{liu2023omnigrok, humayun2024deep}. Existing explanations largely frame grokking as a phase transition measurable through complexity or information-theoretic quantities \cite{liu2023omnigrok, merrill2023a, clauw2024information, carvalho2025grokking, demoss2025complexity}. Our work instead contributes to the line of research relating grokking to the emergence of structure in learned representations \cite{zheng2024delays}.


Within this perspective, \cite{nanda2023progress} used Fourier analysis to show that transformers trained on modular arithmetic learn representations consistent with circular structure, linking grokking to the emergence of latent geometric organization. Relatedly, \cite{yildirim2026geometric} reduced grokking delay by constraining representations to lie on a sphere. Our approach is observational rather than interventional: using persistent homology, we detect the emergence of persistent 1-dimensional topological structure in representations learned during grokking. We also compare our approach with local intrinsic dimension (LID), another geometric diagnostic proposed for studying grokking dynamics \cite{brown2022relating, ruppik2026less}.



Persistent homology has previously been used to analyze neural network representations and transformer embeddings \cite{watanabe2022topological, ballester2023topological, uchendu2024unveiling, lee2025geometric, kushnareva2021artificial, tan_fractal}. To our knowledge, however, this is the first work applying persistent homology to characterize topological signatures of grokking and relate them to the latent cyclic structure of the training task.


\section{Experimental Setup}
\label{sec:experimental_setup}

\textbf{Task and data.}
We study the modular addition task: given two integers $a, b \in \{0, \ldots, p-1\}$, predict $(a + b) \bmod p$. The full dataset consists of all $p^2$ input pairs, represented as two-token sequences $[a, b]$. We use $p \in \{113, 149, 197\}$ as the modulus, following prior work on grokking~\citep{power2022grokking}. Models are trained on a random subset of fraction $\alpha \in \{0.20, 0.25, 0.30\}$ of all pairs, with the remainder held out as a test set. To test that the observed signals are not artifacts, as a control setting we additionally run experiments where training labels are independently permuted at random while the test set retains the original labels.

\textbf{Model architecture.}
We study two architectures on the same task.
\begin{enumerate}
    \item \textbf{Transformer.} We use a two-layer transformer encoder with pre-layer-norm. Token embeddings and learned positional embeddings of dimension $d_\text{model} = 128$ are summed and passed through two standard encoder blocks, each with $h = 4$ attention heads, key/query dimension $d_\text{attn} = 32$ per head, and a two-layer MLP with hidden dimension $d_\text{ff} = 256$ and GELU activations. After the encoder processes both tokens via self-attention, only the hidden state at the second token position is extracted, passed through a layer norm, and projected linearly to produce logits over the $p$ output classes. No dropout is applied.
    
    \item \textbf{MLP.} As a non-attention baseline, we use a feed-forward network with a shared token embedding of dimension $d_\text{embed} = 128$. The embeddings of $a$ and $b$ are concatenated into a $2d_\text{embed}$-dimensional vector and passed through three hidden layers of width $512$ with GELU activations, followed by a linear readout to $p$ classes. No positional encoding or self-attention is used.
\end{enumerate}

\textbf{Training.} Models are optimized with AdamW (\(\beta_1 = 0.9\), \(\beta_2 = 0.98\), \(\epsilon = 10^{-6}\), weight decay \(\lambda = 0.1\)) using a learning rate of \(3 \times 10^{-3}\). Training uses a brief linear warm-up over the first 10 steps, after which the learning rate is held constant. Models are trained for \(6 \times 10^4\) gradient steps with a fixed batch size of 512. Model weights, optimizer states, and train/test metrics are checkpointed every 500 steps. All experiments are run with fixed random seeds for reproducibility.

Experiments were conducted on a local workstation equipped with a single NVIDIA GeForce RTX 3070 Laptop GPU (8GB memory) and an AMD Ryzen 7 5800H CPU. The software environment used Windows Subsystem for Linux 2 (WSL2), configured with 10 CPU threads and 10GB system memory. Each experiment used a single GPU.

For each prime and training fraction, models were trained over five random seeds (46--50). Training a single model required approximately 8--10 minutes for both transformer and MLP architectures. The CPU-based Vietoris--Rips PH and LID analyses required approximately 2 and 6 minutes per model, respectively.

\textbf{Analysis.}
We characterize the geometry of learned representations through three complementary lenses.

\begin{enumerate}
    \item \textbf{Topological data analysis.}
    At selected checkpoints we construct point clouds from (i) the rows of the token embedding matrix $\texttt{tok\_emb.weight} \in \mathbb{R}^{p \times d_\text{model}}$, i.e.\ the $p$ learned token vectors prior to the addition of positional embeddings, and (ii) the second-token hidden states after each encoder layer on the test set. Prior to computing PH, each point cloud is centered and normalized. We then compute Vietoris--Rips PH in dimensions 0 and 1 using \textsc{Ripser}~\citep{bauer2021ripser}.
    
    \item \textbf{Fourier analysis.}
    Following \citet{nanda2023progress}, we compute the 2-D discrete Fourier transform of the $p \times p$ logit tensor and measure the restricted and excluded accuracies. We also compute per-head 1-D Fourier spectra of the embedding, key, query, and value weight rows to identify the dominant frequencies learned at each layer.
    \item \textbf{Local intrinsic dimension.} 
    We estimate the local intrinsic dimension (LID) of the same model checkpoints used for the TDA analysis, using the TwoNN estimator~\citep{facco2017estimating} as implemented in \texttt{skdim}. The point cloud is constructed from the layer-2 hidden states of both token positions on the test set, subsampled to 2,000 points, and we track how the mean LID and its distribution evolve across training checkpoints to characterize the progressive compression or expansion of learned geometry.
\end{enumerate}

\textbf{Comparison with previous local intrinsic dimension studies~\citep{ruppik2026less}.}
\citet{ruppik2026less} study LID dynamics during grokking on modular addition using a closely related setup, which we broadly follow but differ from in several respects. Their experiments use a single modulus $p = 197$ and train fractions ranging from $10\%$ to $50\%$, whereas we vary $p \in \{113,149,197\}$ and focus on $\alpha \in \{0.20,0.25,0.30\}$. Both works train for $60{,}000$ optimization steps, but the architectures and input representations differ: their setup uses four-token sequences of the form ``$a+b=$'', while ours uses only the two input tokens $(a,b)$. 

For LID estimation, they subsample $3{,}000$ points with a fixed neighborhood size of $L=64$, whereas we subsample $2{,}000$ points with the same neighborhood size. Additionally, their LID point clouds are constructed from all token positions in the sequence, whereas we restrict attention to the two input-token positions at layer 2.



\section{Results and Discussion}

We investigate the topological signatures of grokking across architectures, datasets, and training regimes, beginning with modular arithmetic and then considering ablations and MNIST as a contrasting real-data setting.

\subsection{Transformer Grokking on Modular Arithmetic}
\label{sec:results_transformer}

\begin{figure}[t]
    \centering
    \includegraphics[width=0.45\textwidth]{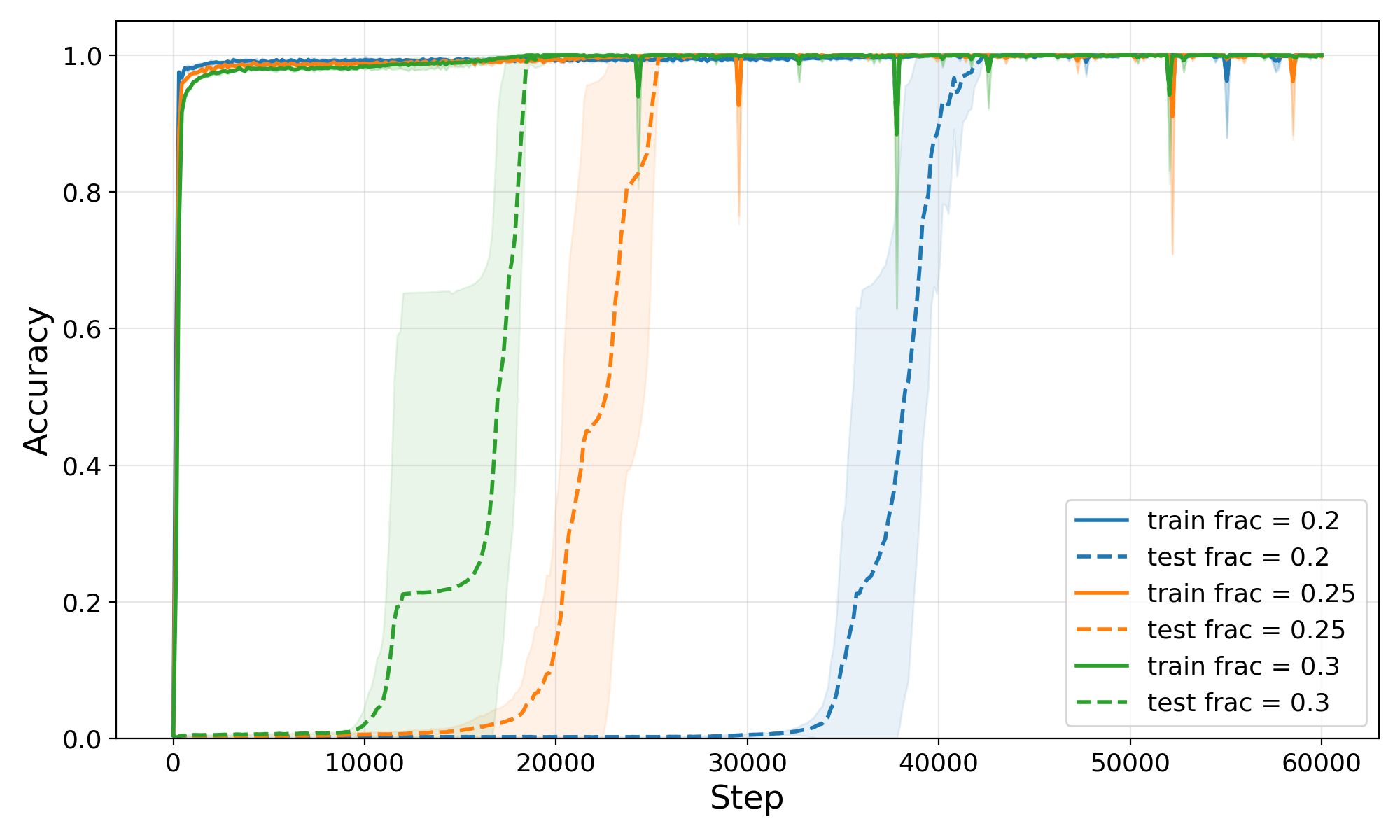}%
    \includegraphics[width=0.45\textwidth]{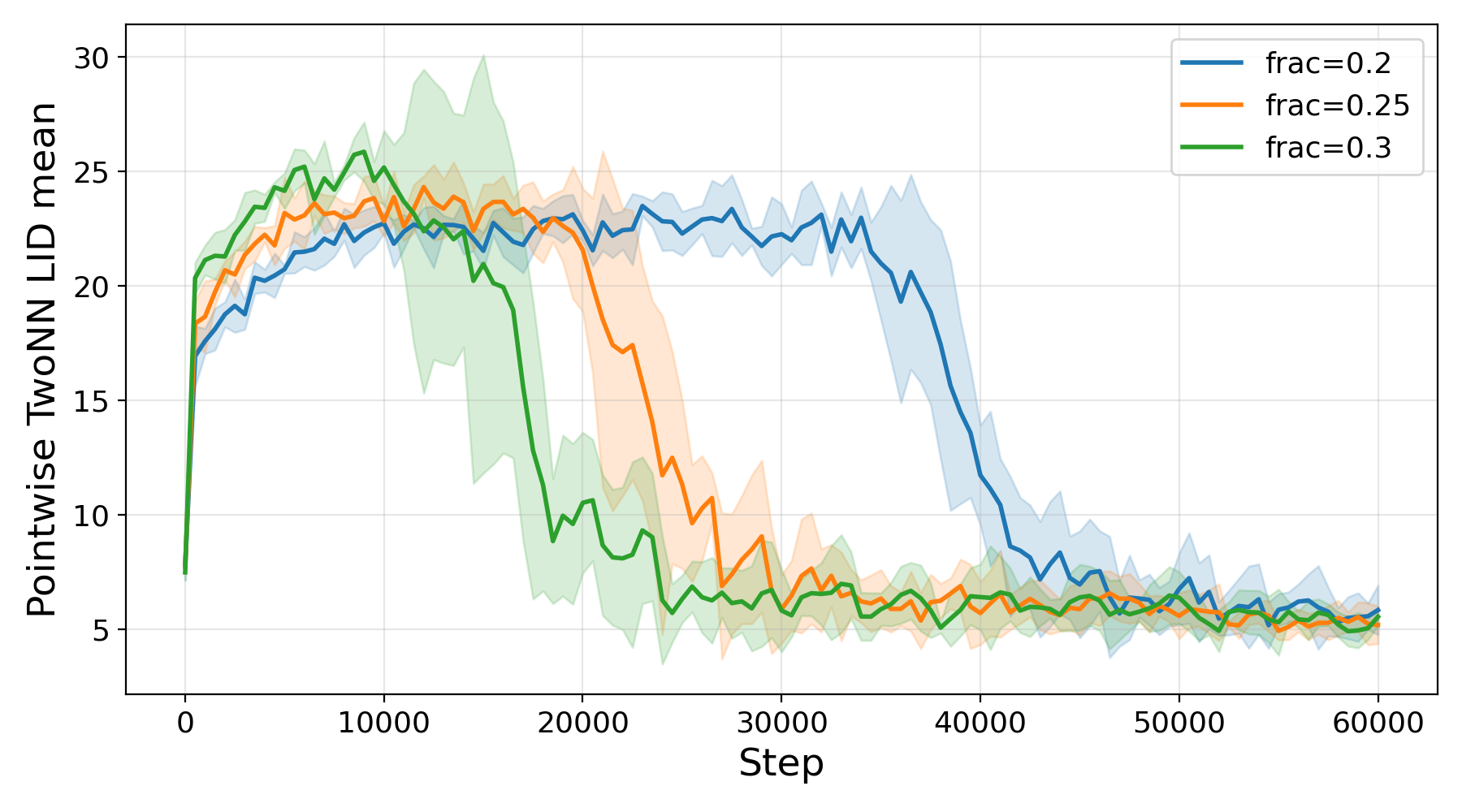}\\[4pt]
    \includegraphics[width=0.45\textwidth]{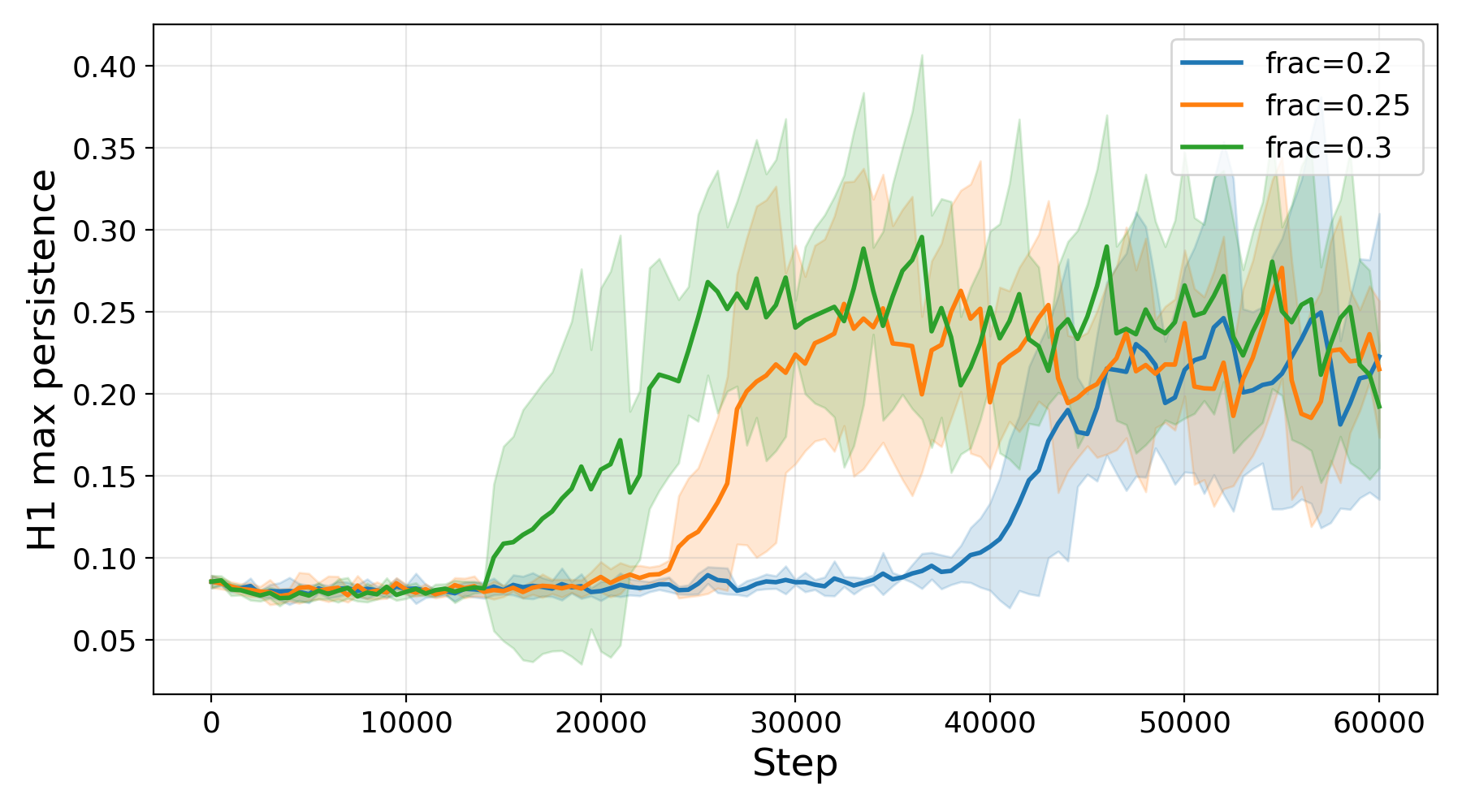}%
    \includegraphics[width=0.45\textwidth]{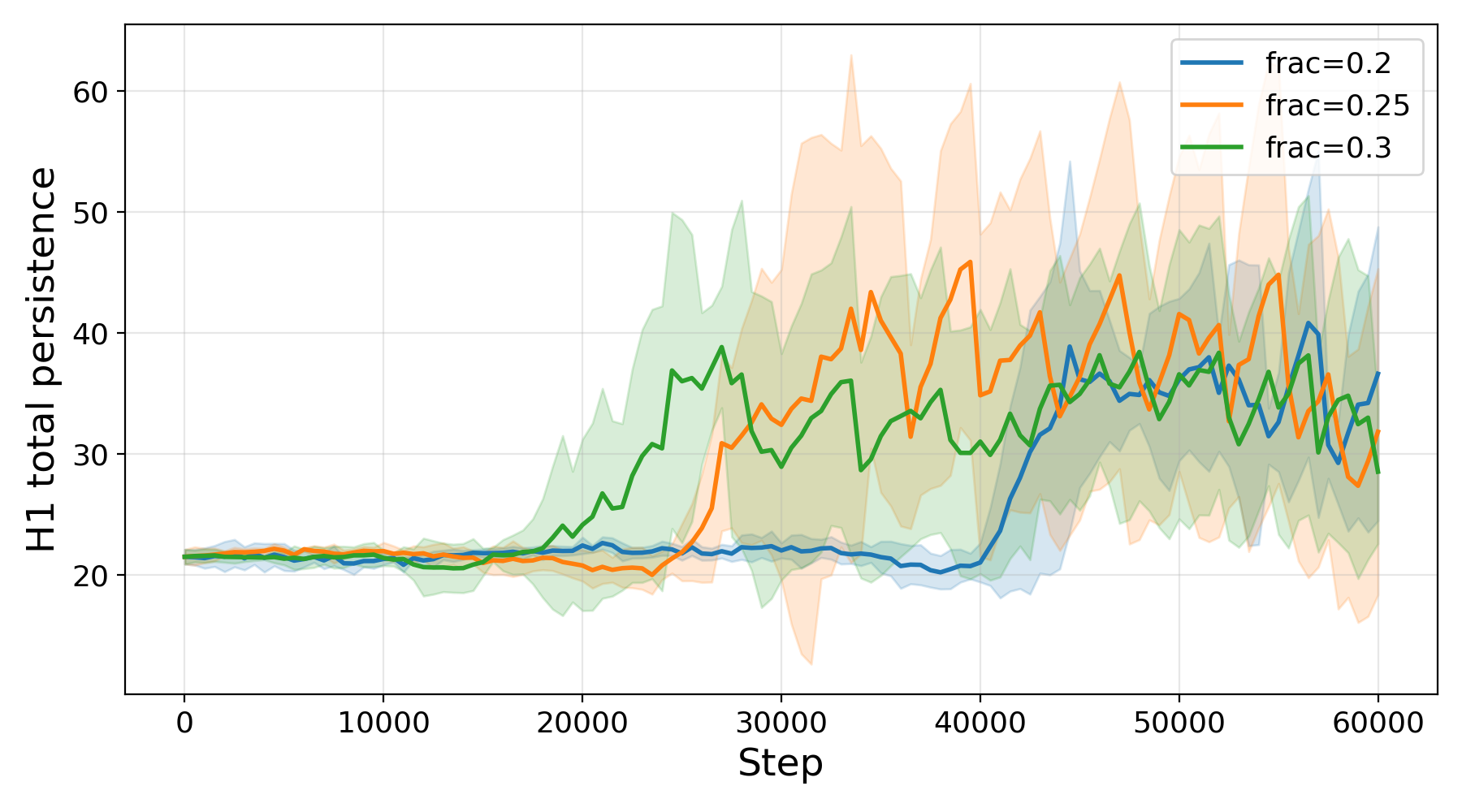}
    \caption{Transformer on modular addition, $p = 197$, averaged across seeds ($\pm 1$ std.\ shaded).  Top row: train/test accuracy (left) and mean pointwise LID on the test set (right).  Bottom row: H1 max persistence (left) and H1 total persistence (right), both computed on the token embedding matrix.  Both H1 metrics rise while LID drops simultaneously at grokking, across all training fractions.}
    \label{fig:transformer_197}
\end{figure}

\textbf{Grokking dynamics.}
Figure~\ref{fig:transformer_197} summarizes training for the transformer on modular addition with $p = 197$, the same modulus studied by \citet{ruppik2026less}, across fractions $\alpha \in \{0.2, 0.25, 0.3\}$, averaged over multiple seeds with $\pm 1$ standard deviation shading.  The accuracy curves (top-left panel) confirm the grokking pattern: all fractions achieve near-perfect training accuracy within a few thousand steps, while test accuracy rises with a delay that grows as $\alpha$ decreases.  For $\alpha = 0.3$, generalization occurs around step 15{,}000--20{,}000; for $\alpha = 0.2$ it is delayed to approximately step 35{,}000--45{,}000.

\begin{figure}[htbp]
    \centering
    
    \begin{subfigure}[b]{0.24\textwidth}
        \centering
        \includegraphics[width=\textwidth]{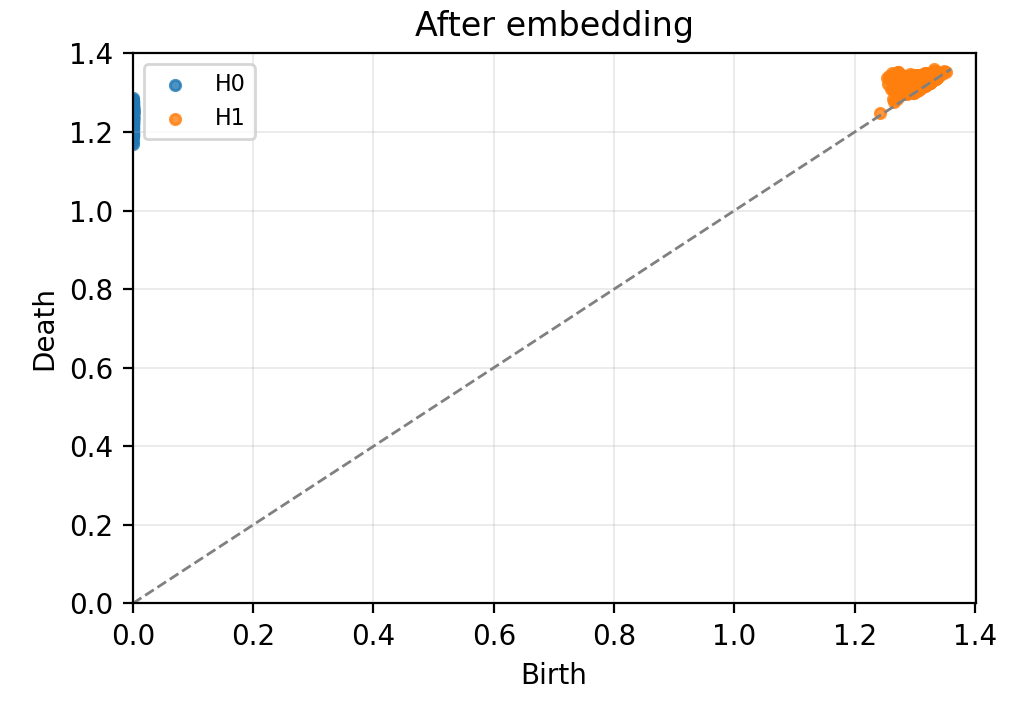}
        \caption{Step 1.}
    \end{subfigure}
    \hfill
    \begin{subfigure}[b]{0.24\textwidth}
        \centering
        \includegraphics[width=\textwidth]{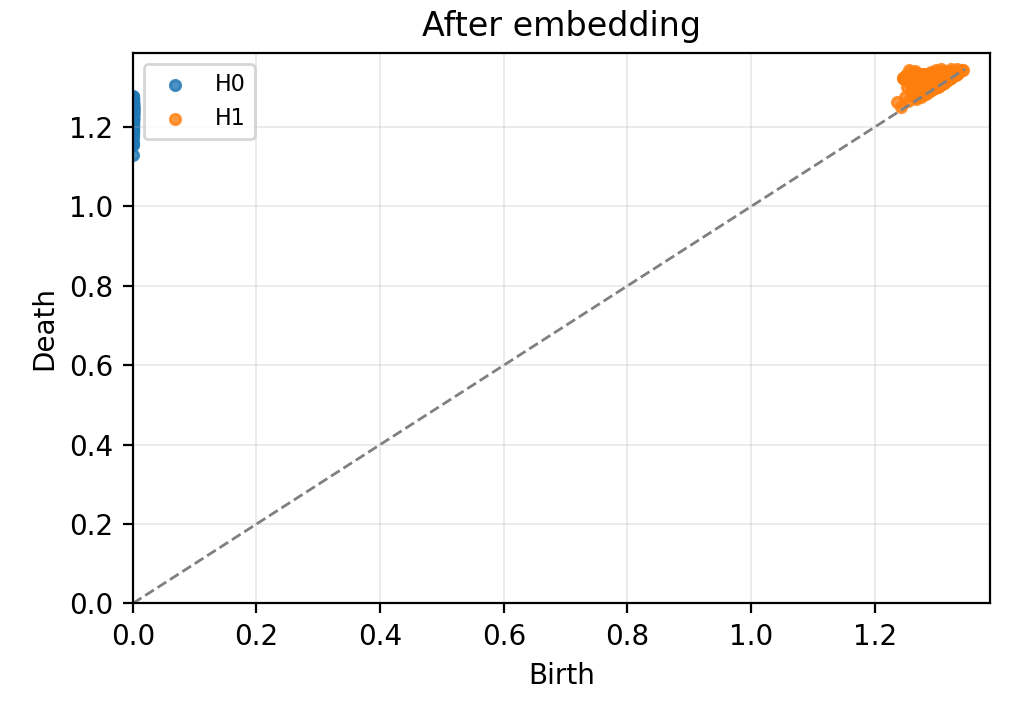}
        \caption{Step 20,000.}
    \end{subfigure}
    \hfill
    \begin{subfigure}[b]{0.24\textwidth}
        \centering
        \includegraphics[width=\textwidth]{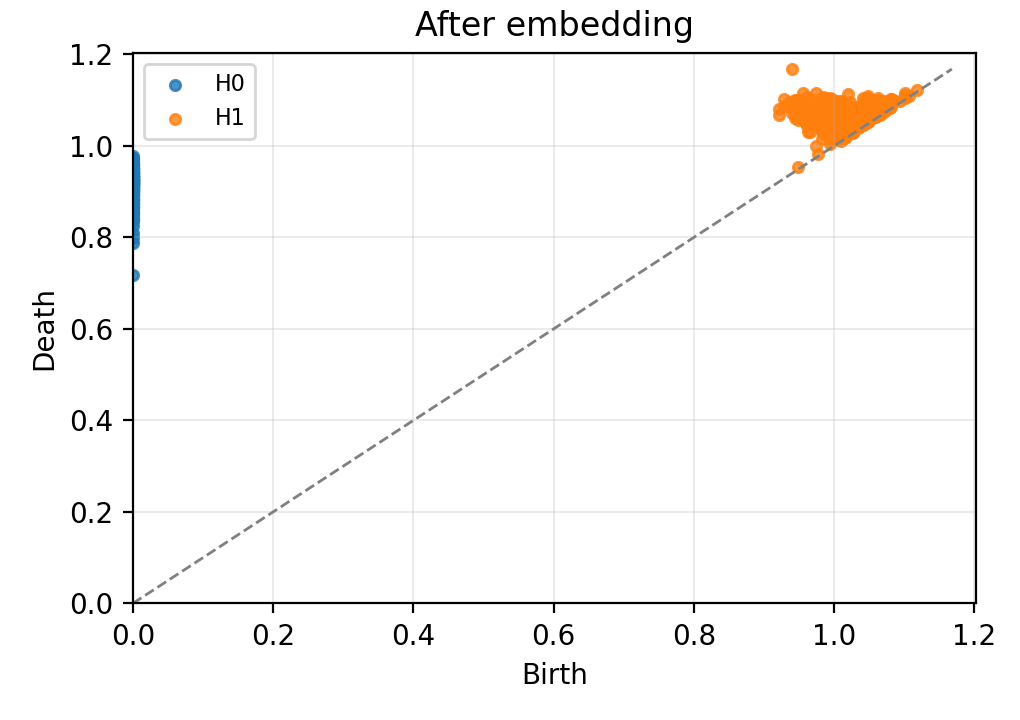}
        \caption{Step 30,000.}
    \end{subfigure}
    \hfill
    \begin{subfigure}[b]{0.24\textwidth}
        \centering
        \includegraphics[width=\textwidth]{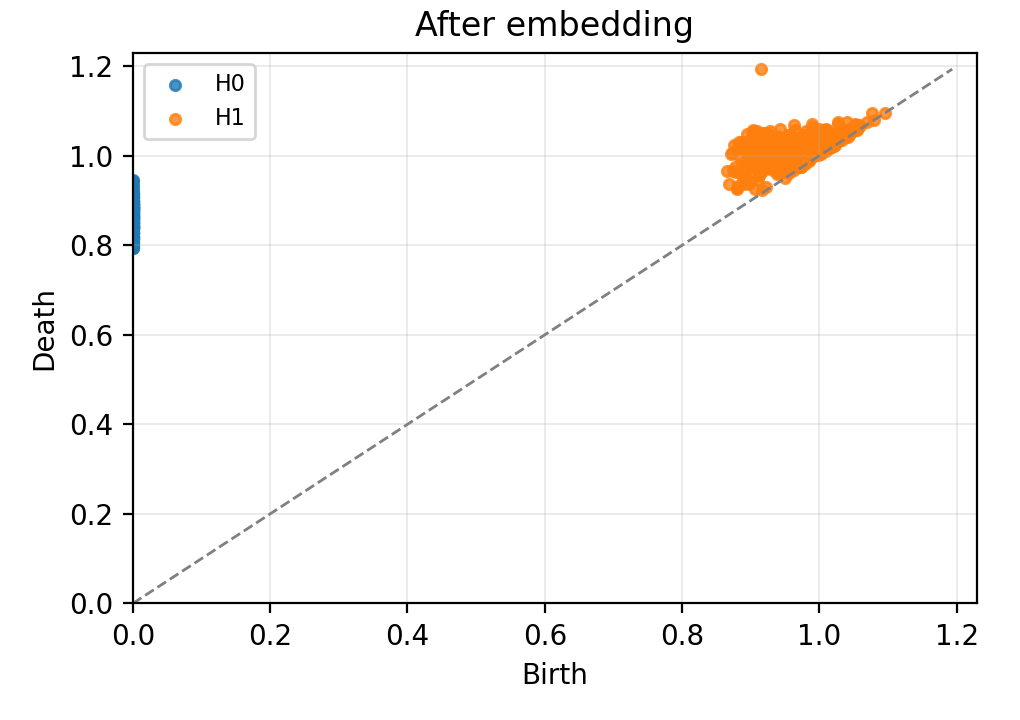}
        \caption{Step 50,000.}
    \end{subfigure}
    
    \caption{Vietoris--Rips persistence diagrams computed from the learned representations of the Transformer model at different stages of training on the modular addition task with prime modulus $p=197$ and training fraction $30\%$. Panels correspond to training steps 1, 20{,}000, 30{,}000, and 50{,}000, respectively. As training progresses and the model transitions into the generalized regime, a dominant long-lived $H_1$ feature increasingly separates from the remaining topological features. This emergence of a highly persistent one-dimensional cycle is consistent with the formation of structured periodic organization in the learned representations following grokking.
}
    \label{fig:pd_example}
\end{figure}

\textbf{Topological grokking signatures.} The maximum persistence value of a point in the degree-1 persistence diagram (H1 max persistence, bottom-left panel of Figure~\ref{fig:transformer_197}) exhibits a clear and reproducible transition: after remaining near its baseline of $\approx 0.075$--$0.08$ throughout the memorization phase, it rises sharply with generalization and stabilizes at $0.20$--$0.25$. The transition timing tracks the grokking step for each fraction, with larger fractions transitioning first. The sum of the persistences of all points in the degree-1 persistence diagram (total H1 persistence, bottom-right panel) shows a corresponding increase from a baseline of $\approx 20$ to values in the range $30$--$50$, consistently across all three fractions. Persistence diagrams across training further reveal the emergence of a dominant long-lived $H_1$ feature that increasingly separates from the remaining topological features as grokking occurs (Figure~\ref{fig:pd_example}). Together, these metrics indicate that grokking coincides with the emergence of coherent 1-dimensional topological structure in the token embedding space, consistent with the cyclic geometry underlying modular arithmetic.


\textbf{Comparison of PH with local intrinsic dimension.} The LID curves (top-right panel of Figure~\ref{fig:transformer_197}) display a complementary but inverse pattern: LID rises to roughly $20$--$25$ during memorization and then drops sharply at the grokking step, stabilizing near $5$.  The initial rise reflects the network assembling a high-dimensional, unstructured representation of the training data; the subsequent drop signals compression onto a low-dimensional manifold.  PH captures the dual aspect that LID misses: the global loop structure that becomes geometrically apparent only after this compression.  Neither metric alone characterizes the transition, but together they paint a coherent picture of a representational phase transition.

\begin{figure}[htbp]
    \centering

    \begin{subfigure}[b]{0.4\textwidth}
        \centering
        \includegraphics[width=\linewidth]{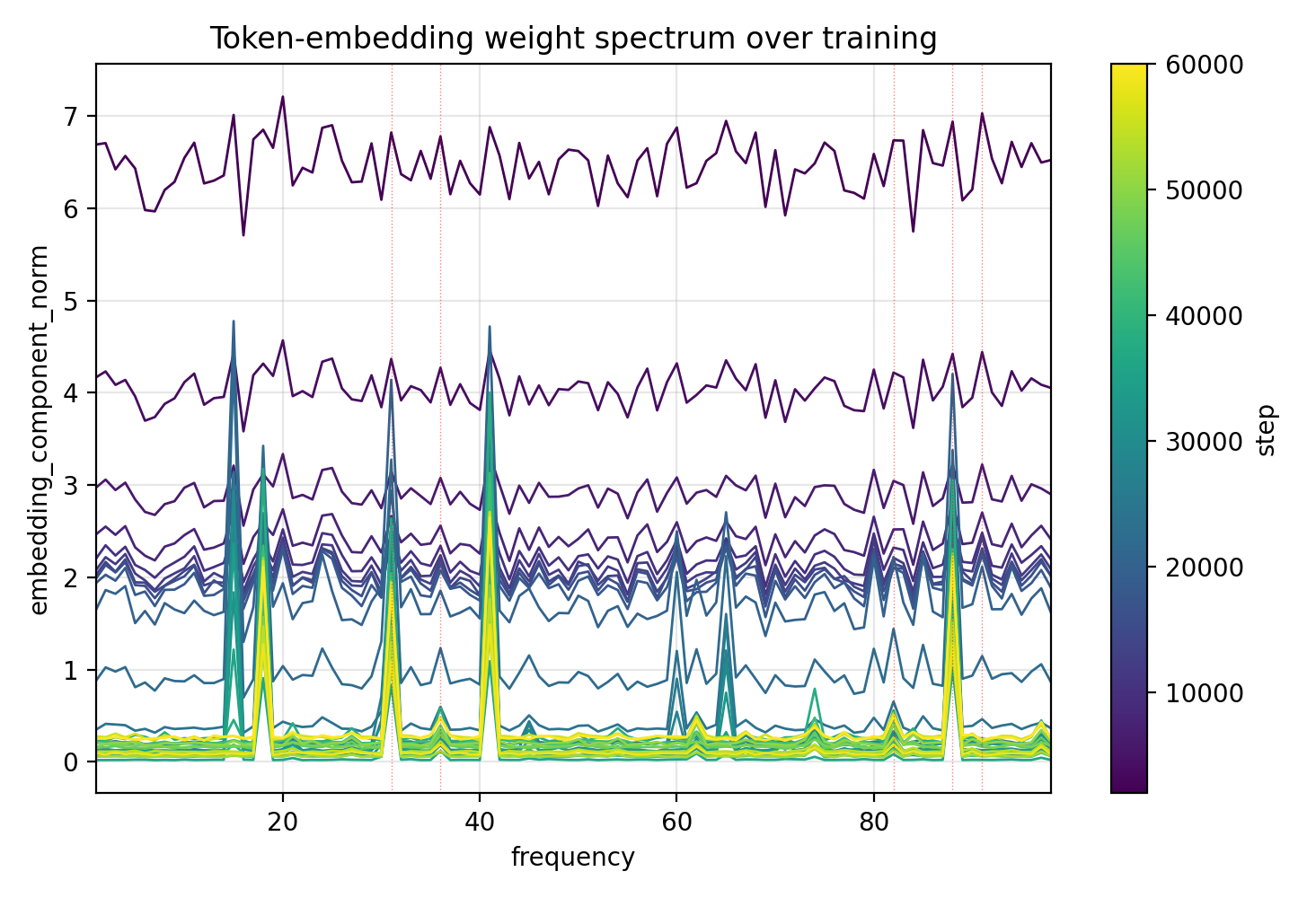}
        \caption{Evolution of the token-embedding Fourier spectrum over training.}
        \label{fig:fourier1}
    \end{subfigure}
    \hfill
    \begin{subfigure}[b]{0.45\textwidth}
        \centering
        \includegraphics[width=\linewidth]{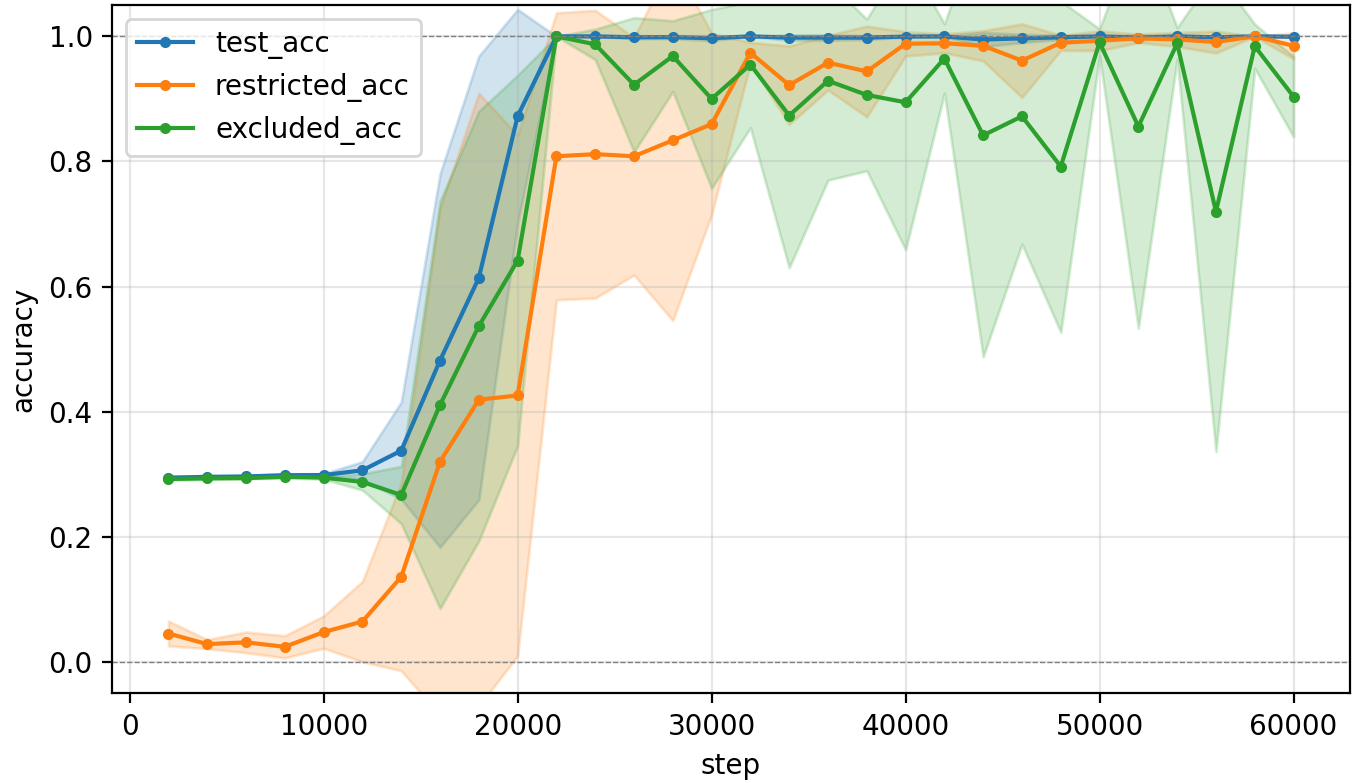}
        \caption{Comparison of test accuracy with restricted and excluded accuracies.}
        \label{fig:fourier2}
    \end{subfigure}

    \caption{Fourier-based mechanistic analysis of grokking on modular addition ($p=197$, $\alpha=0.3$). (a) Evolution of the token-embedding Fourier spectrum during training. (b) Test accuracy together with restricted and excluded Fourier accuracies, showing the emergence of dominant frequency components during grokking.}
    \label{fig:fourier}
\end{figure}

\textbf{Comparison with Fourier analysis \citep{nanda2023progress}.} To contextualize the PH results, we also examine the Fourier-based mechanistic analysis of grokking \citep{nanda2023progress}. We study a representative training run with $\alpha = 0.3$. As shown in Figure~\ref{fig:fourier1}, the Fourier spectrum of the token embedding matrix is initially diffuse but progressively concentrates around a small number of dominant frequencies over the course of training. Figure~\ref{fig:fourier2} shows that the accuracies obtained by restricting to or excluding these dominant frequencies closely track the test accuracy, reflecting the onset of grokking.

These observations provide a complementary spectral perspective on the emergence of structure during training. Whereas Fourier analysis identifies dominant frequency components associated with the learned modular arithmetic circuit, PH captures the evolving geometric and topological organization of the learned representations.


\textbf{Consistency across primes.} Results for $p = 113$ and $p = 149$ are shown in Figures~\ref{fig:transformer_113} and~\ref{fig:transformer_149} in the Appendix.  The H1 transitions remain aligned with the respective grokking steps across all fractions, and the LID inversion is consistently observed across all three primes.

\subsection{MLP Grokking on Modular Arithmetic}
\label{sec:results_mlp}

\begin{figure}[t]
    \centering
    \includegraphics[width=0.45\textwidth]{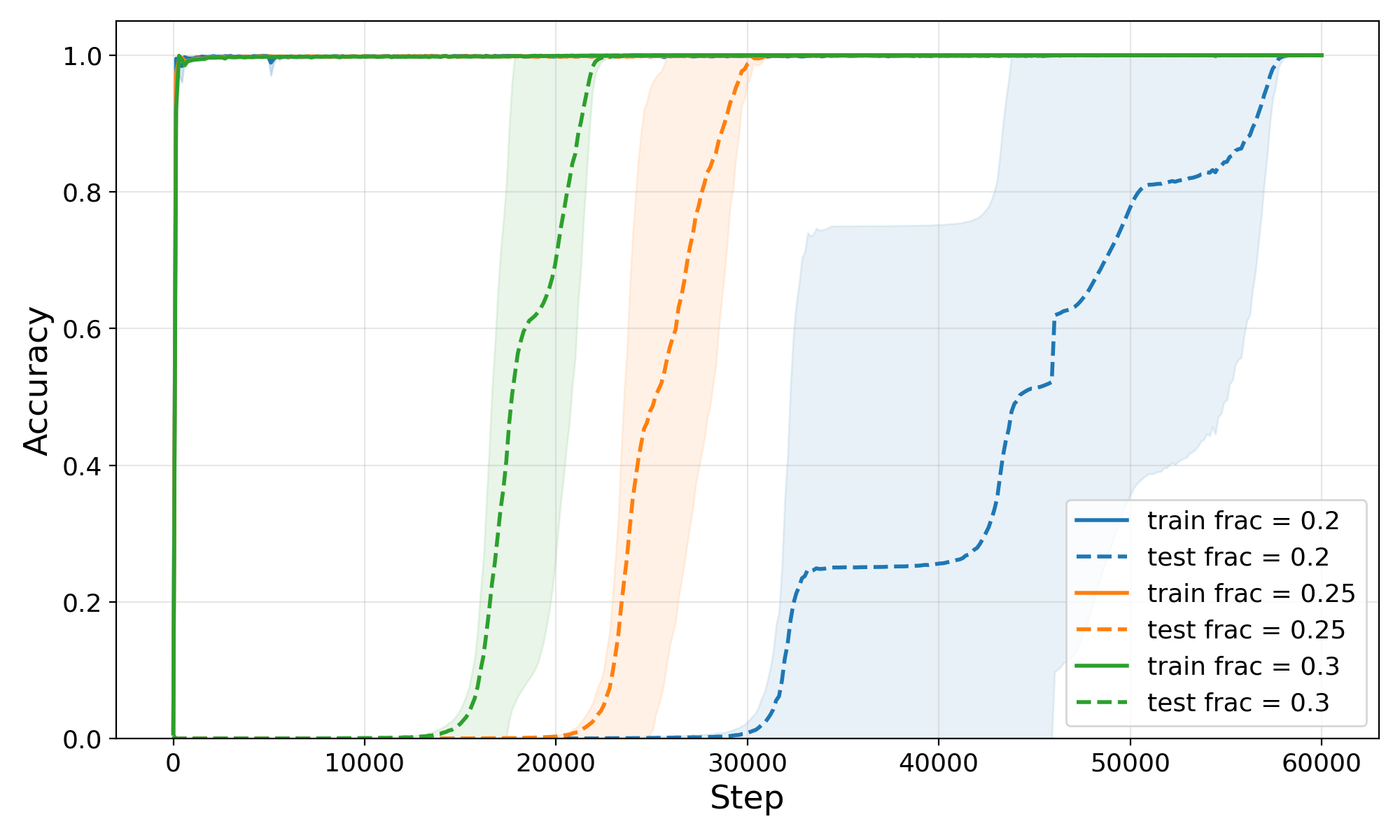}\\[4pt]
    \includegraphics[width=0.45\textwidth]{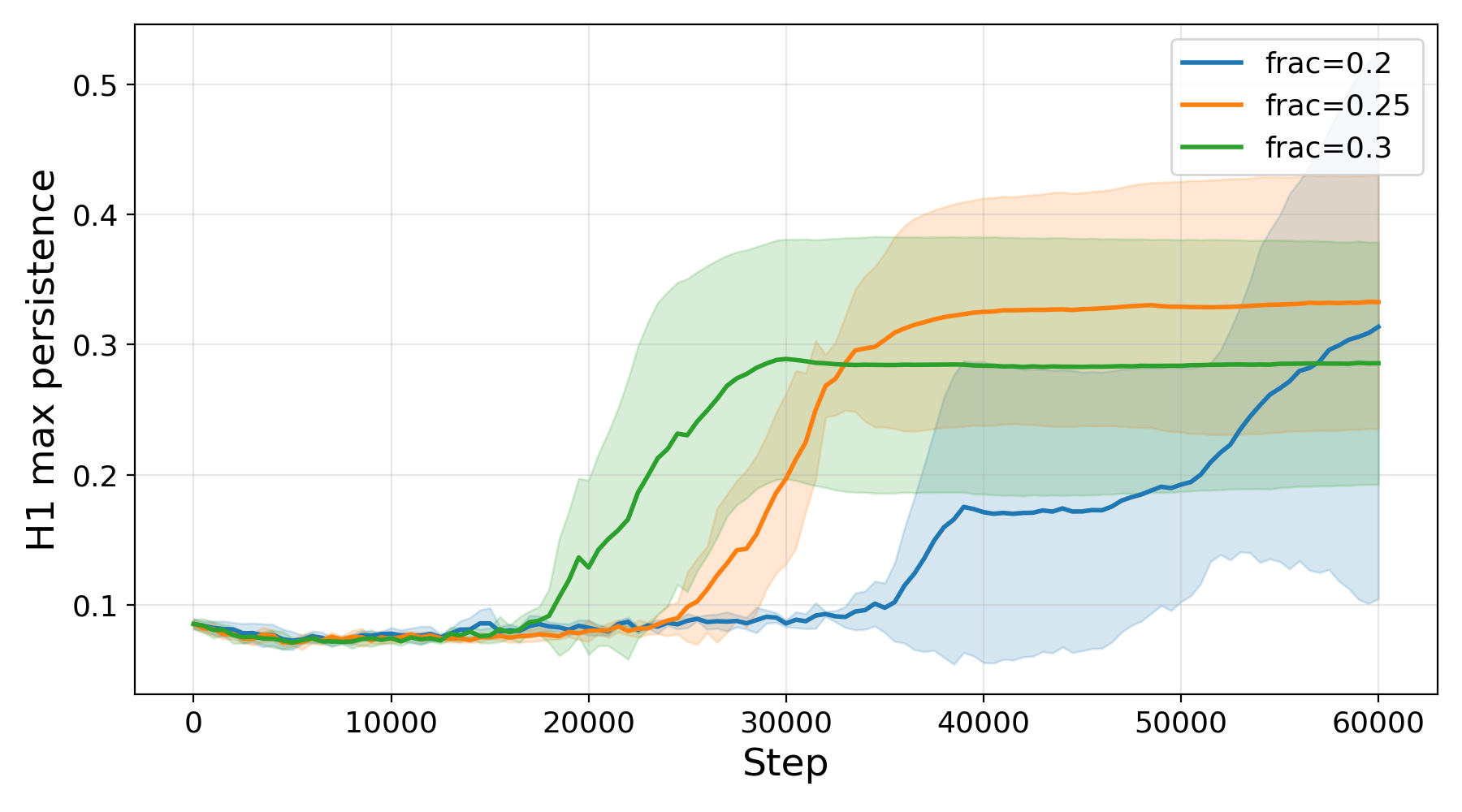}%
    \includegraphics[width=0.45\textwidth]{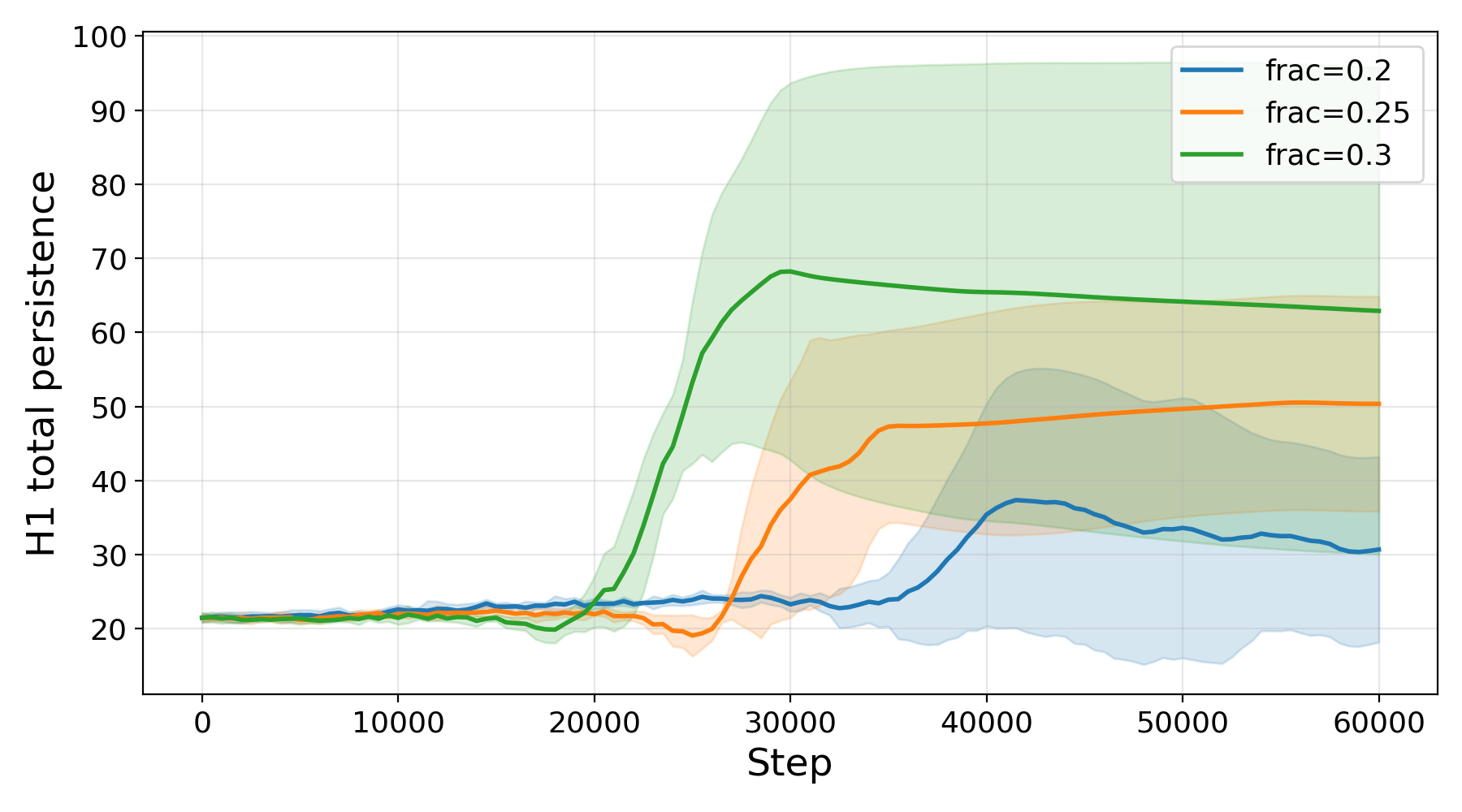}\\[4pt]
    \includegraphics[width=0.45\textwidth]{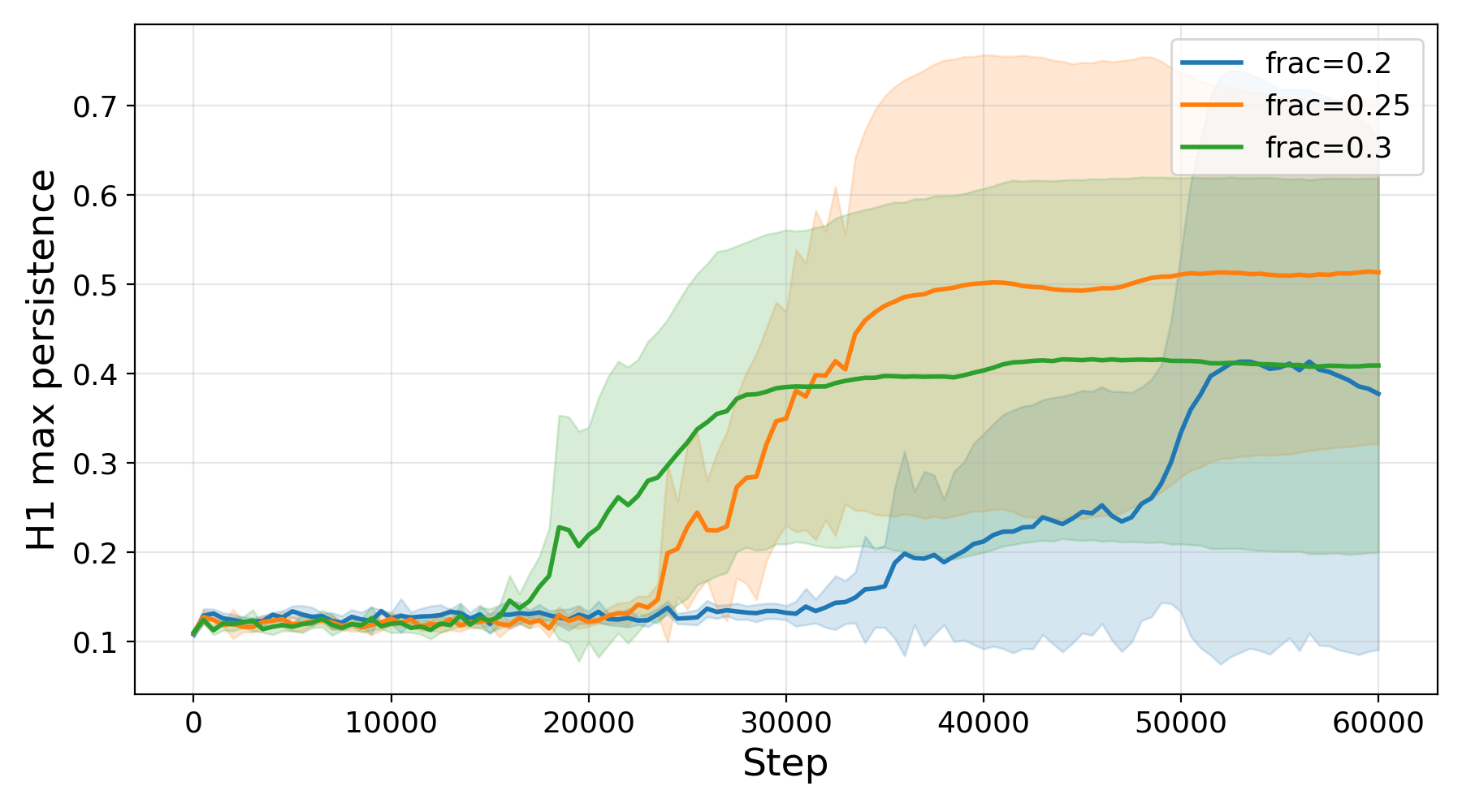}%
    \includegraphics[width=0.45\textwidth]{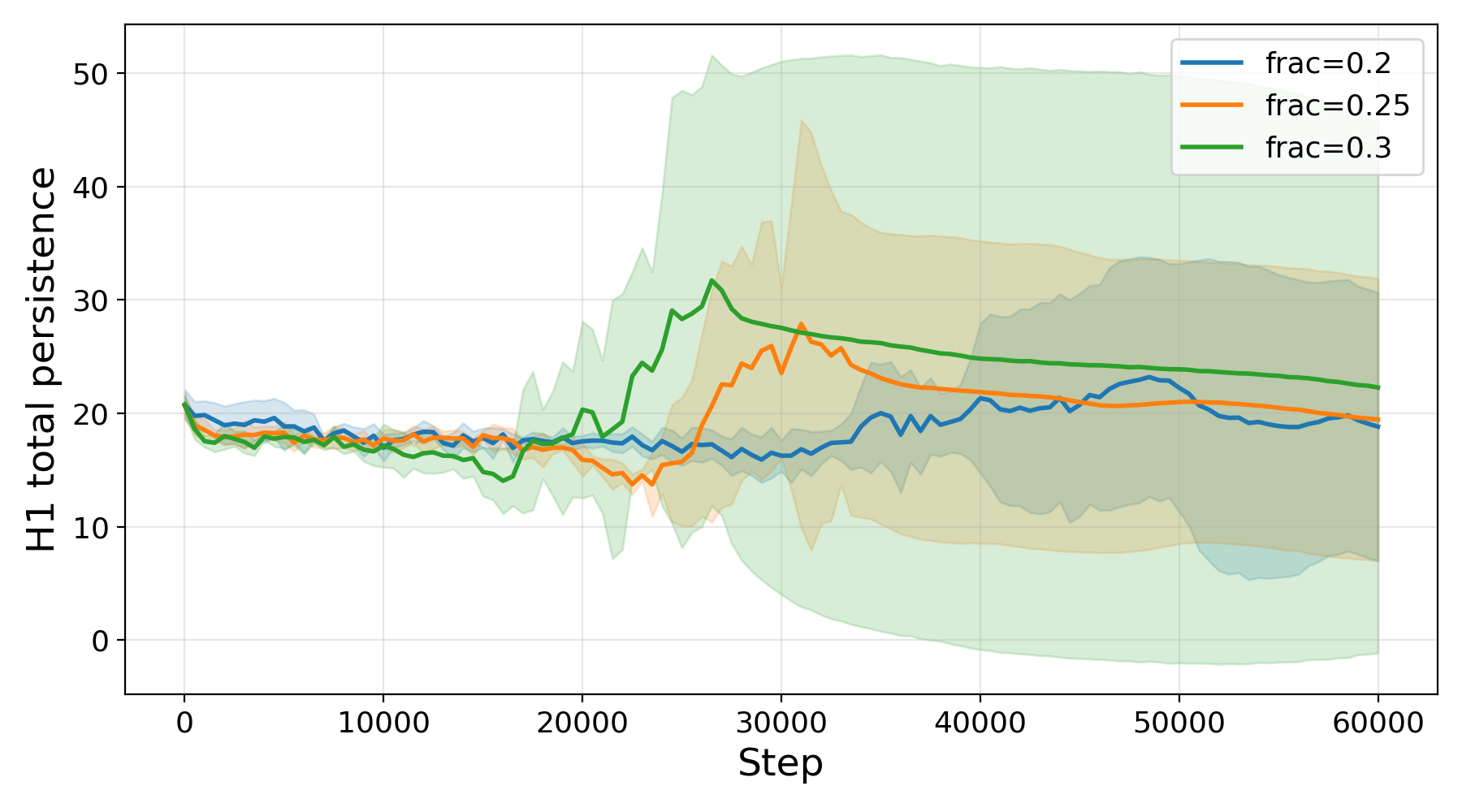}
    \caption{MLP on modular addition, $p = 197$.  Top: train/test accuracy.  Middle row: H1 max (left) and H1 total persistence (right) at the token embedding layer (layer~0) — both rise at grokking for all three fractions, with H1 total showing the most dramatic increase across all settings.  Bottom row: same metrics at the third hidden layer (layer~3) — H1 max rises further, and H1 total also increases modestly, suggesting topological structure strengthens through depth for this larger group.}
    \label{fig:mlp_197}
\end{figure}

\textbf{Grokking dynamics.}
Figure~\ref{fig:mlp_197} shows training for the MLP on modular addition with $p = 197$, the same modulus studied by \citet{ruppik2026less}.  As with the transformer, all fractions memorize quickly while test accuracy generalizes with a fraction-dependent delay, confirming that grokking is not specific to the attention-based architecture.

\textbf{Topological signatures.} H1 max persistence at the token embedding layer (layer~0, middle row of Figure~\ref{fig:mlp_197}) rises from a baseline near $0.08$ to approximately $0.29$--$0.35$ at grokking.  The rise in H1 total persistence is even more striking: starting from a baseline of $\approx 20$, it increases to the range $30$--$70$ for all three fractions at the grokking step.  Both metrics rise for all fractions, including $\alpha = 0.2$, indicating that the topological signal is robust across data regimes.

\textbf{Layer-wise behavior.}
At the third hidden layer (bottom row of Figure~\ref{fig:mlp_197}), H1 max rises even further (to $0.4$--$0.5$), suggesting that the cyclic structure becomes more geometrically pronounced in deeper representations.  Unlike the smaller primes $p = 113$ and $p = 149$ (see Appendix~\ref{app:mlp_primes}), where H1 total at layer~3 \emph{decreases}, reflecting a consolidation into fewer but stronger cycles, for $p = 197$ the H1 total at layer~3 shows a modest positive increase alongside the max.  We attribute this difference to the larger group size: $\mathbb{Z}/197\mathbb{Z}$ sustains more topological structure across all layers, rather than compressing it into a single dominant feature.  The fact that the signal is visible as early as the embedding layer confirms that structure emerges at the very first stage of processing, consistent with the analysis of \citet{nanda2023progress}.

\textbf{Consistency across primes.}
Results for $p = 113$ and $p = 149$ are shown in Figures~\ref{fig:mlp_113} and~\ref{fig:mlp_149} in the Appendix.  The embedding-layer H1 max rise is present in both, and the layer-3 inversion (max up, total down) is clearly visible, particularly for $p = 149$.

\subsection{Ablation Study: Label Permutation and Topological Robustness}
\label{sec:ablations}

To assess whether the observed topological transitions are specifically associated with model generalization, rather than reflecting generic optimization dynamics or memorization, we performed a label-permutation ablation study. By progressively corrupting the training objective through controlled label permutation, we create settings in which the underlying modular structure becomes increasingly difficult to recover. This allows us to identify the point at which grokking fails and, in turn, evaluates whether our PH measures continue to track the representational phase transition.

The training configurations are as described in Section \ref{sec:experimental_setup}, with the Transformer model trained using a $0.3$ training fraction and the MLP model using a $0.2$ training fraction. For each architecture, we introduced a label permutation fraction, denoted $P_{\mathrm{frac}}$, corresponding to the proportion of training labels randomly permuted among the training samples and explored $P_{\mathrm{frac}}$ ranging from $0\%$ to $100\%$. All models were trained for a fixed budget of 60,000 optimization steps to capture both grokking and post-grokking dynamics.

We focus on the label-permutation ablation results for the Transformer model, while the corresponding MLP results for modular arithmetic are provided in Appendix~\ref{app:mlp_ablations}. Consistent with the Transformer, the MLP exhibits qualitatively similar behavior, with improvements in generalization strongly coupled to the topological signatures captured by the PH statistics.

\begin{figure}[htbp]
    \centering

    \begin{subfigure}[b]{0.6\textwidth}
        \centering
        \includegraphics[width=\linewidth]{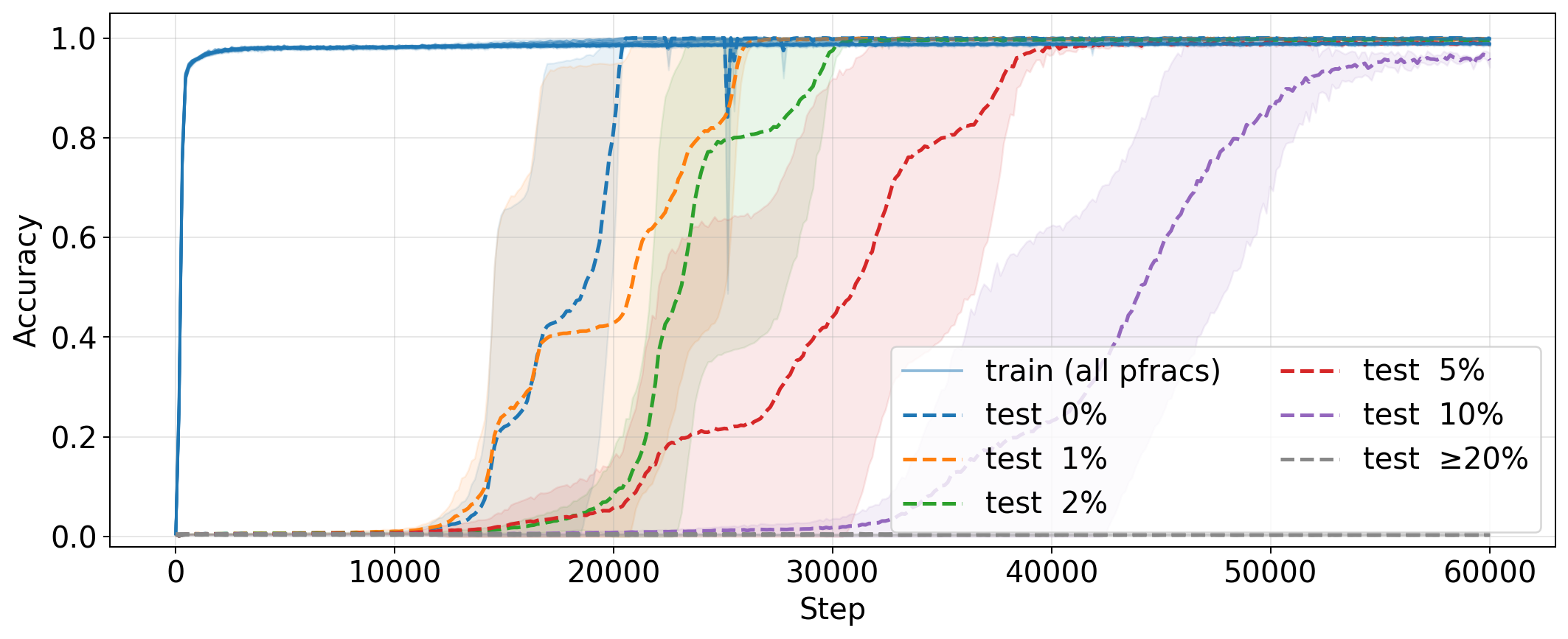}
    \end{subfigure}

    \vspace{0.5em}

    \begin{subfigure}[b]{0.49\textwidth}
        \centering
        \includegraphics[width=\linewidth]{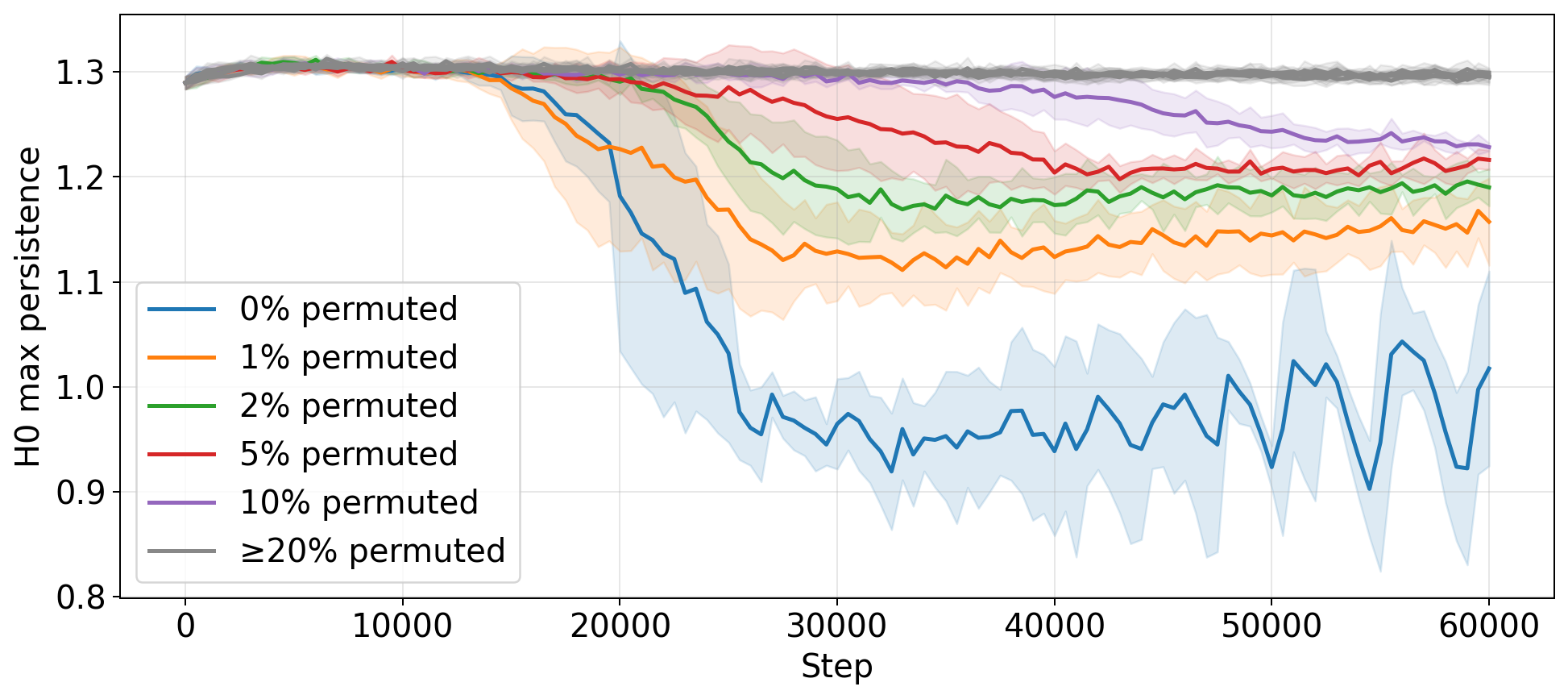}
    \end{subfigure}
    \hfill
    \begin{subfigure}[b]{0.49\textwidth}
        \centering
        \includegraphics[width=\linewidth]{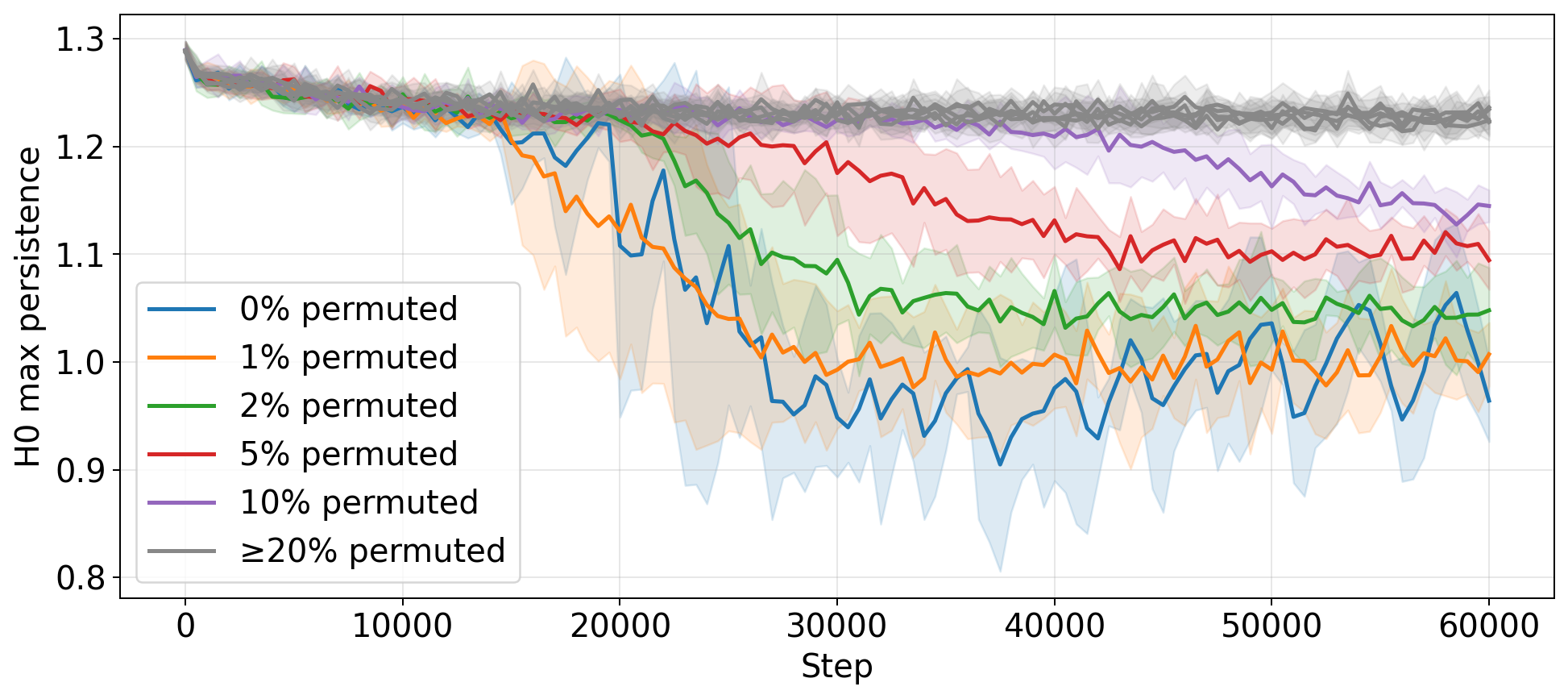}
    \end{subfigure}




    \begin{subfigure}[b]{0.49\textwidth}
        \centering
        \includegraphics[width=\linewidth]{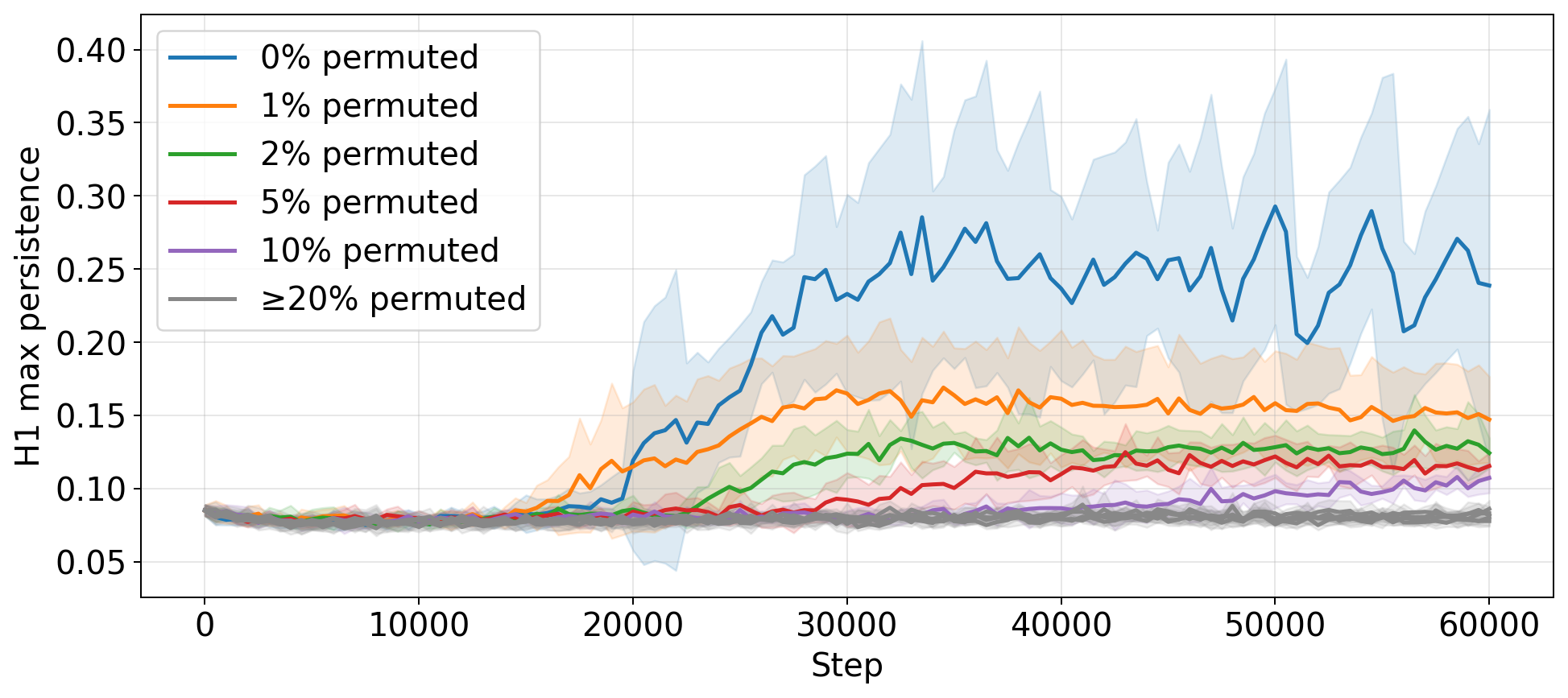}
    \end{subfigure}
    \hfill
    \begin{subfigure}[b]{0.49\textwidth}
        \centering
        \includegraphics[width=\linewidth]{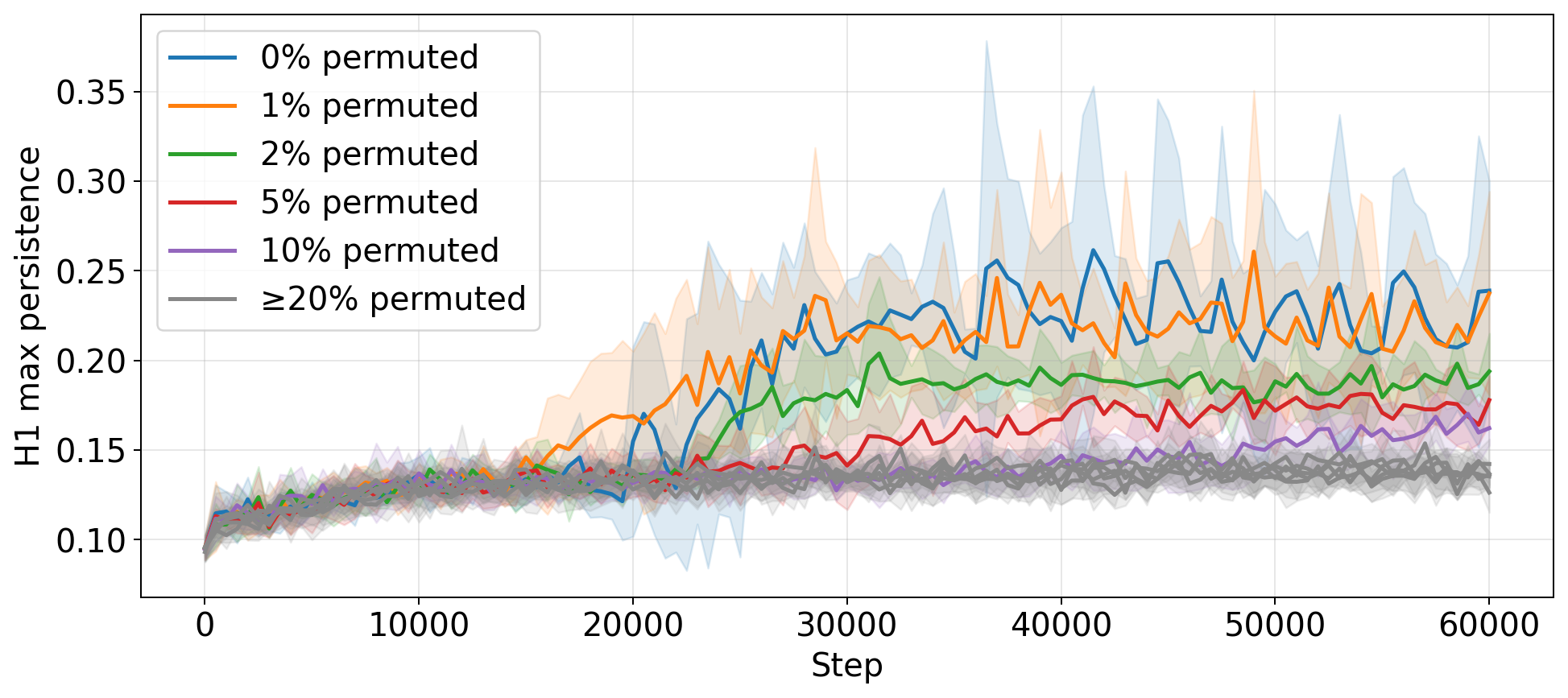}
    \end{subfigure}



    \caption{Training dynamics under label permutation for the Transformer model. 
    The top panel shows test accuracy across training for different label permutation fractions $P_{\mathrm{frac}}$. 
    The remaining panels show the corresponding Vietoris--Rips PH statistics computed from the learned representations, with the left and right columns corresponding to the embedding layer (layer 0) and first layer (layer 1), respectively. 
    Rows 2 and 3 report $H_0$ and $H_1$ maximum persistence, respectively.
    For low permutation levels, successful grokking is accompanied by a pronounced decrease in $H_0$ persistence and the emergence of a dominant $H_1$ cycle, consistent with the formation of a structured periodic representation. 
    As label corruption increases, both generalization and the associated topological transition progressively deteriorate.}
    \label{fig:transformer_ablation}
\end{figure}

\textbf{Topological signatures with grokking.} For permutation levels up to $P_{\mathrm{frac}} \leq 10\%$, the Transformer model consistently exhibited grokking behavior. Within this regime, we observed strong and highly consistent associations between model generalization performance and the PH statistics, particularly in the embedding layer and first layer (Appendix Table~\ref{tab:ph_correlations_transformer}). Across these early layers, $H_0$-based statistics exhibited strong negative correlations with test accuracy, with Spearman coefficients reaching as low as $\rho=-0.91$ for total $H_0$ persistence and remaining consistently below $-0.49$ across all grokking permutation levels. Conversely, maximum $H_1$ persistence exhibited strong positive correlation with test accuracy, reaching $\rho=0.81$ in layer 1 and remaining consistently positive across the same regimes. These results indicate that improved generalization is accompanied by increased connectivity together with the emergence of a dominant one-dimensional topological feature in the learned representation space. Figure~\ref{fig:transformer_ablation} illustrates the evolution of these PH signatures over training under increasing label permutation, further illustrations can be found in the Appendix Section \ref{app:transformer_ablations}.

\textbf{Breakdown under increased label corruption.} 
At higher permutation levels ($P_{\mathrm{frac}} \geq 20\%$), the model was not able to generalize within the 60,000-step training window. Instead, the model remained in a memorization-dominated regime characterized by high training accuracy but poor out-of-sample performance.

Crucially, under these conditions, the previously observed correlations between test accuracy and PH measures largely disappeared. This suggests that the topological signatures identified by our PH measures are not simply artifacts of training duration or loss minimization, but instead are specifically associated with the emergence of model generalization.

\textbf{Temporal dynamics.}
To further characterize the relationship between predictive performance and topological reorganization, we analyzed the cross-correlation function (CCF) between the first-order differences of test accuracy and the PH measures.

Across all grokking runs of the Transformer model, cross-correlation analysis revealed a consistent temporal lag centered around 1,000 training steps between changes in predictive performance and the corresponding topological transition. Changes in test accuracy consistently preceded shifts in the PH statistics (e.g., total $H_0$ persistence), with the full set of cross-correlation plots provided in Appendix~\ref{app:transformer_ablations}. This suggests a two-stage learning dynamic in which the model first discovers a functionally correct input--output mapping, followed by a delayed geometric reorganization of the latent space into a more compact and topologically coherent representation.

    

\subsection{Topological Behavior on MNIST}
\label{sec:results_mnist}

\begin{figure}[t]
    \centering
    \includegraphics[width=0.9\textwidth]{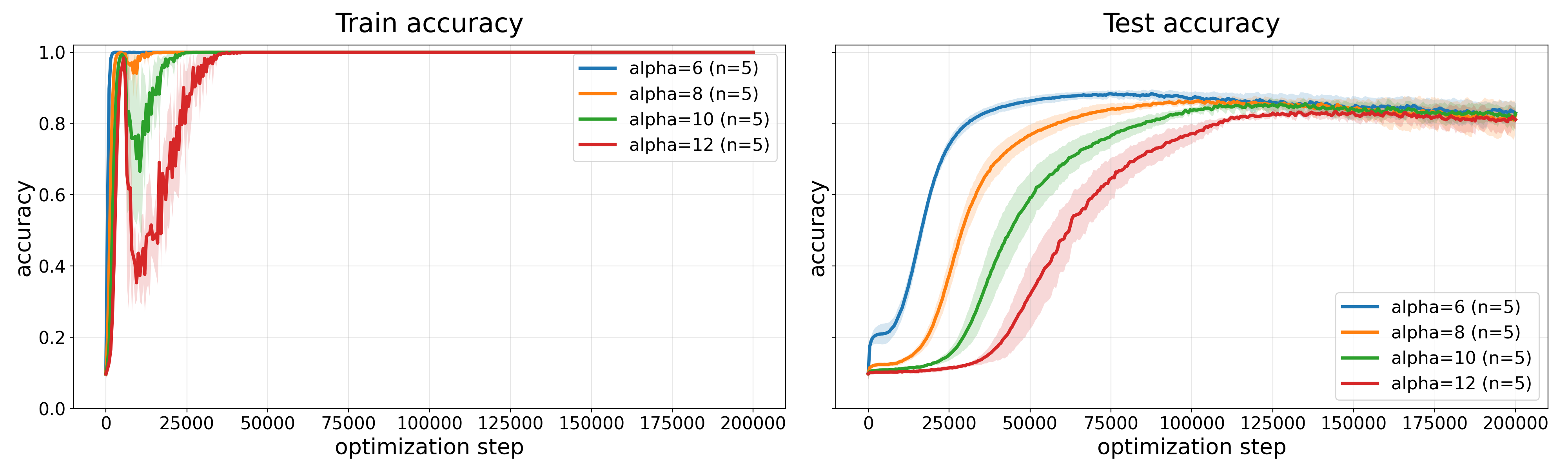}\\%
    \includegraphics[width=0.45\textwidth]{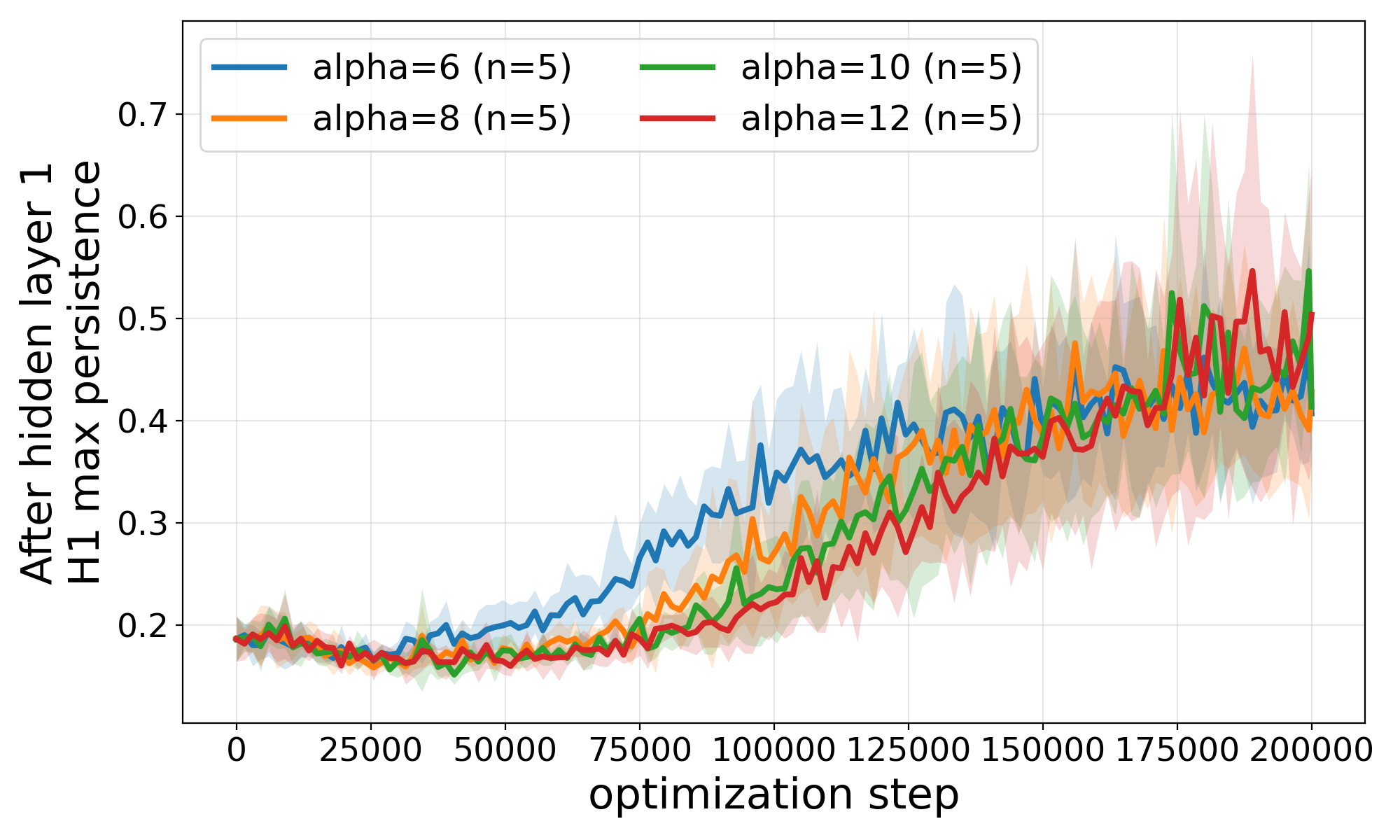}%
    \includegraphics[width=0.45\textwidth]{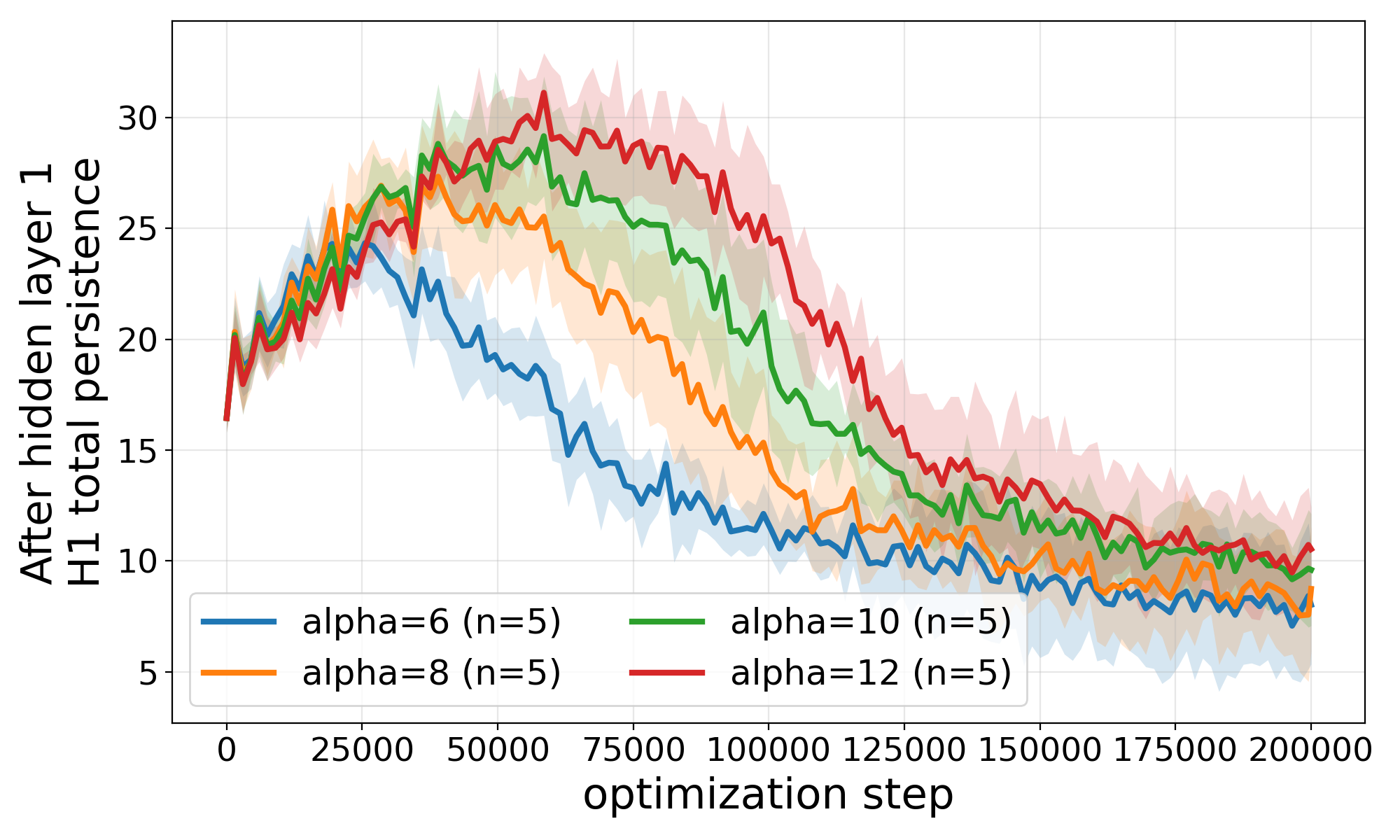}
    \caption{MNIST results, layer~1 (after the first hidden layer). Top row: train/test accuracy. Bottom row, left to right: $H_1$ max persistence and $H_1$ total persistence. Unlike the modular arithmetic setting, $H_1$ max persistence increases gradually without a sharp transition, while $H_1$ total persistence peaks early and subsequently declines. No clearly dominant long-lived cycle emerges, consistent with the absence of a simple global cyclic structure in the MNIST task.}
    \label{fig:mnist}
\end{figure}

To assess whether the observed topological signatures generalize beyond tasks with latent cyclic structure, we apply the same analysis to representations learned on MNIST, a classification task whose underlying geometry is substantially more heterogeneous and not naturally cyclic. Experiments follow the implementation of \citet{liu2023omnigrok}, with varying weight initialization scales. In this setting, $\alpha$ denotes the initialization scaling factor relative to the default PyTorch initialization rather than the training fractions used in the modular arithmetic experiments. Results are averaged over five random seeds. As shown in Figure~\ref{fig:mnist}, training accuracy on MNIST saturates quickly for all values of $\alpha$, while test accuracy increases more gradually.

\textbf{$H_1$ max persistence.} The $H_1$ max persistence (left panel in the bottom row of Figure~\ref{fig:mnist}) increases gradually throughout training across all conditions. Unlike the modular arithmetic setting, no sharp transition is observed: the increase is comparatively slow and noisy rather than concentrated around a distinct grokking step.

\textbf{$H_1$ total persistence.}
The $H_1$ total persistence (right panel in the bottom row of Figure~\ref{fig:mnist}) displays a qualitatively different trajectory from the modular arithmetic case. It rises early in training to a peak around steps 25{,}000--50{,}000, before gradually declining throughout the remainder of training. This behavior suggests an initial increase in distributed topological complexity during early fitting, followed by a progressive consolidation of the learned representations. In contrast to modular arithmetic, no sustained increase in total persistence is observed, and no clearly dominant long-lived $H_1$ feature emerges.

\section{Conclusion: Limitations and Future Work}
\label{sec:conclusion-and-limitations}

We studied grokking through the lens of topology using PH on learned neural representations. Across modular arithmetic tasks, architectures, and training regimes, we identified a consistent topological signature of grokking characterized by increasing $H_1$ persistence and the emergence of dominant long-lived topological features. Ablation studies further showed that these signatures disappear when grokking fails to occur, indicating that they are associated with generalization rather than memorization.

\textbf{Limitations.} While the observed topological transitions are clear in modular arithmetic, the behavior on MNIST is substantially more diffuse, suggesting that interpretable global topological structure may depend strongly on the underlying geometry of the task. More broadly, PH provides descriptive geometric summaries rather than mechanistic explanations of learning dynamics.

Overall, our results suggest that PH offers a useful framework for studying representation learning and the emergence of structure during training. Future work includes extending these analyses to tasks with richer latent geometry, larger-scale architectures, and more expressive topological summaries.


\clearpage
\bibliographystyle{abbrvnat}
\bibliography{grokking-ph}


\clearpage
\appendix

\section{Additional Transformer Results}
\label{app:transformer_primes}

\begin{figure}[h]
    \centering
    \includegraphics[width=0.49\textwidth]{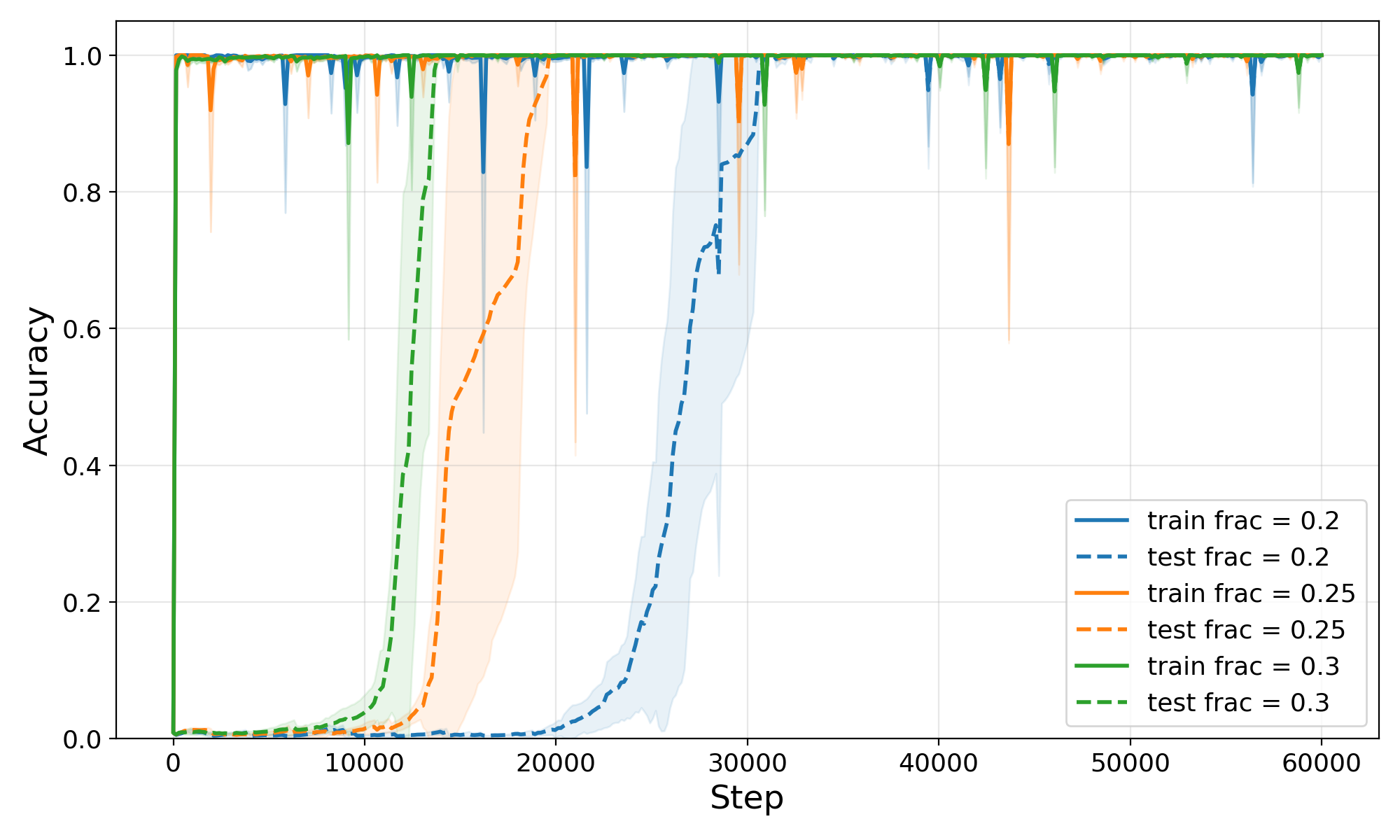}%
    \includegraphics[width=0.49\textwidth]{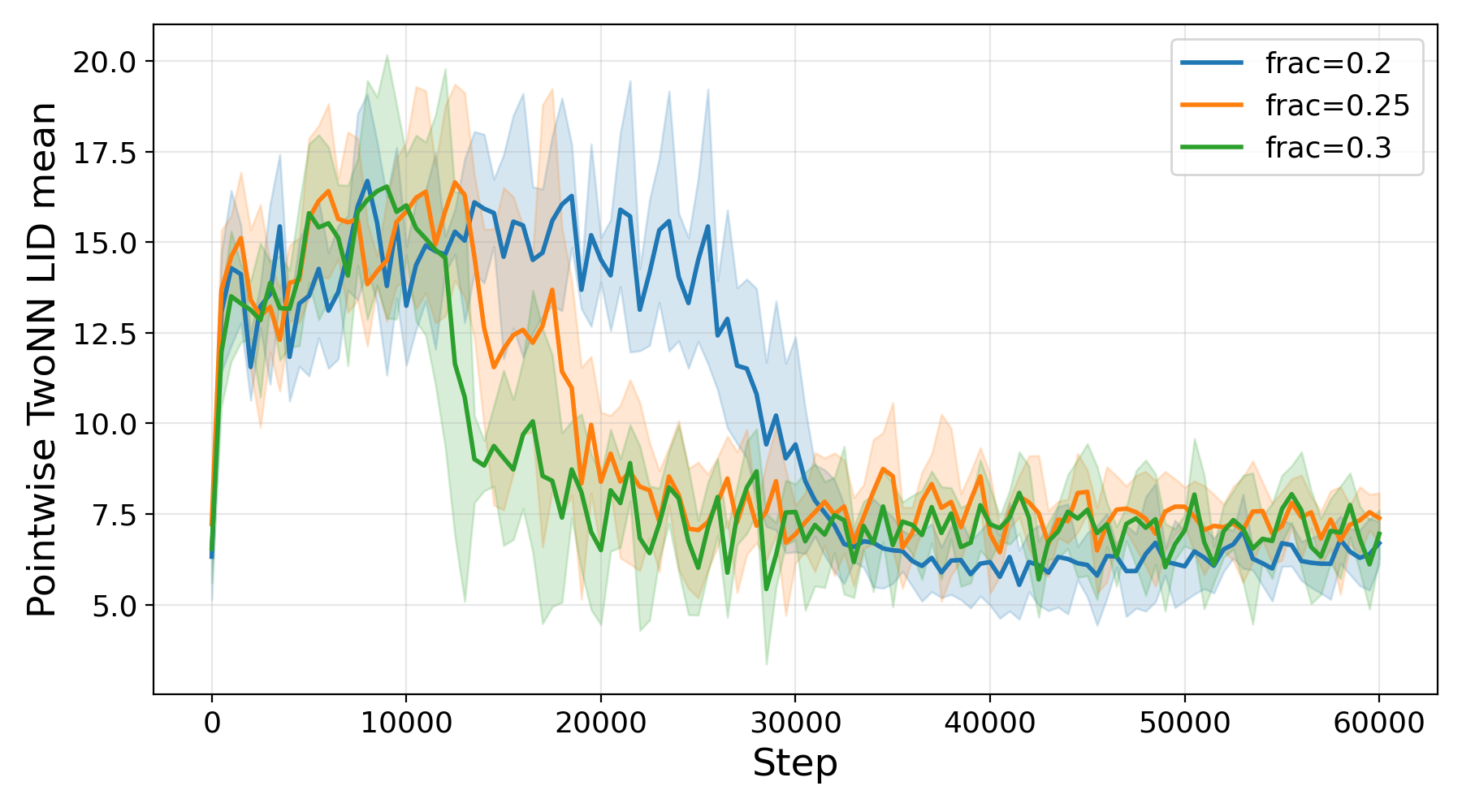}\\[4pt]
    \includegraphics[width=0.49\textwidth]{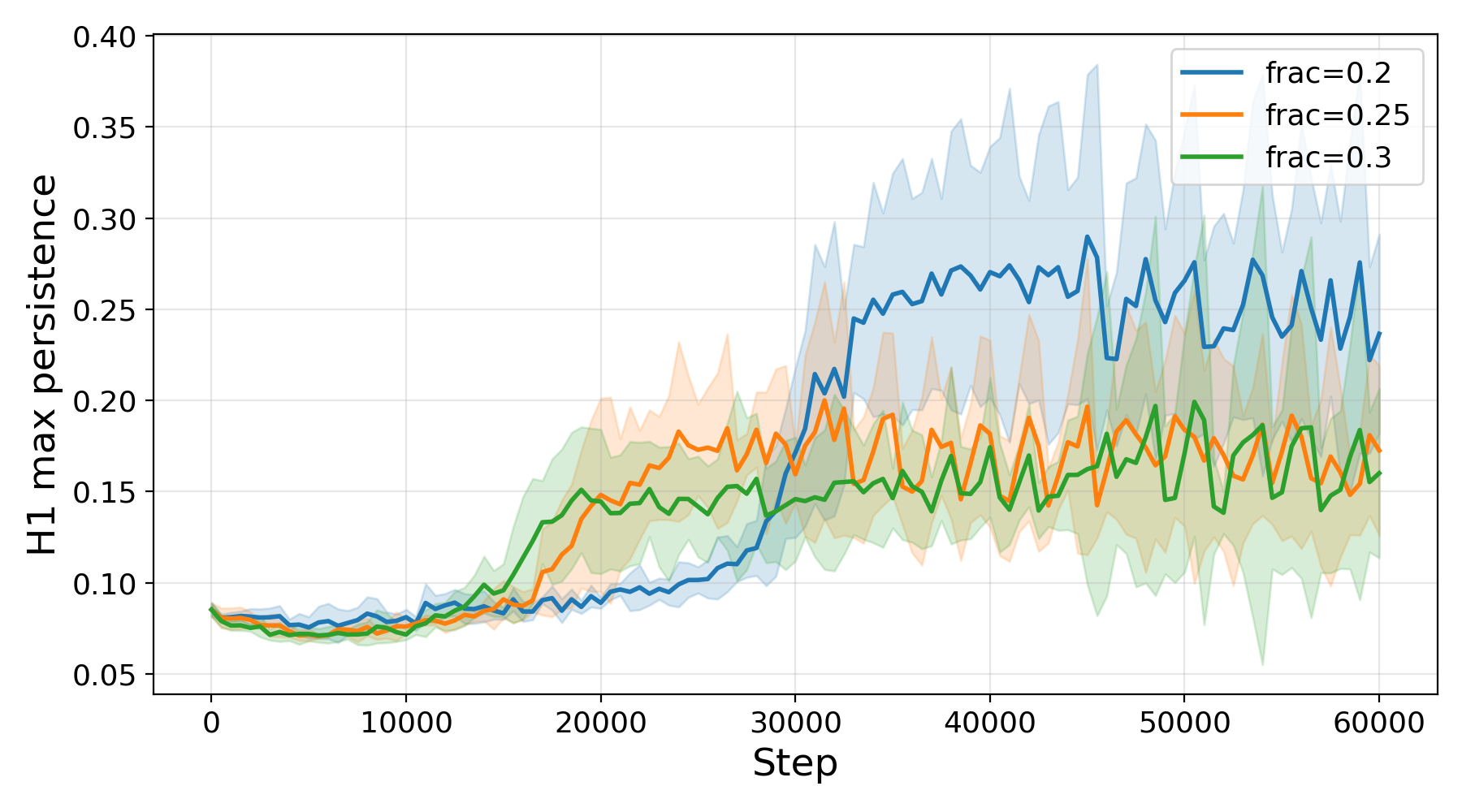}%
    \includegraphics[width=0.49\textwidth]{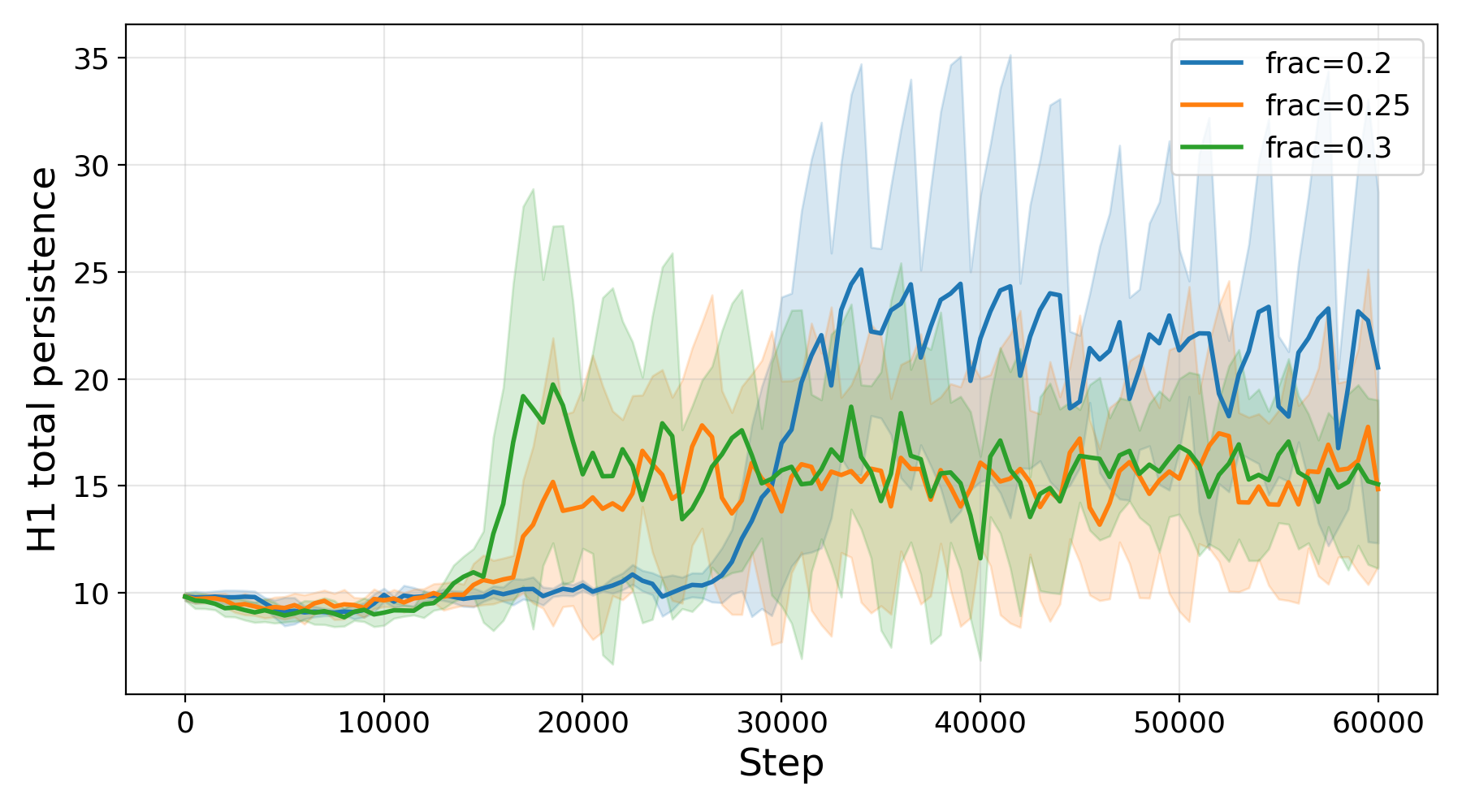}
    \caption{Transformer on modular addition, $p = 113$.  Same layout as Figure~\ref{fig:transformer_197}: train/test accuracy (top-left), LID (top-right), H1 max persistence (bottom-left), H1 total persistence (bottom-right).}
    \label{fig:transformer_113}
\end{figure}

\begin{figure}[h]
    \centering
    \includegraphics[width=0.49\textwidth]{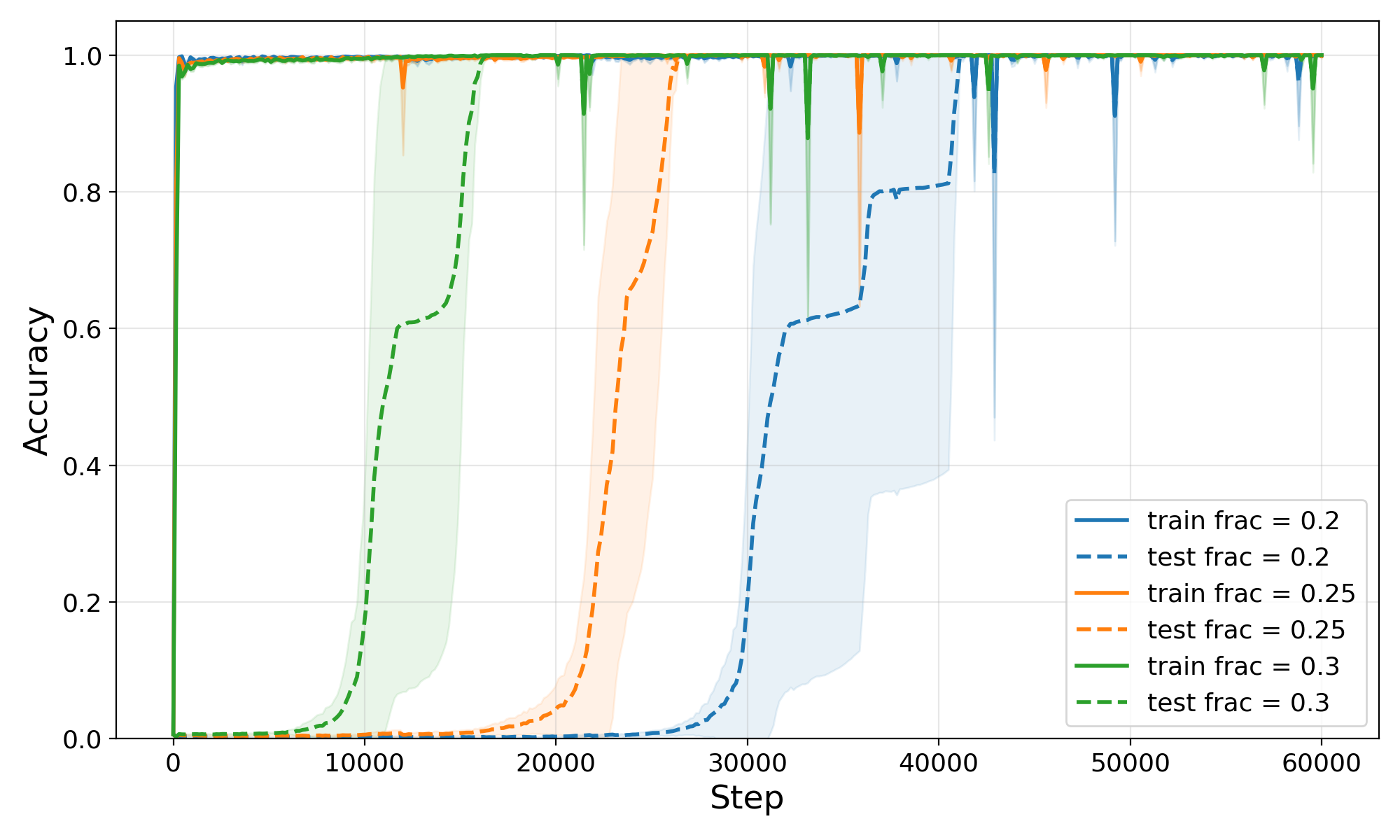}%
    \includegraphics[width=0.49\textwidth]{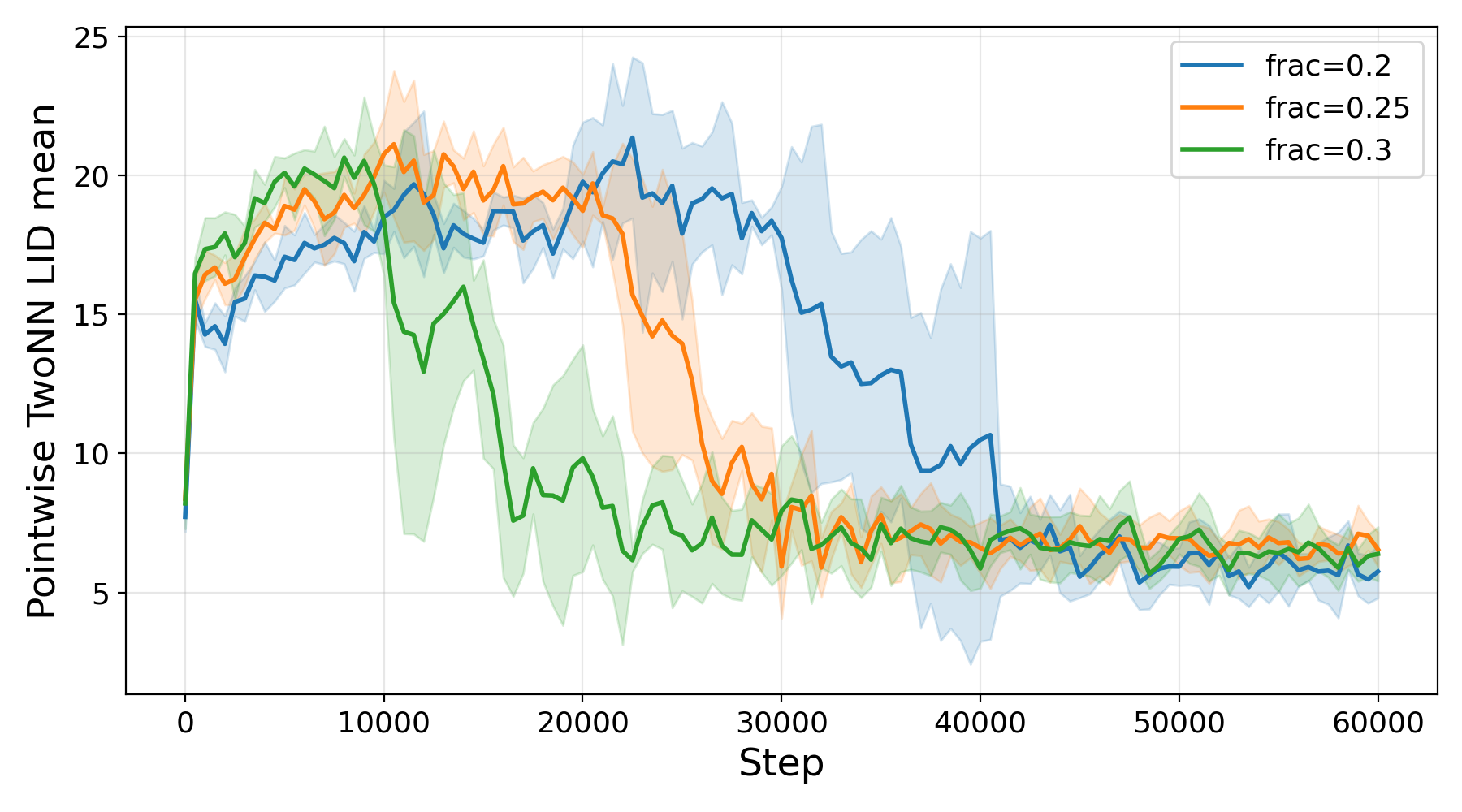}\\[4pt]
    \includegraphics[width=0.49\textwidth]{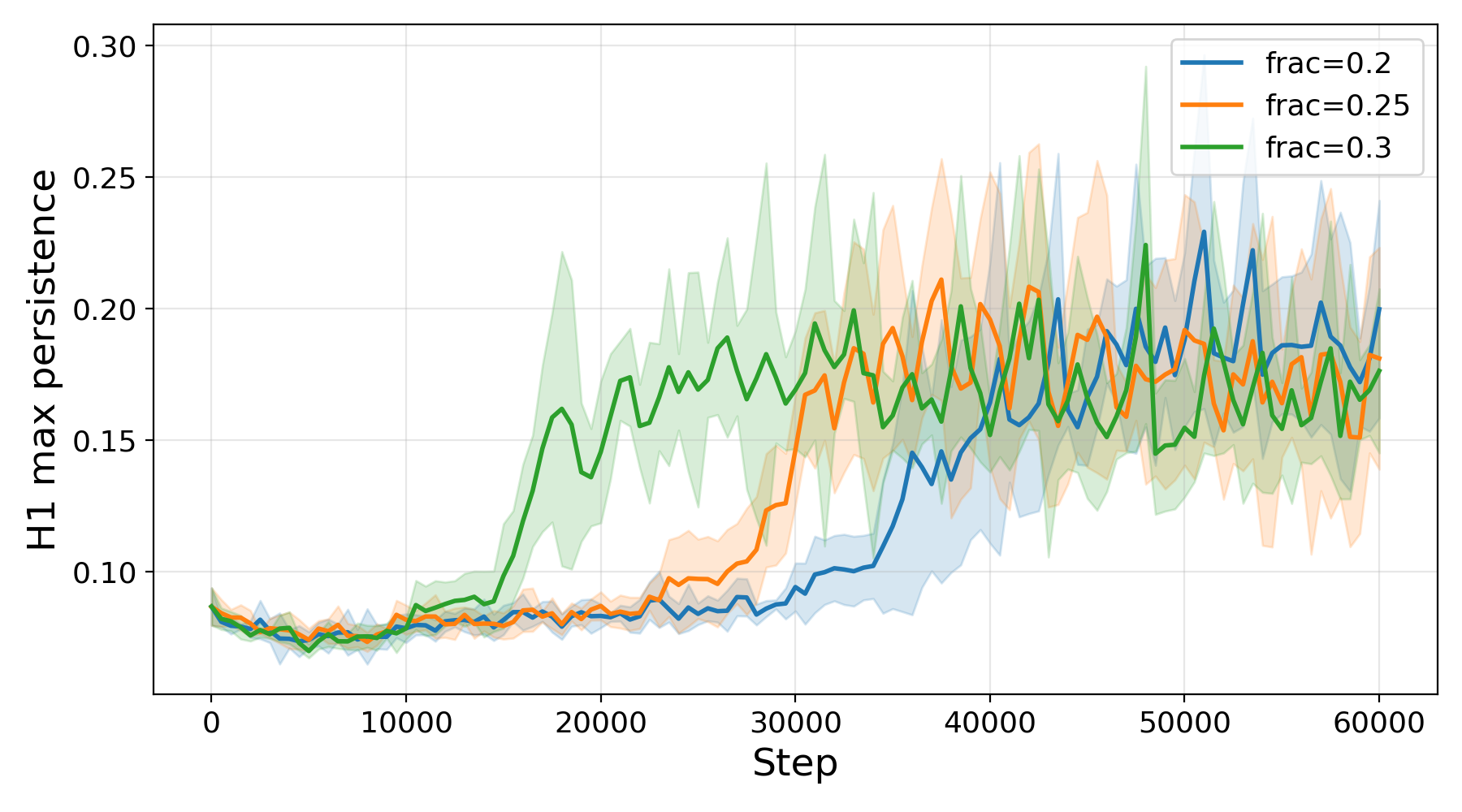}%
    \includegraphics[width=0.49\textwidth]{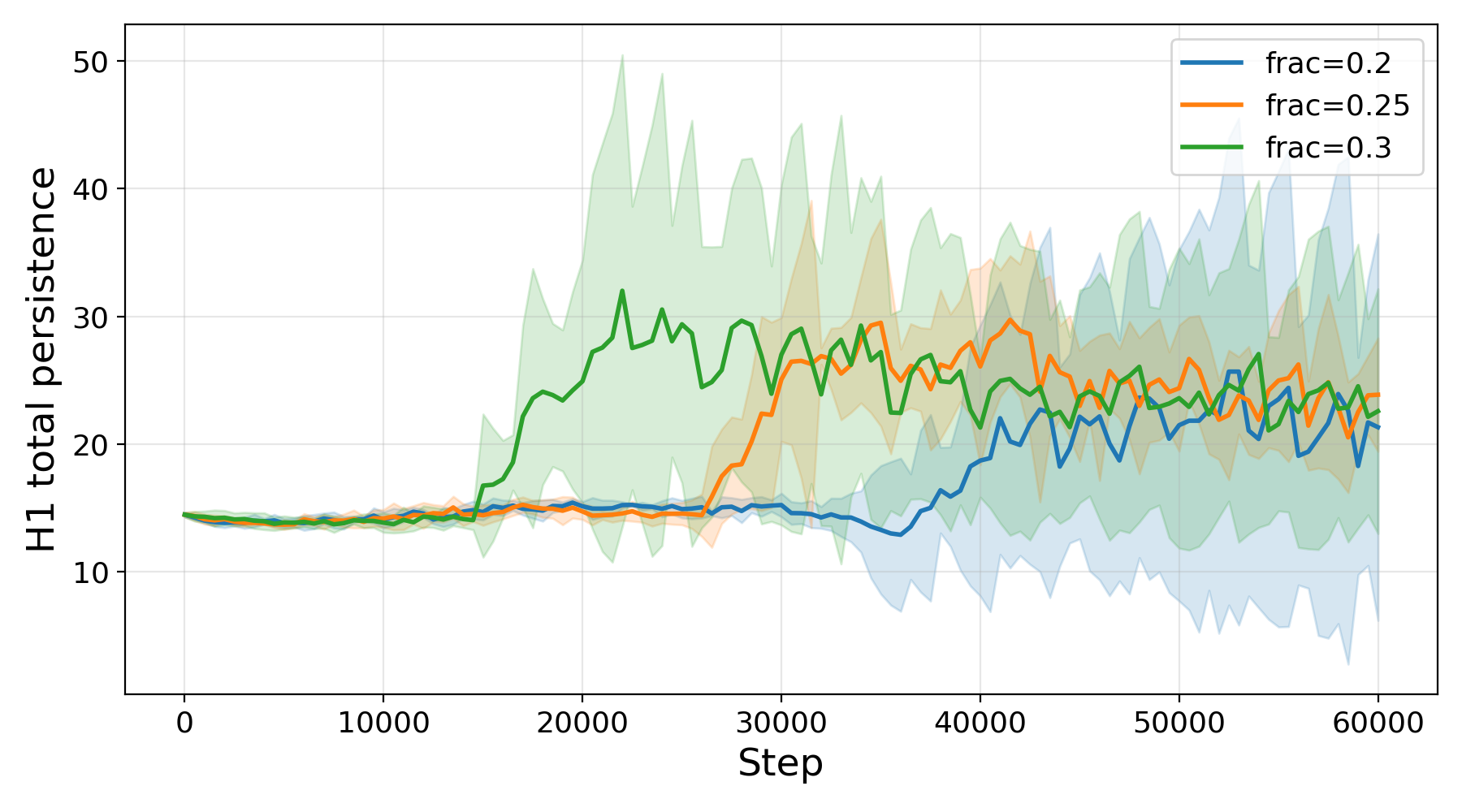}
    \caption{Transformer on modular addition, $p = 149$.  Same layout as Figure~\ref{fig:transformer_197}.}
    \label{fig:transformer_149}
\end{figure}

Figures~\ref{fig:transformer_113} and~\ref{fig:transformer_149} show the full transformer results for $p = 113$ and $p = 149$, using the same layout as Figure~\ref{fig:transformer_197} in the main text.  The topological signature and LID inversion are reproduced in both cases across all training fractions.

\section{Additional MLP Results}
\label{app:mlp_primes}

\begin{figure}[h]
    \centering
    \includegraphics[width=0.6\textwidth]{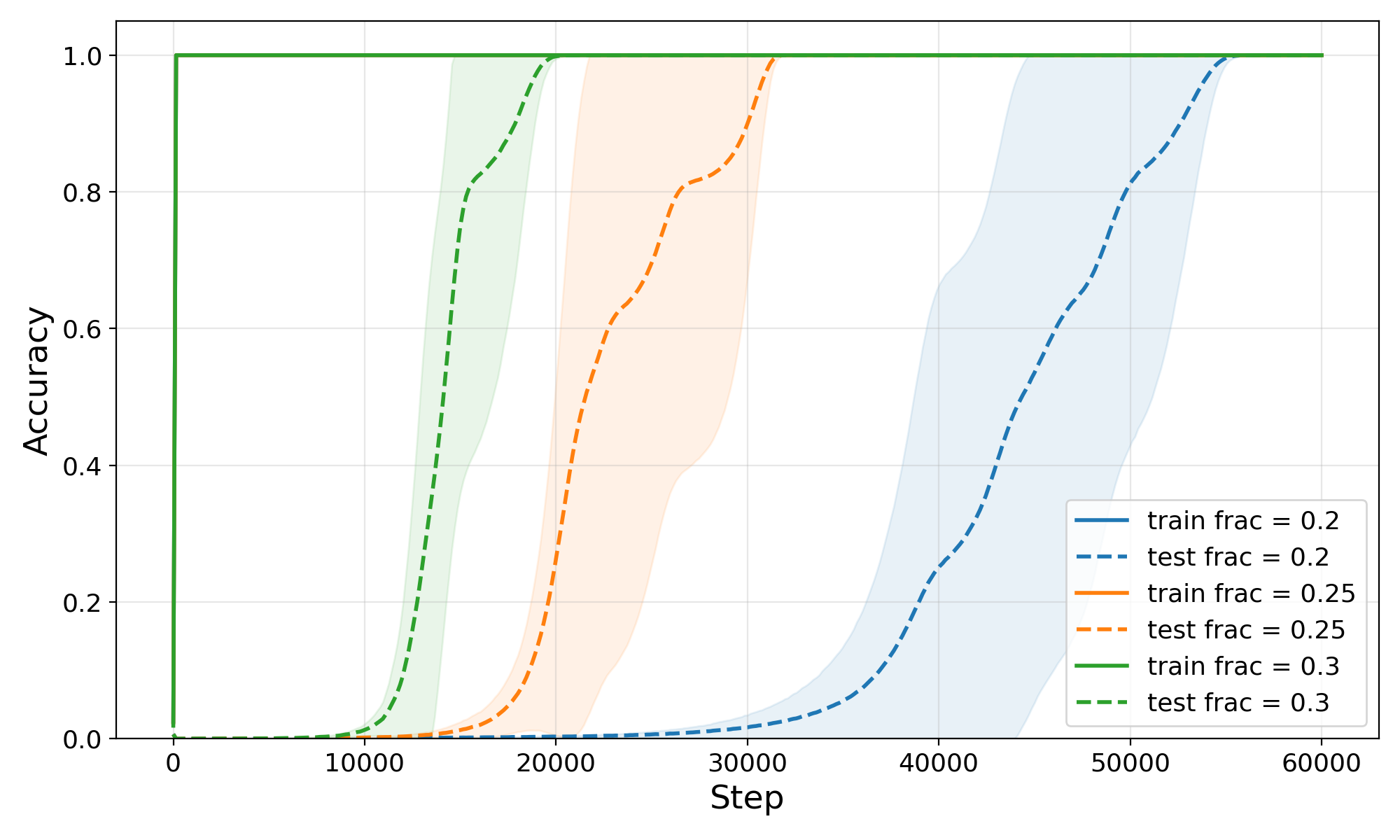}\\[4pt]
    \includegraphics[width=0.49\textwidth]{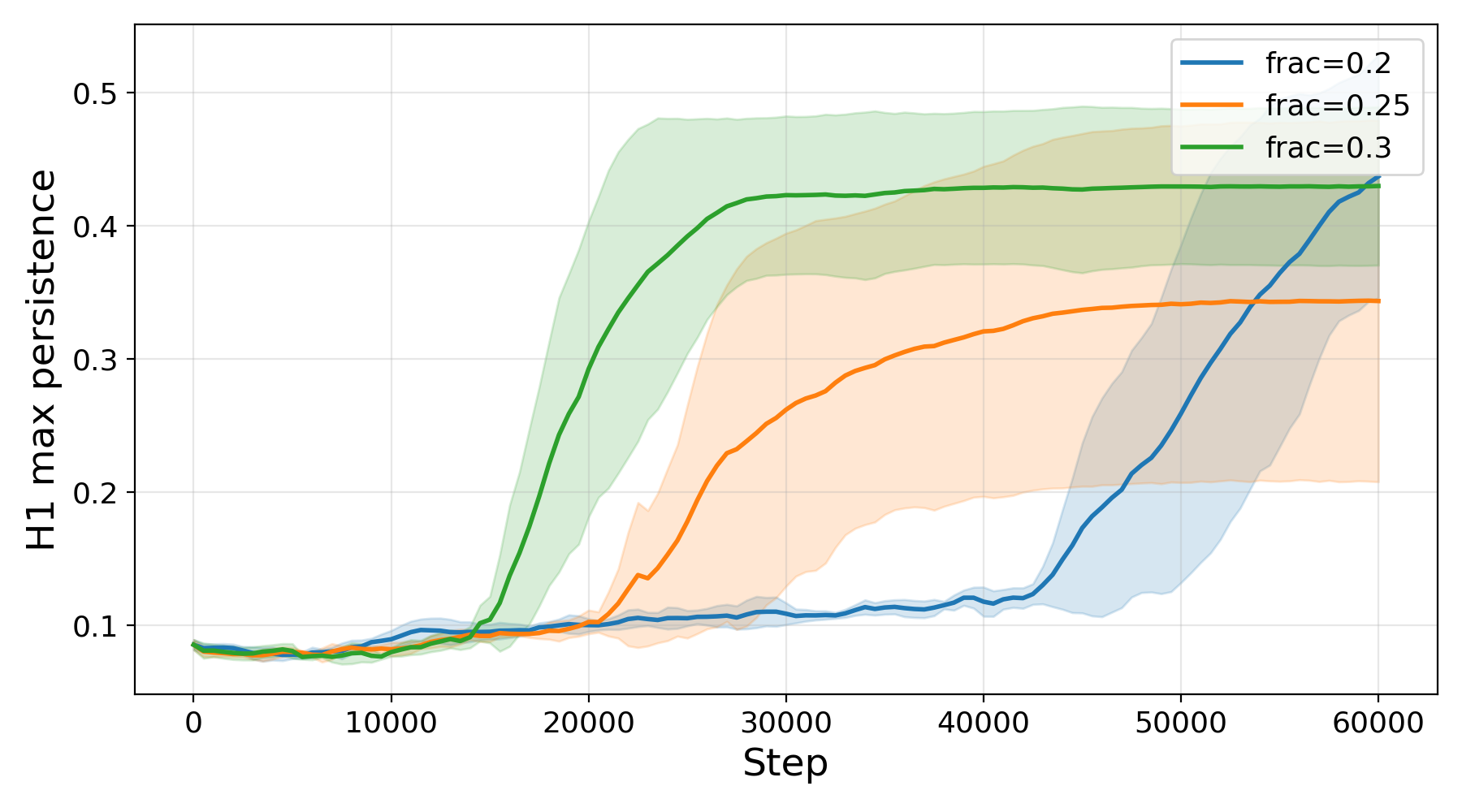}%
    \includegraphics[width=0.49\textwidth]{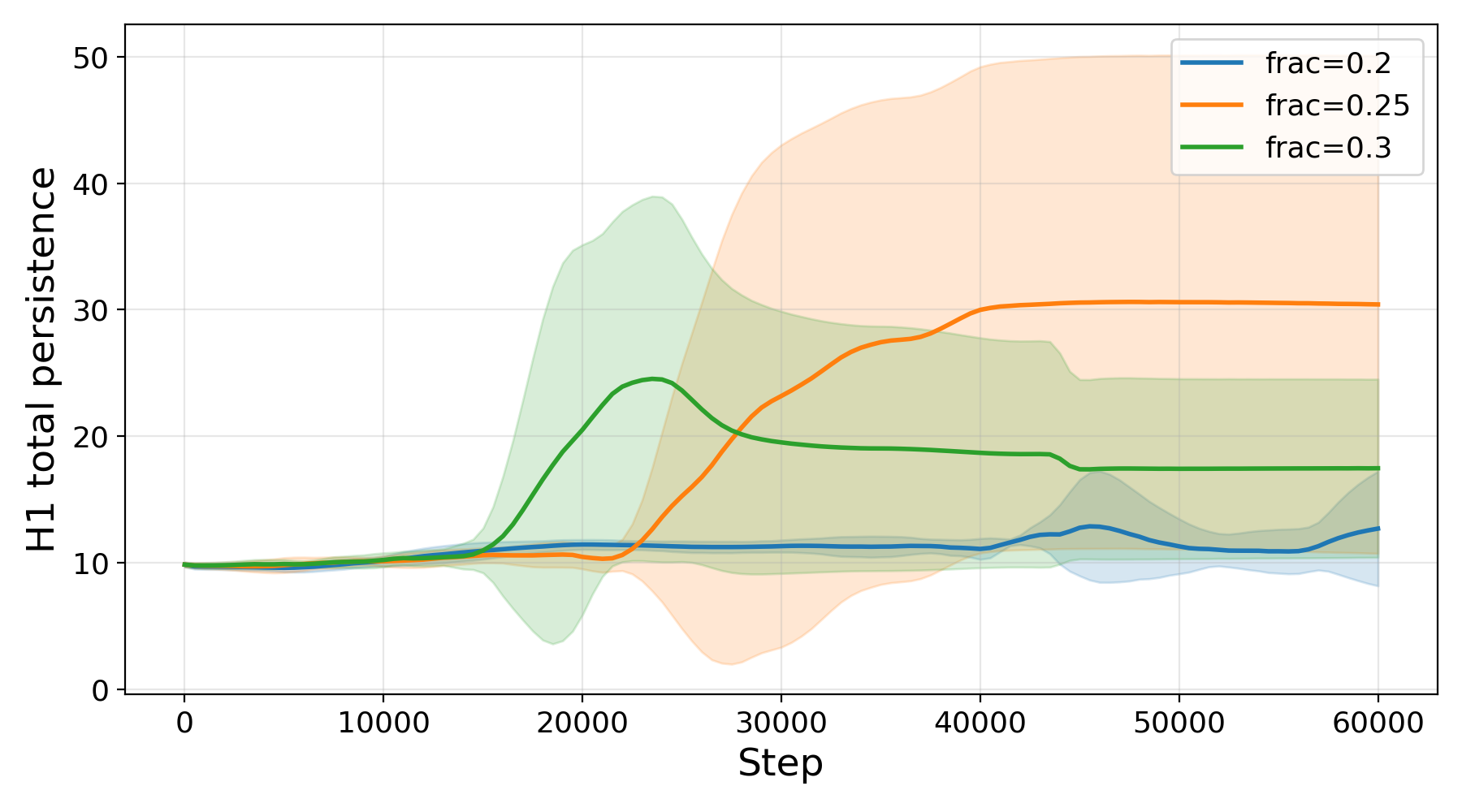}\\[4pt]
    \includegraphics[width=0.49\textwidth]{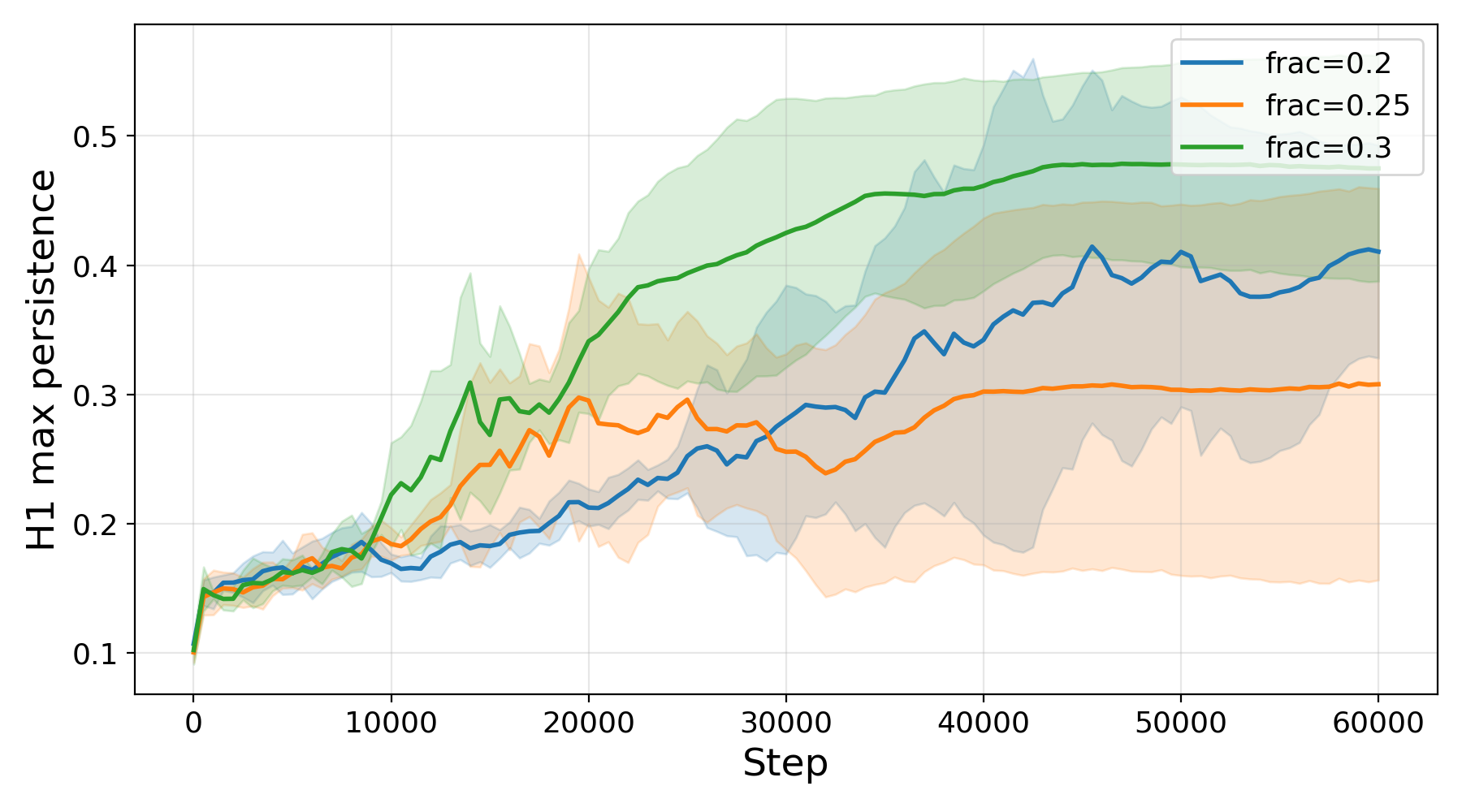}%
    \includegraphics[width=0.49\textwidth]{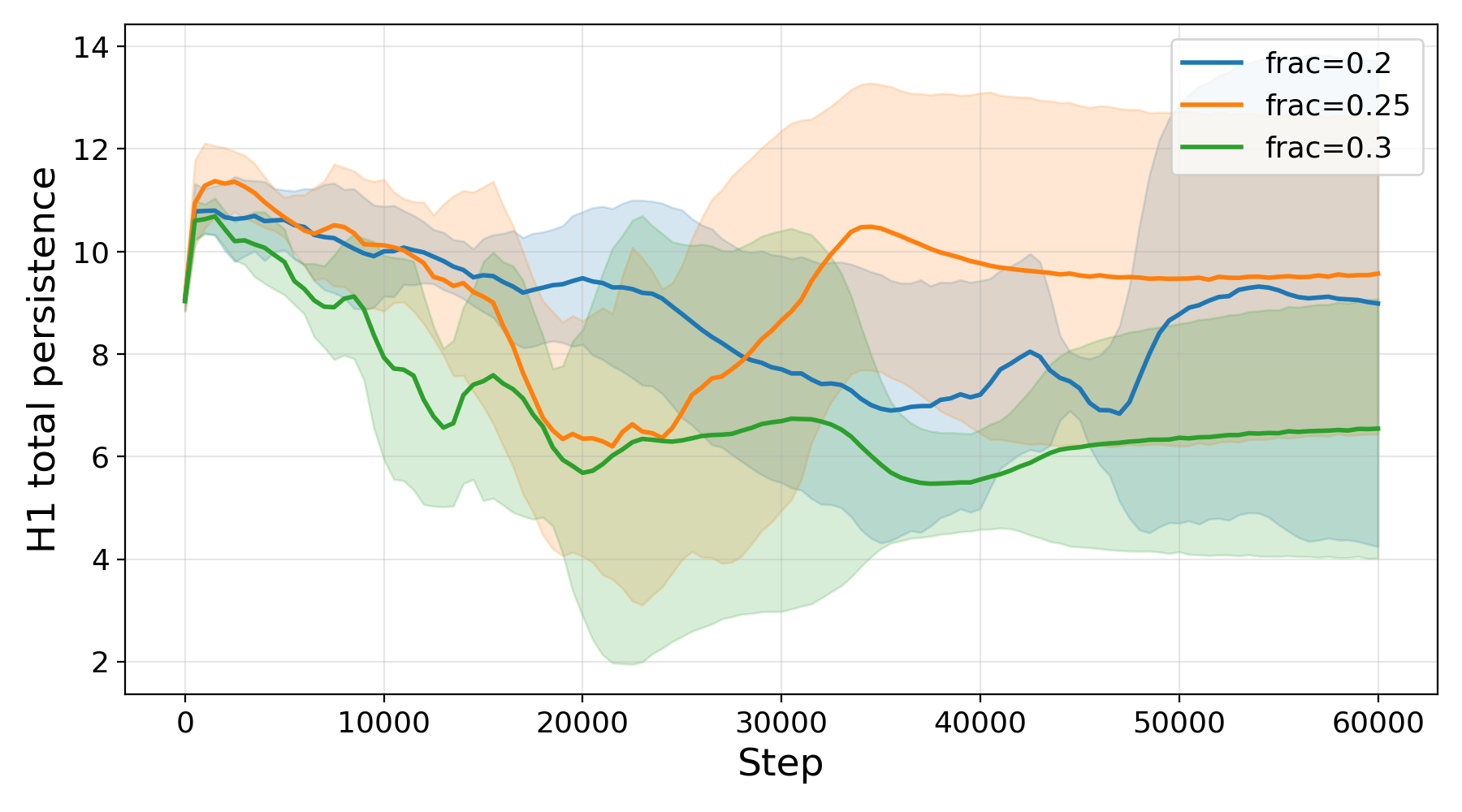}
    \caption{MLP on modular addition, $p = 113$.  Same layout as Figure~\ref{fig:mlp_197}.  H1 total at the embedding layer rises for $\alpha \in \{0.25, 0.3\}$ but only modestly for $\alpha = 0.2$.  H1 total at layer~3 decreases, showing the inversion pattern.}
    \label{fig:mlp_113}
\end{figure}

\begin{figure}[h]
    \centering
    \includegraphics[width=0.6\textwidth]{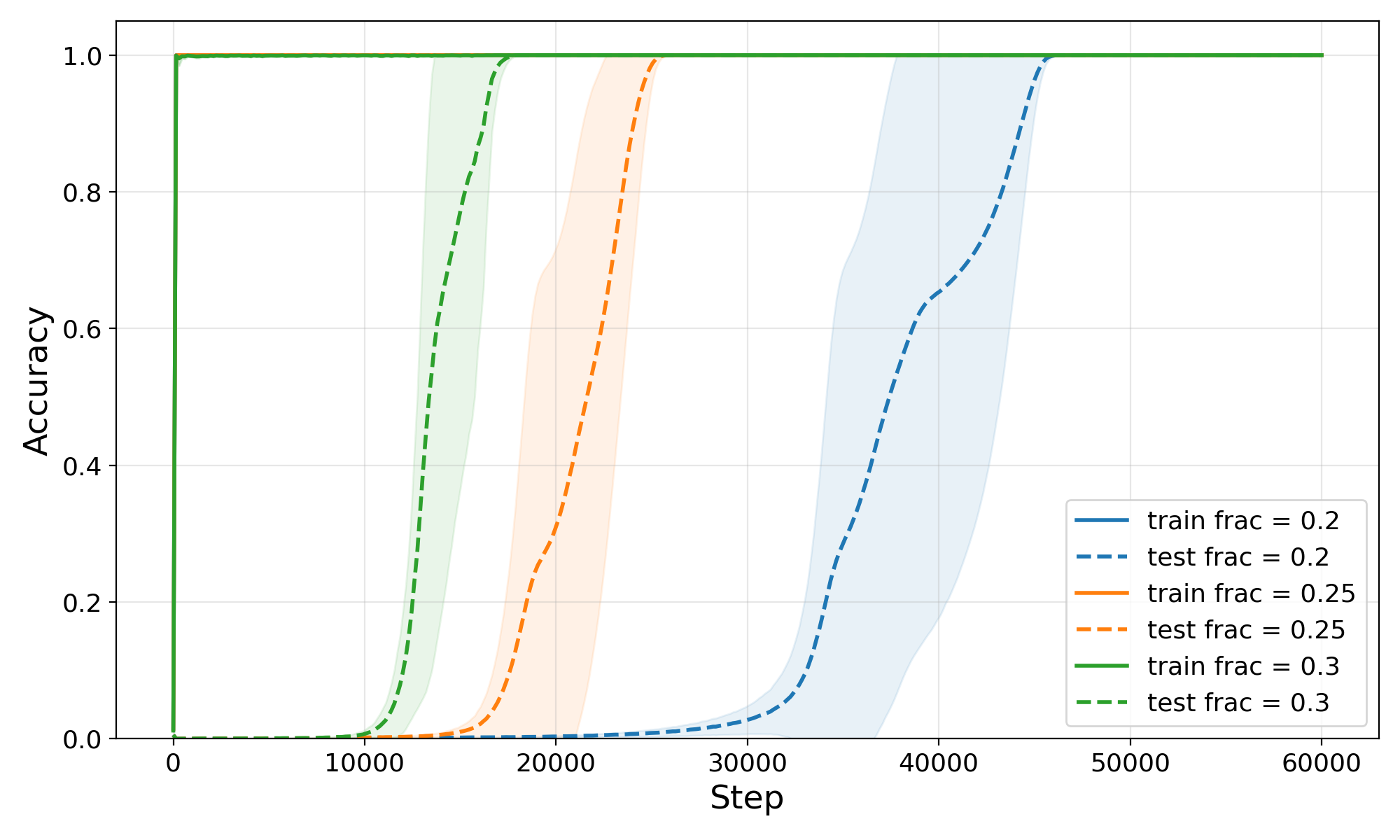}\\[4pt]
    \includegraphics[width=0.49\textwidth]{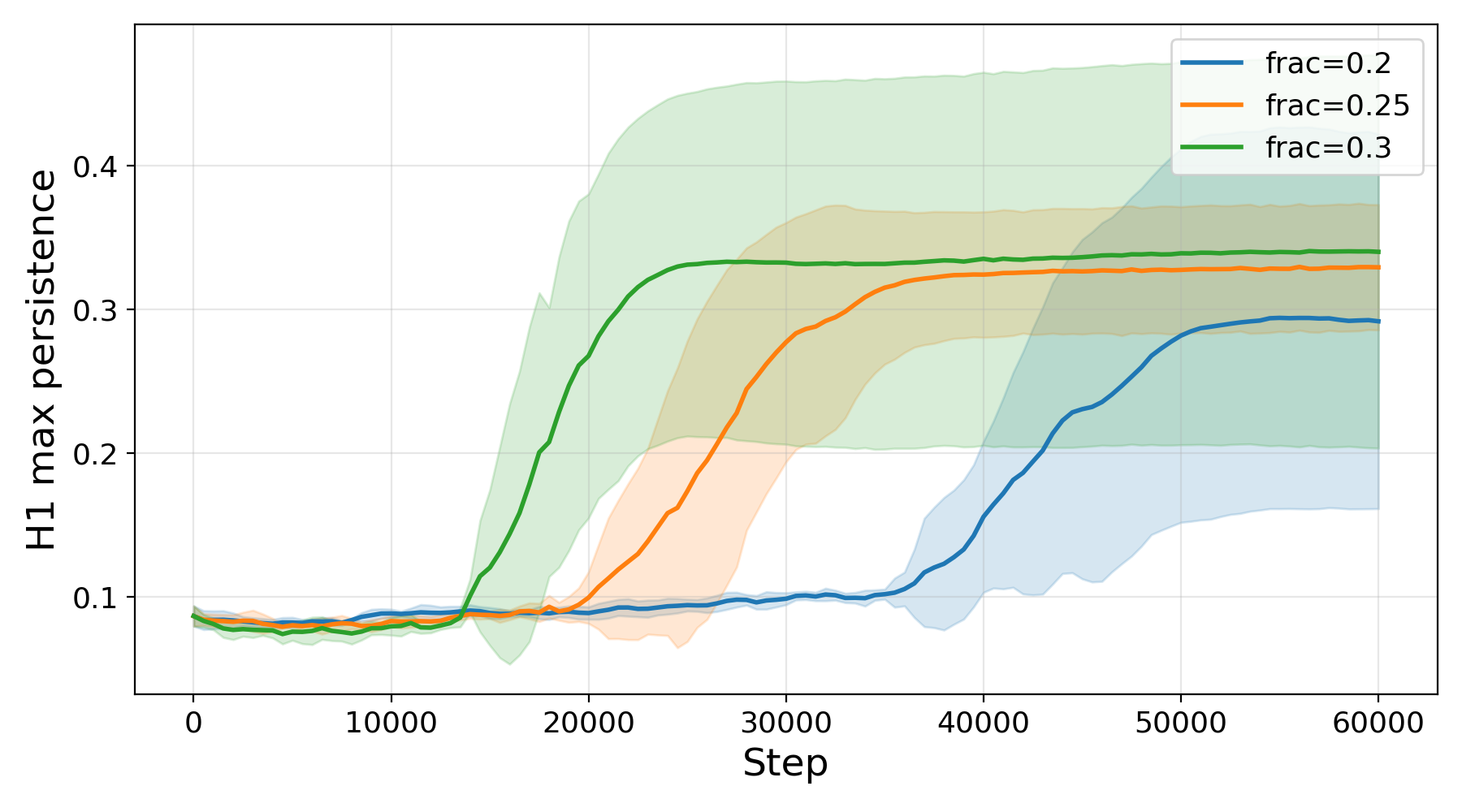}%
    \includegraphics[width=0.49\textwidth]{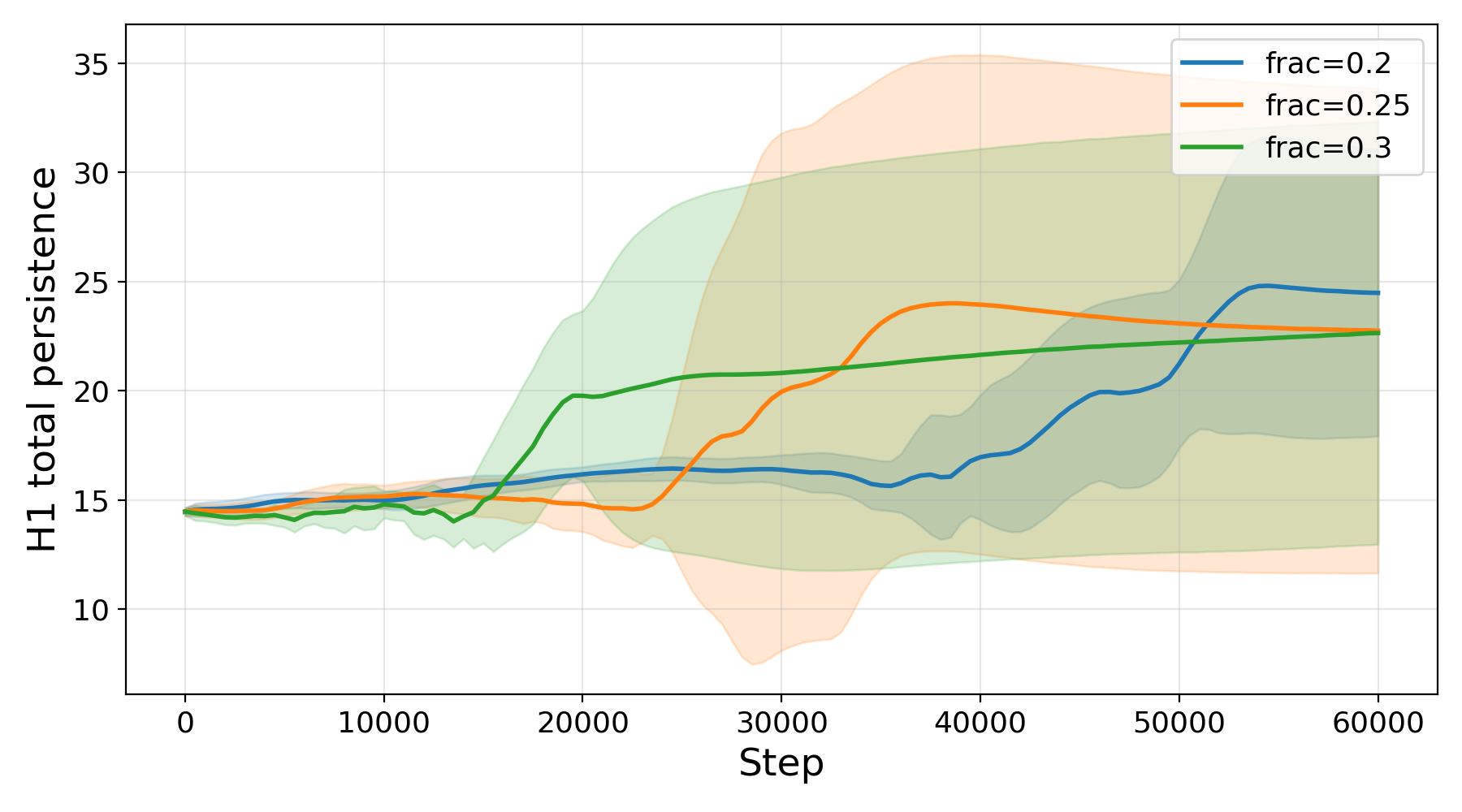}\\[4pt]
    \includegraphics[width=0.49\textwidth]{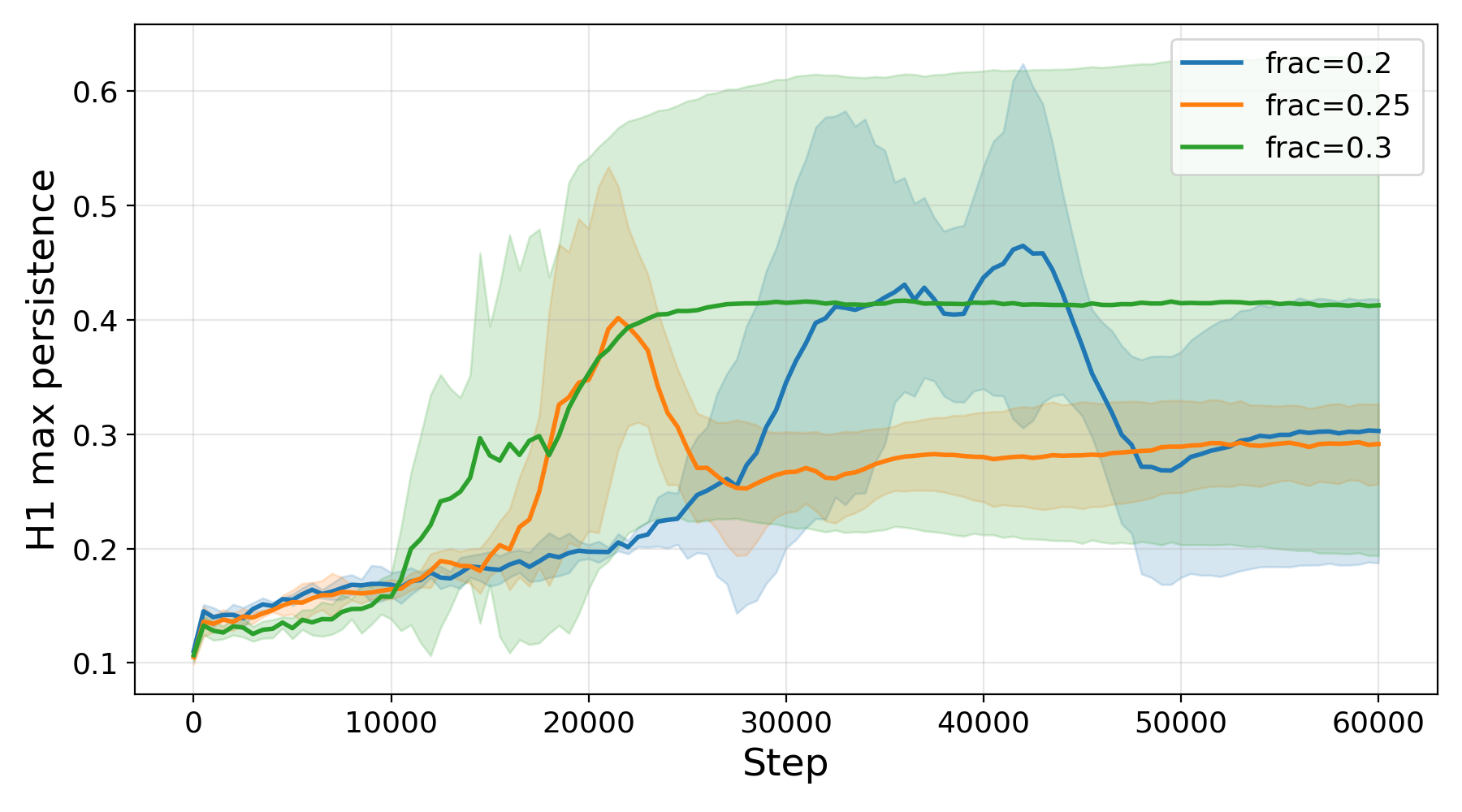}%
    \includegraphics[width=0.49\textwidth]{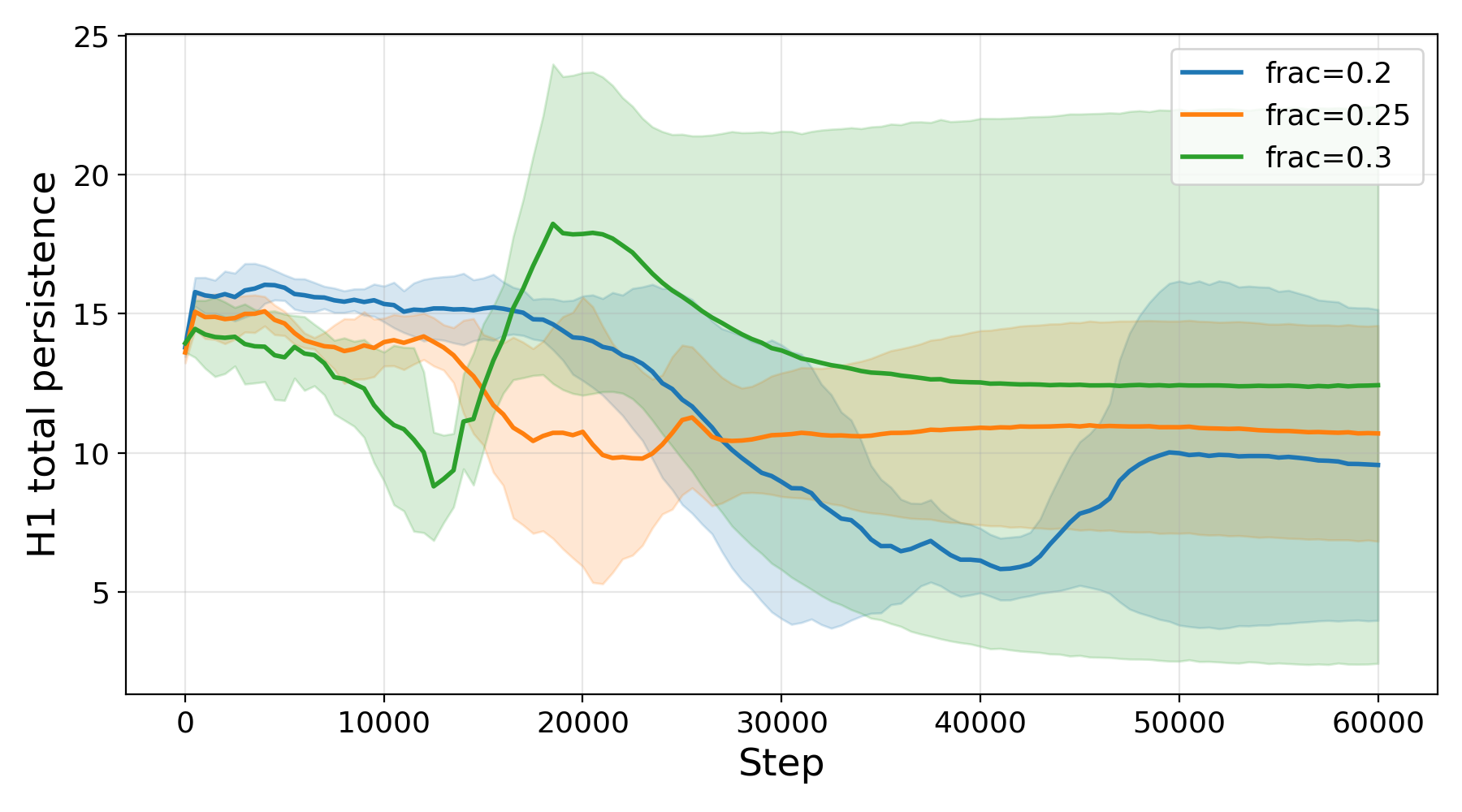}
    \caption{MLP on modular addition, $p = 149$.  Same layout as Figure~\ref{fig:mlp_197}.  All three fractions show a clear rise in both H1 metrics at the embedding layer.  The layer-3 inversion is the most pronounced of the three primes: H1 total drops from $\approx 15$ to $8$--$10$ while H1 max rises.}
    \label{fig:mlp_149}
\end{figure}

Figures~\ref{fig:mlp_113} and~\ref{fig:mlp_149} show the full MLP results for $p = 113$ and $p = 149$, using the same layout as Figure~\ref{fig:mlp_197} in the main text.  The layer-3 inversion (H1 max rising while H1 total falls) is most clearly visible here, particularly for $p = 149$.  For $p = 113$, H1 total at the embedding layer rises clearly for $\alpha \in \{0.25, 0.3\}$ but only modestly for $\alpha = 0.2$.

\section{Additional Ablation Results}
\label{app:ablations}

\subsection{Transformer Ablation Results}
\label{app:transformer_ablations}

\begin{figure}[htbp]
\centering

    \begin{subfigure}[b]{0.8\textwidth}
        \centering
        \includegraphics[width=\linewidth]{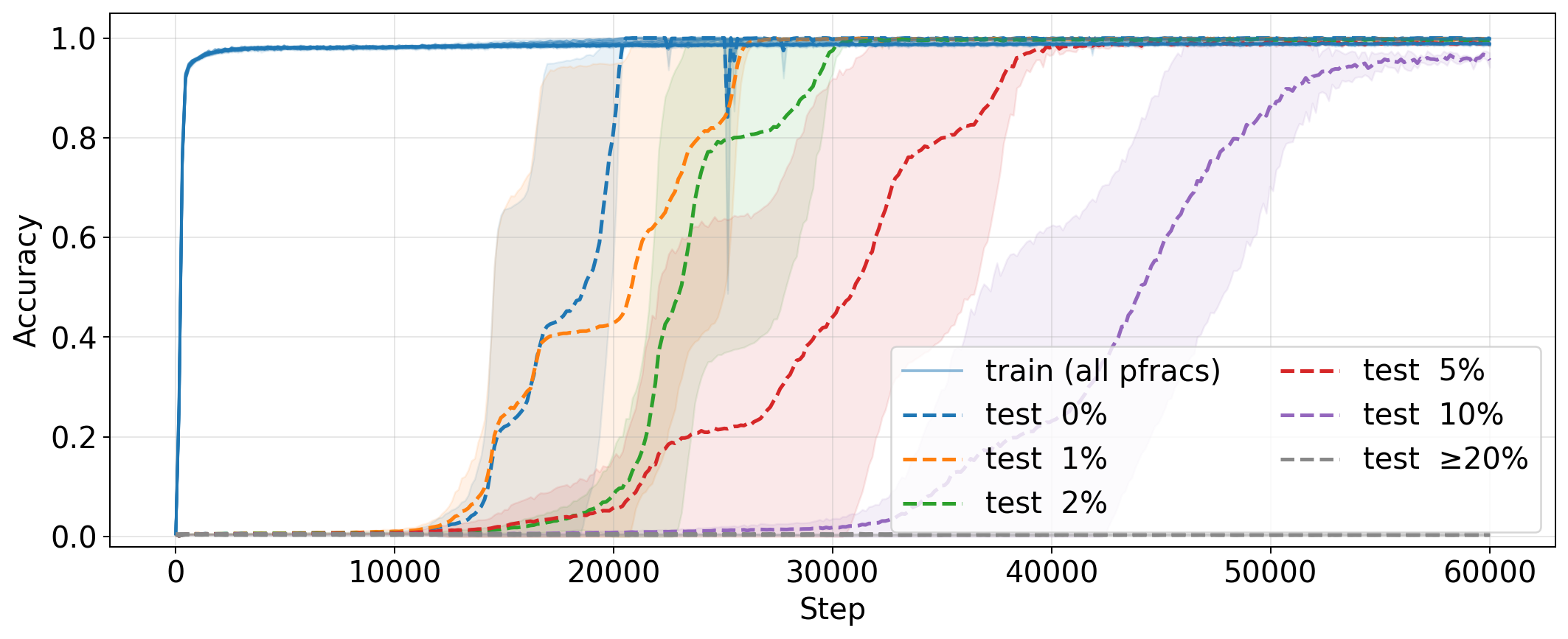}
    \end{subfigure}

    \vspace{0.5em}
    \begin{subfigure}[b]{0.49\textwidth}
        \centering
        \includegraphics[width=\linewidth]{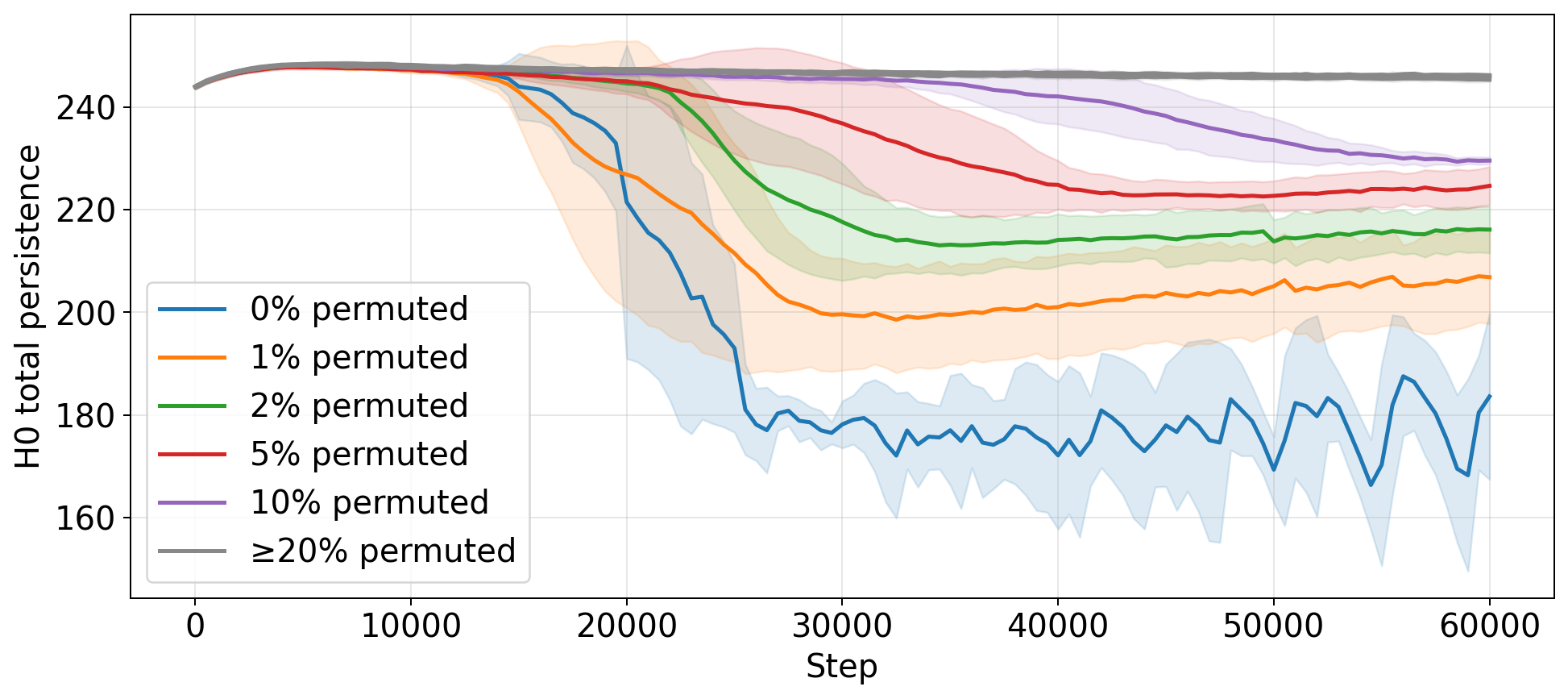}
    \end{subfigure}
    \hfill
    \begin{subfigure}[b]{0.49\textwidth}
        \centering
        \includegraphics[width=\linewidth]{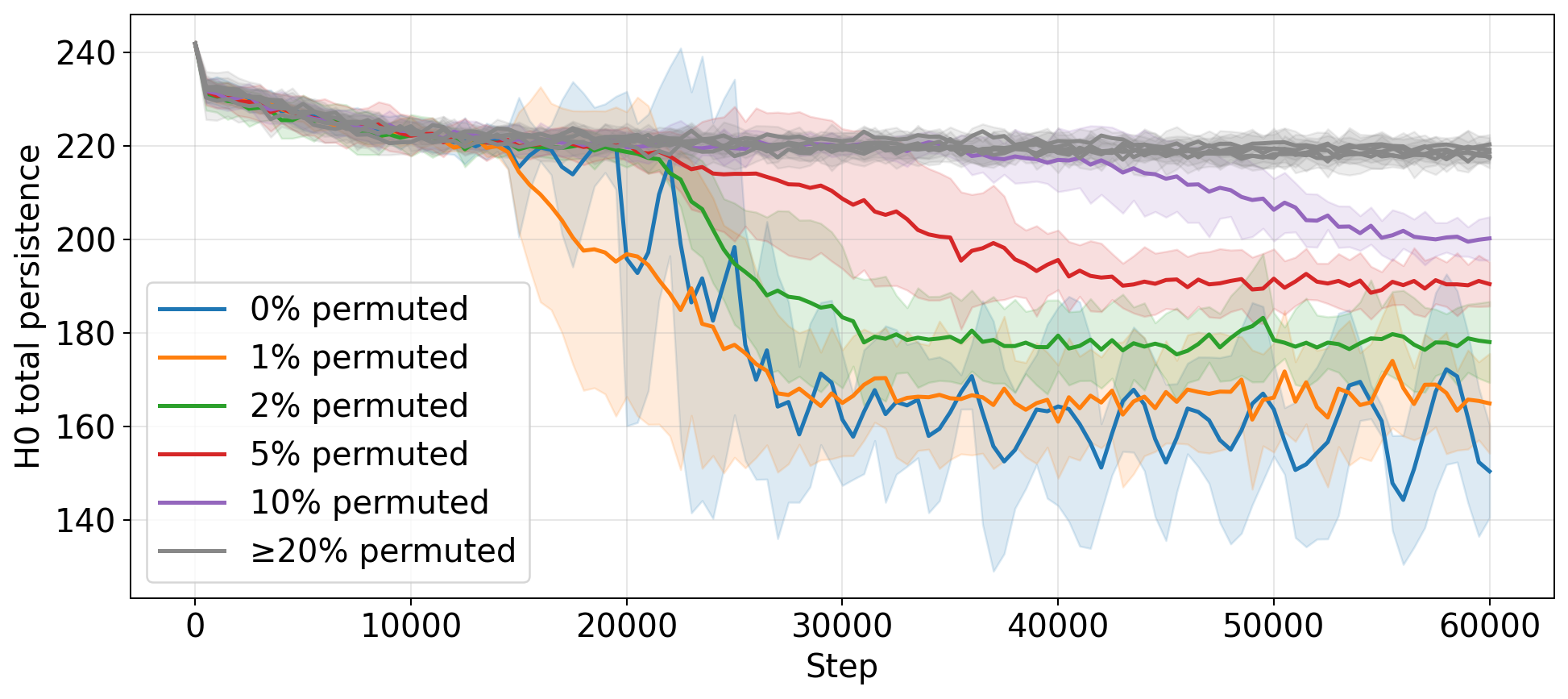}
    \end{subfigure}

    \vspace{0.5em}

    \begin{subfigure}[b]{0.49\textwidth}
        \centering
        \includegraphics[width=\linewidth]{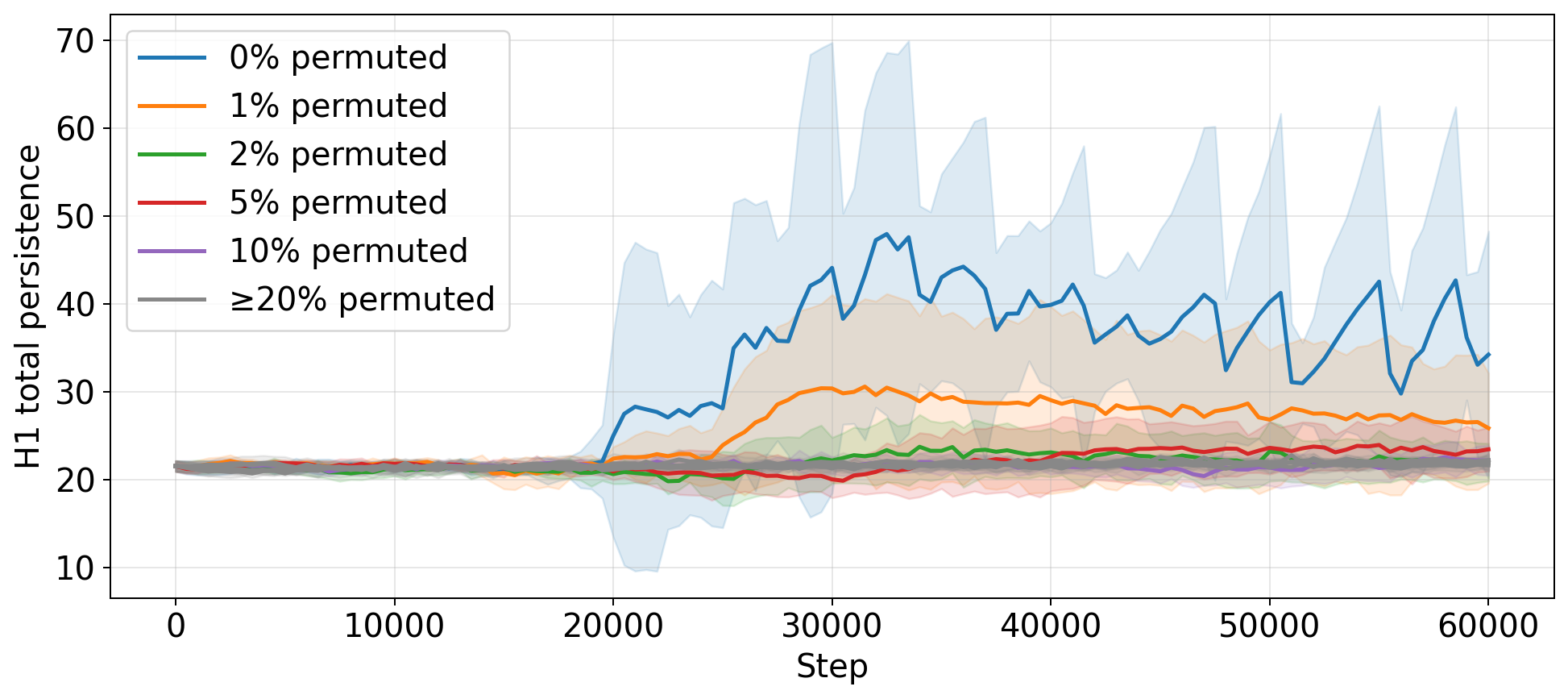}
    \end{subfigure}
    \hfill
    \begin{subfigure}[b]{0.49\textwidth}
        \centering
        \includegraphics[width=\linewidth]{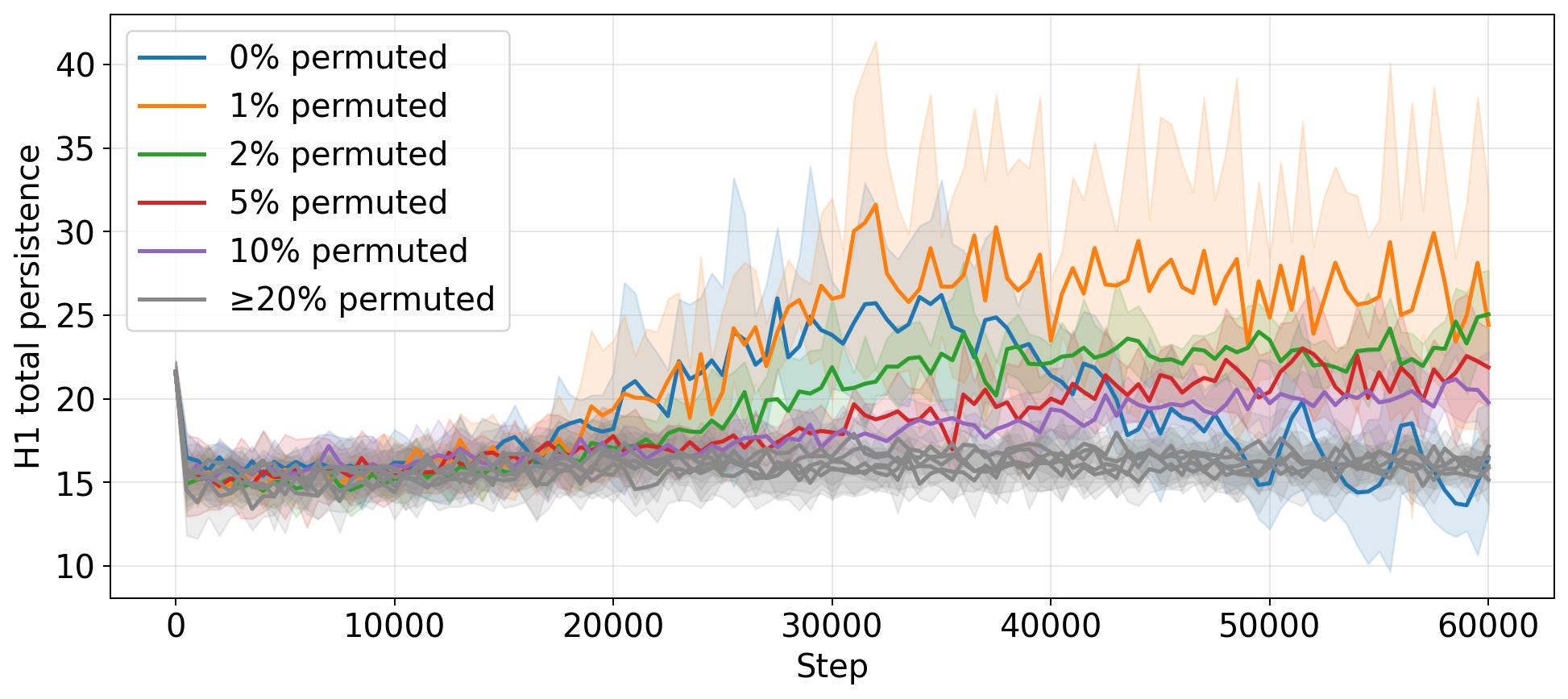}
    \end{subfigure}

    \caption{Training dynamics under label permutation for the Transformer model. 
    The top panel shows test accuracy across training for different label permutation fractions $P_{\mathrm{frac}}$. 
    The remaining panels show the corresponding Vietoris--Rips PH statistics computed from the learned representations, with the left and right columns corresponding to the embedding layer (layer 0) and first layer (layer 1), respectively. 
    Rows 2 and 3 report $H_0$ total persistence, and $H_1$ total persistence respectively.
    For low permutation levels, successful grokking is accompanied by a pronounced decrease in $H_0$ persistence and the emergence of a dominant $H_1$ cycle, consistent with the formation of a structured periodic representation. 
    As label corruption increases, both generalization and the associated topological transition progressively deteriorate.}
    \label{fig:transformer_ablation_sup}
\end{figure}

\begin{figure}[htbp]
    \centering

    \begin{subfigure}[b]{0.48\linewidth}
        \centering
        \includegraphics[width=\linewidth]{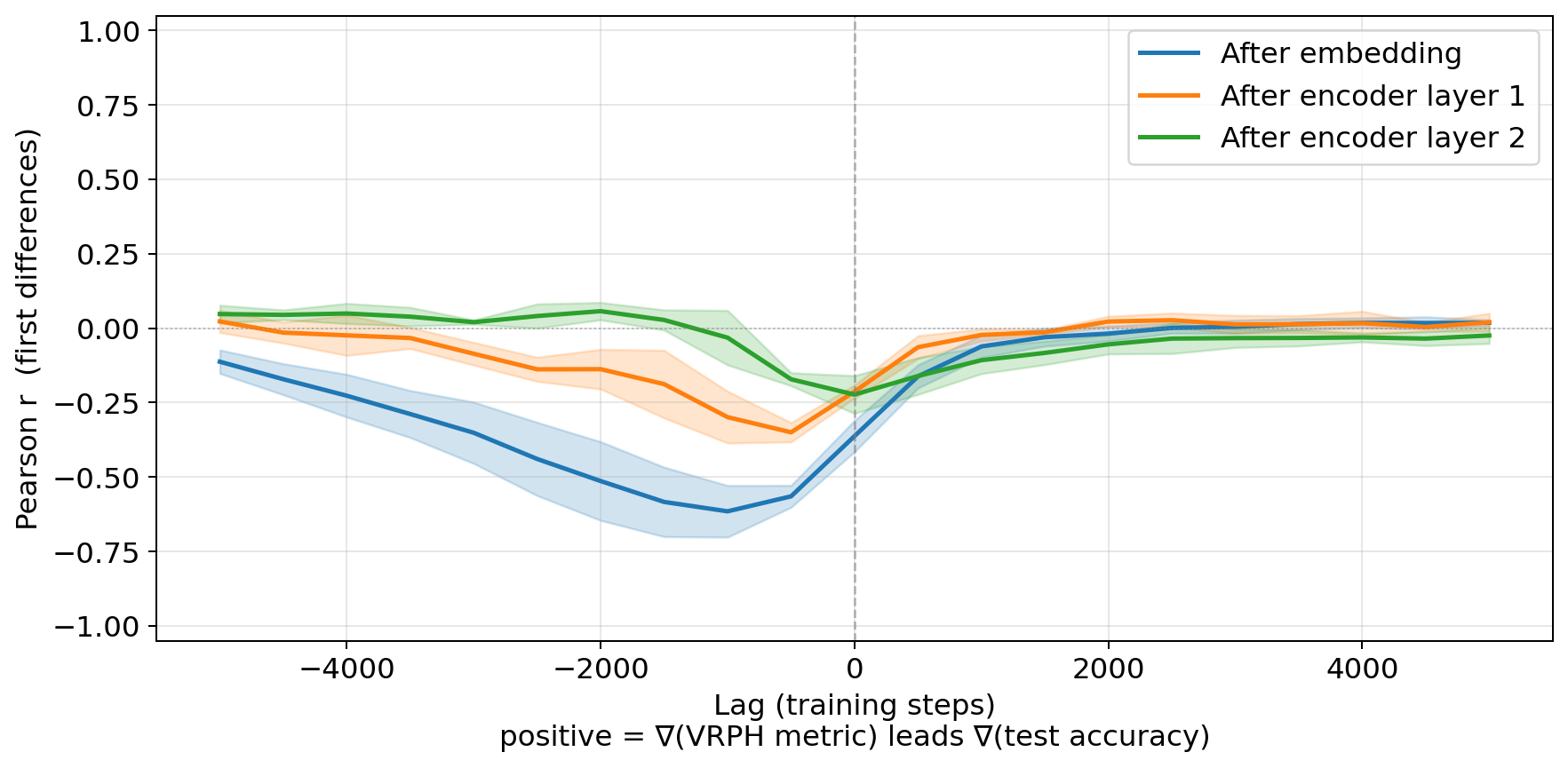}
        \caption{$P_{\mathrm{frac}}=1\%$}
    \end{subfigure}
    \hfill
    \begin{subfigure}[b]{0.48\linewidth}
        \centering
        \includegraphics[width=\linewidth]{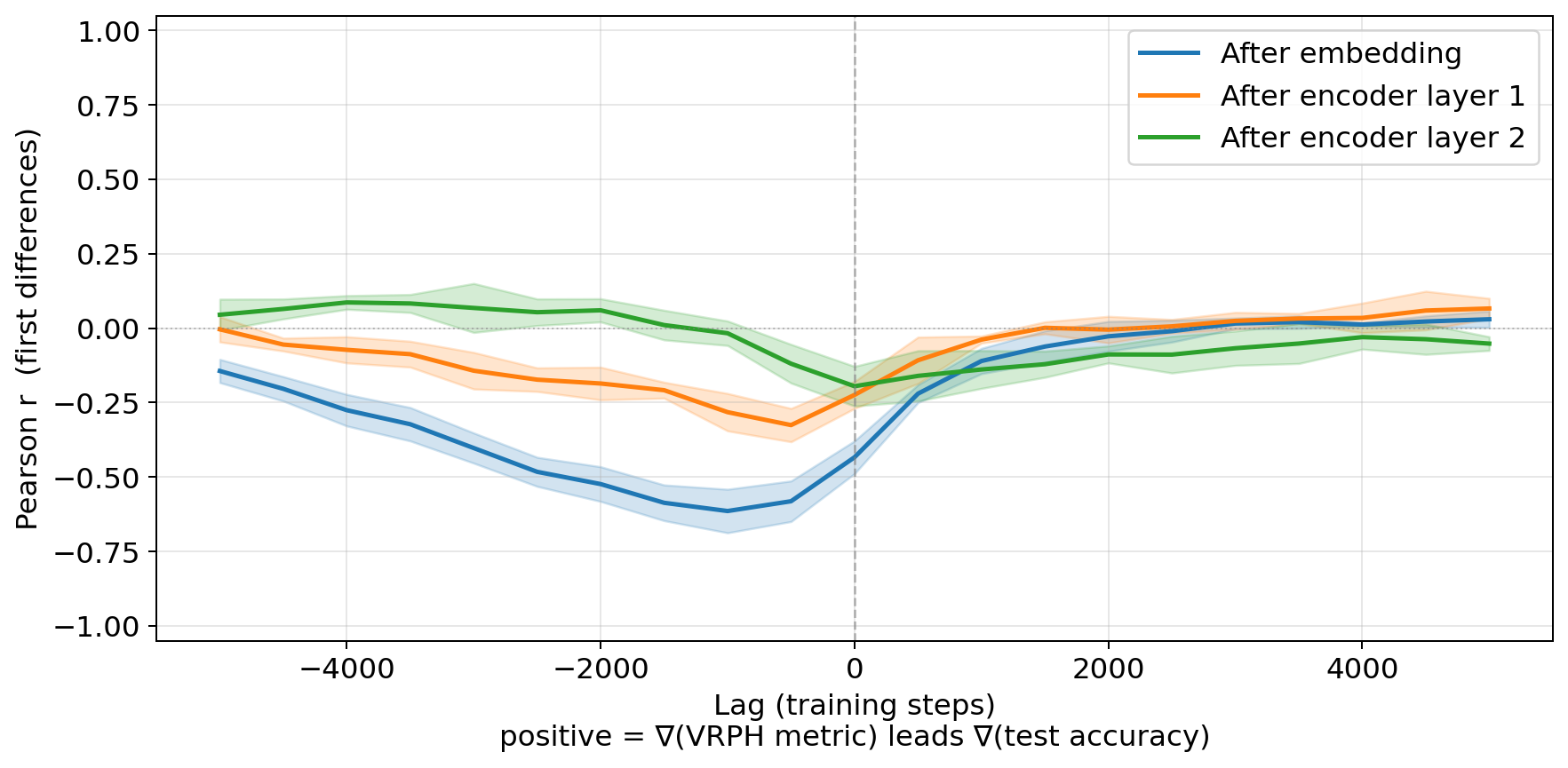}
        \caption{$P_{\mathrm{frac}}=2\%$}
    \end{subfigure}

    \vspace{0.5em}

    \begin{subfigure}[b]{0.48\linewidth}
        \centering
        \includegraphics[width=\linewidth]{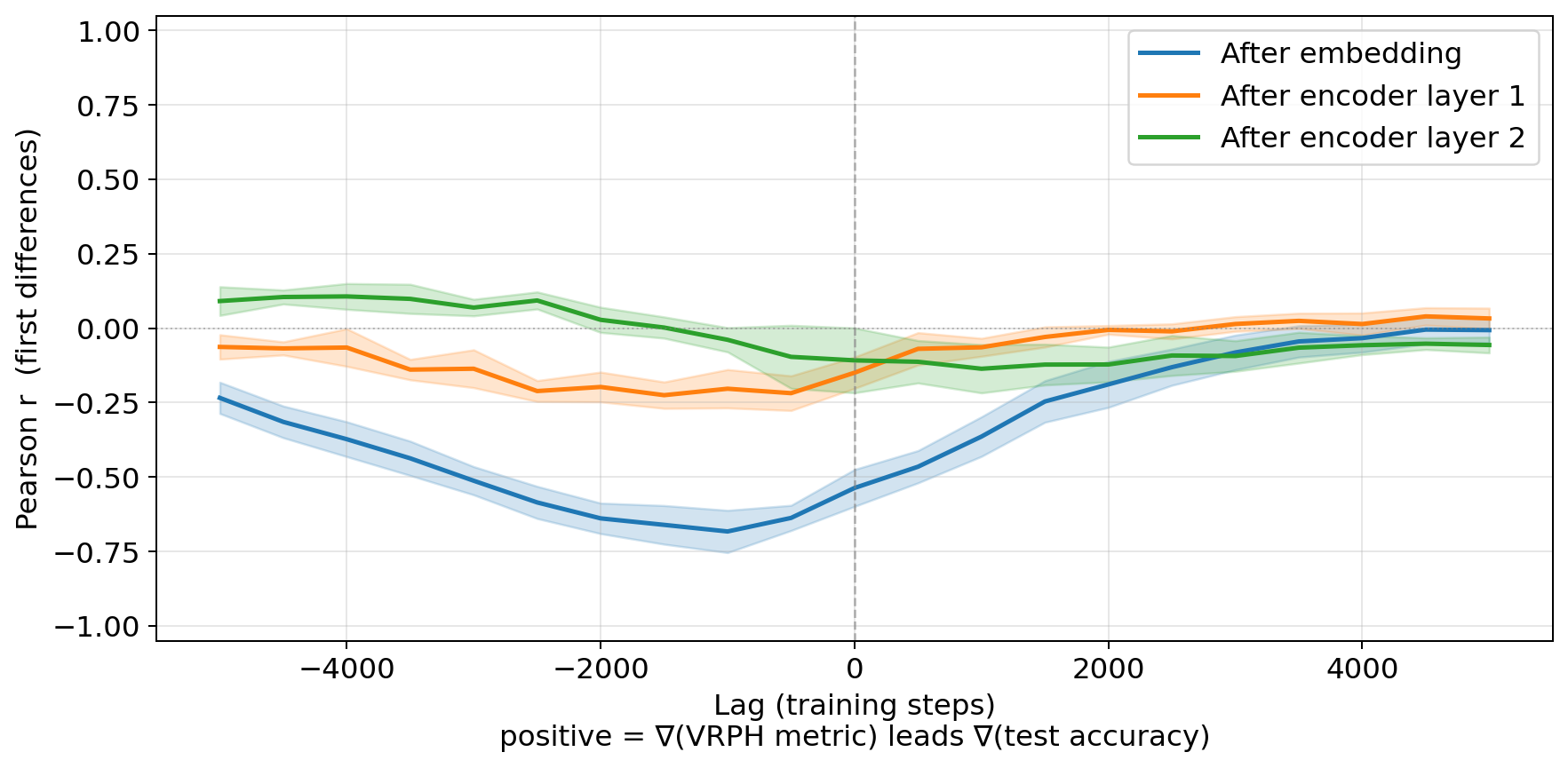}
        \caption{$P_{\mathrm{frac}}=5\%$}
    \end{subfigure}
    \hfill
    \begin{subfigure}[b]{0.48\linewidth}
        \centering
        \includegraphics[width=\linewidth]{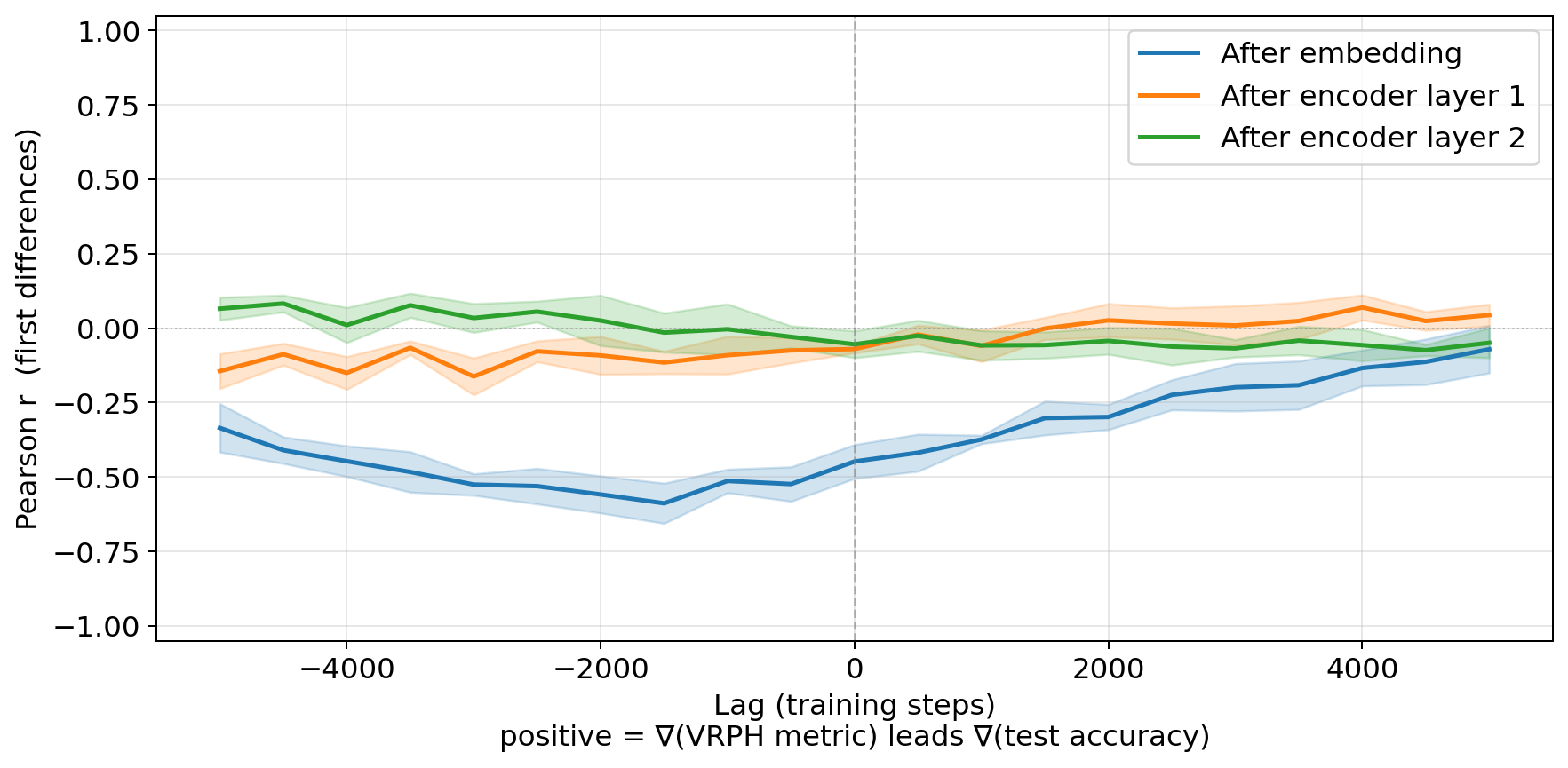}
        \caption{$P_{\mathrm{frac}}=10\%$}
    \end{subfigure}

    \caption{
    Cross-correlation functions (CCFs) between the first-order differences of test accuracy and total $H_0$ persistence for the Transformer model under varying levels of label permutation. Across all grokking regimes ($P_{\mathrm{frac}} \in \{1\%, 2\%, 5\%, 10\%\}$), the peak correlation consistently occurs at a negative lag, indicating that improvements in predictive performance precede the corresponding topological restructuring of the learned representation space.
    }
    \label{fig:ccf}
\end{figure}

In this section, we provide additional cross-correlation results and ablation analyses to supplement the analysis presented in the main text. Table \ref{tab:ph_correlations_transformer} shows a comprehensive summary of Spearman correlations between PH measures and test accuracy across all layers and permutation levels for the Transformer model. This table provides a detailed breakdown of the relationships discussed in the main text. Figure \ref{fig:transformer_ablation_sup} provides a further illustration of additional topological statistics across the training dynamics for different levels of permutation at layers 0 and 1. Figure \ref{fig:ccf} shows the relationship between the first order difference for the test accuracy and the total $H_0$ persistence under varying levels of label permutation for the Transformer model. These results supplement the main finding that improvements in generalization consistently precede the corresponding topological transition.

\begin{table}[ht]
\centering
\caption{Spearman Rank Correlation ($\rho$) between PH measures and test accuracy (mean $\pm$ SD). Bolded values indicate $p < 0.05$.}
\label{tab:ph_correlations_transformer}
\resizebox{\textwidth}{!}{%
\begin{tabular}{llccccc}
\toprule
\textbf{Metric} & \textbf{Layer} & \textbf{0\%} & \textbf{1\%} & \textbf{5\%} & \textbf{10\%} & \textbf{20\%} \\ 
\midrule
$H_0$ Max   & Embed   & \textbf{-0.78 $\pm$ 0.04} & \textbf{-0.72 $\pm$ 0.07} & \textbf{-0.84 $\pm$ 0.07} & \textbf{-0.86 $\pm$ 0.06} & -0.10 $\pm$ 0.32 \\
            & Layer 1 & \textbf{-0.55 $\pm$ 0.08} & \textbf{-0.70 $\pm$ 0.06} & \textbf{-0.89 $\pm$ 0.03} & \textbf{-0.85 $\pm$ 0.07} & -0.05 $\pm$ 0.20 \\
            & Layer 2 & \textbf{-0.47 $\pm$ 0.10} & \textbf{-0.52 $\pm$ 0.14} & \textbf{-0.56 $\pm$ 0.16} & \textbf{-0.67 $\pm$ 0.08} & +0.00 $\pm$ 0.08 \\ 
\addlinespace
$H_0$ Total & Embed   & \textbf{-0.75 $\pm$ 0.03} & \textbf{-0.71 $\pm$ 0.10} & \textbf{-0.87 $\pm$ 0.06} & \textbf{-0.91 $\pm$ 0.03} & -0.20 $\pm$ 0.36 \\
            & Layer 1 & \textbf{-0.49 $\pm$ 0.08} & \textbf{-0.67 $\pm$ 0.09} & \textbf{-0.90 $\pm$ 0.03} & \textbf{-0.88 $\pm$ 0.06} & -0.14 $\pm$ 0.29 \\
            & Layer 2 & -0.06 $\pm$ 0.27          & \textbf{-0.47 $\pm$ 0.14} & \textbf{-0.59 $\pm$ 0.17} & \textbf{-0.82 $\pm$ 0.08} & -0.03 $\pm$ 0.10 \\ 
\addlinespace
$H_1$ Max   & Embed   & \textbf{+0.77 $\pm$ 0.03} & \textbf{+0.71 $\pm$ 0.06} & \textbf{+0.80 $\pm$ 0.06} & \textbf{+0.69 $\pm$ 0.10} & +0.08 $\pm$ 0.10 \\
            & Layer 1 & \textbf{+0.49 $\pm$ 0.08} & \textbf{+0.70 $\pm$ 0.07} & \textbf{+0.81 $\pm$ 0.05} & \textbf{+0.68 $\pm$ 0.14} & +0.09 $\pm$ 0.17 \\
            & Layer 2 & -0.23 $\pm$ 0.23          & \textbf{+0.53 $\pm$ 0.10} & \textbf{+0.59 $\pm$ 0.13} & \textbf{+0.65 $\pm$ 0.11} & +0.05 $\pm$ 0.05 \\ 
\addlinespace
$H_1$ Total & Embed   & \textbf{+0.60 $\pm$ 0.10} & \textbf{+0.42 $\pm$ 0.39} & \textbf{+0.24 $\pm$ 0.49} & \textbf{+0.14 $\pm$ 0.51} & +0.10 $\pm$ 0.21 \\
            & Layer 1 & +0.33 $\pm$ 0.16          & \textbf{+0.71 $\pm$ 0.13} & \textbf{+0.74 $\pm$ 0.12} & \textbf{+0.71 $\pm$ 0.18} & +0.10 $\pm$ 0.18 \\
            & Layer 2 & \textbf{+0.80 $\pm$ 0.04} & \textbf{+0.66 $\pm$ 0.10} & \textbf{+0.62 $\pm$ 0.06} & \textbf{+0.40 $\pm$ 0.31} & +0.05 $\pm$ 0.09 \\ 
\bottomrule
\end{tabular}}
\end{table}

\subsection{MLP Ablation Results}
\label{app:mlp_ablations}

\begin{figure}[htbp]
    \centering

    \begin{subfigure}[b]{0.8\textwidth}
        \centering
        \includegraphics[width=\linewidth]{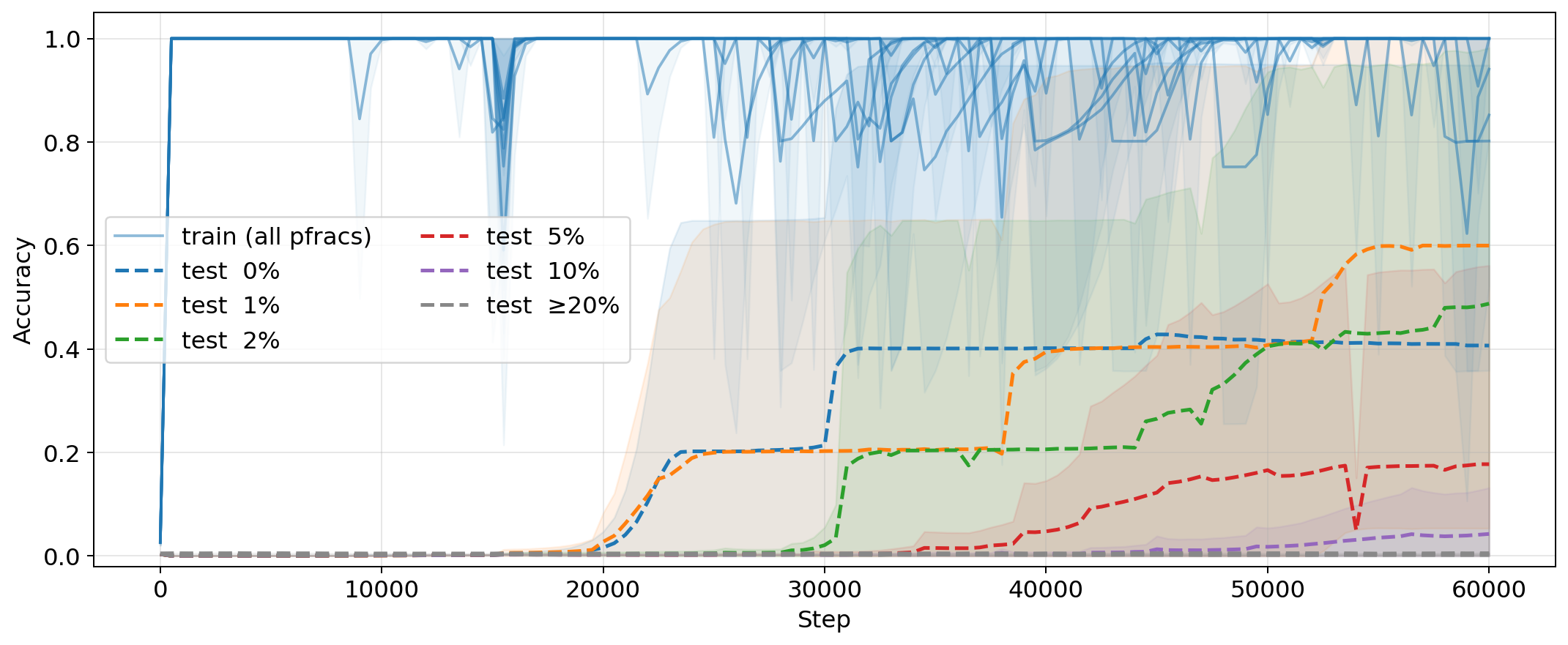}
    \end{subfigure}

    \vspace{0.5em}

    \begin{subfigure}[b]{0.49\textwidth}
        \centering
        \includegraphics[width=\linewidth]{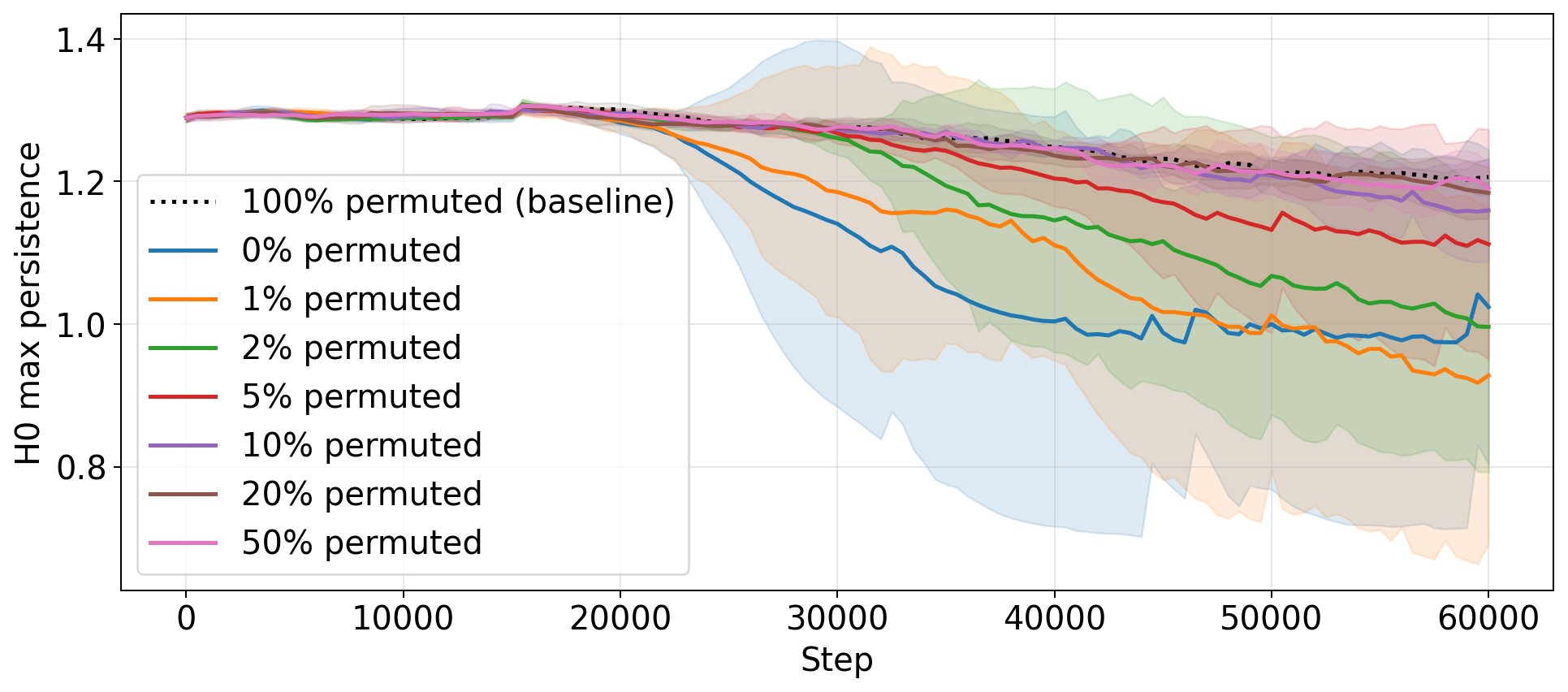}
    \end{subfigure}
    \hfill
    \begin{subfigure}[b]{0.49\textwidth}
        \centering
        \includegraphics[width=\linewidth]{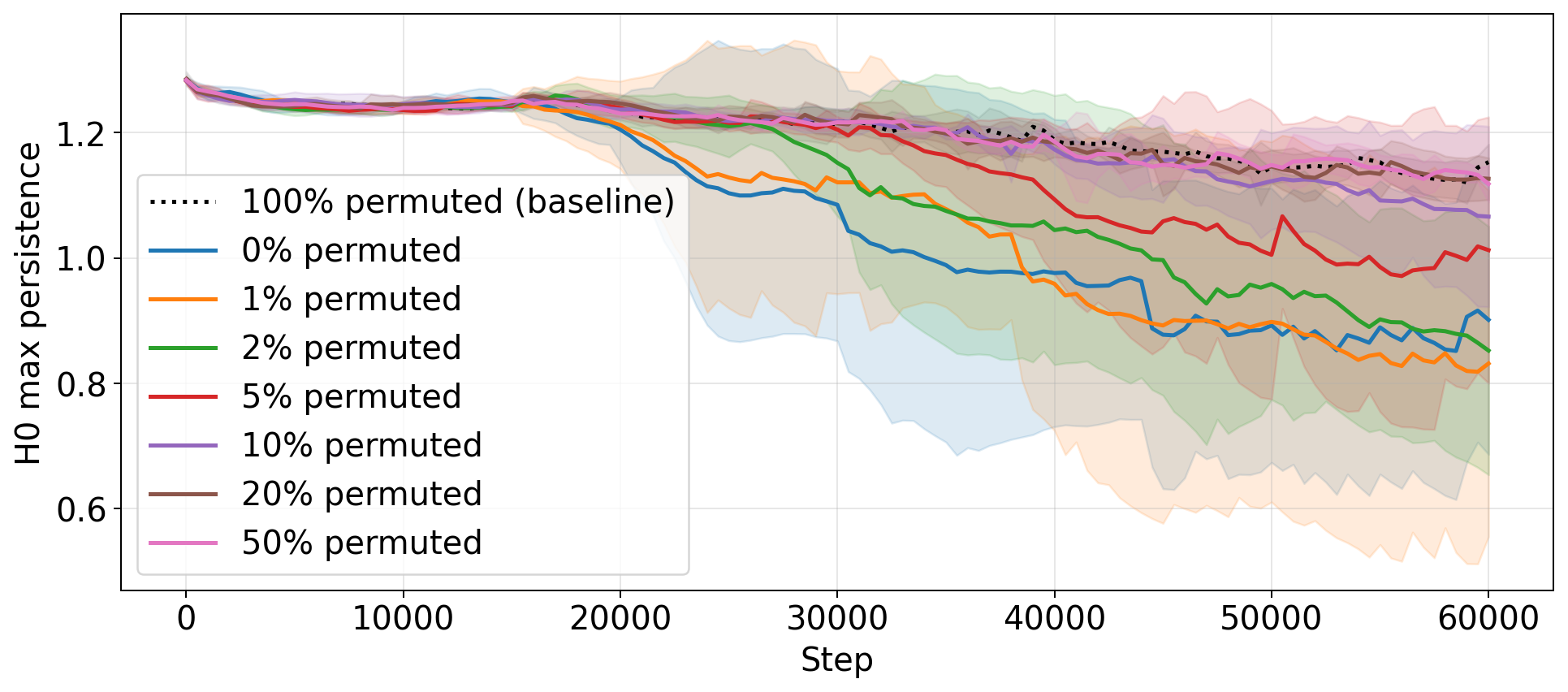}
    \end{subfigure}

    \vspace{0.5em}

    \begin{subfigure}[b]{0.49\textwidth}
        \centering
        \includegraphics[width=\linewidth]{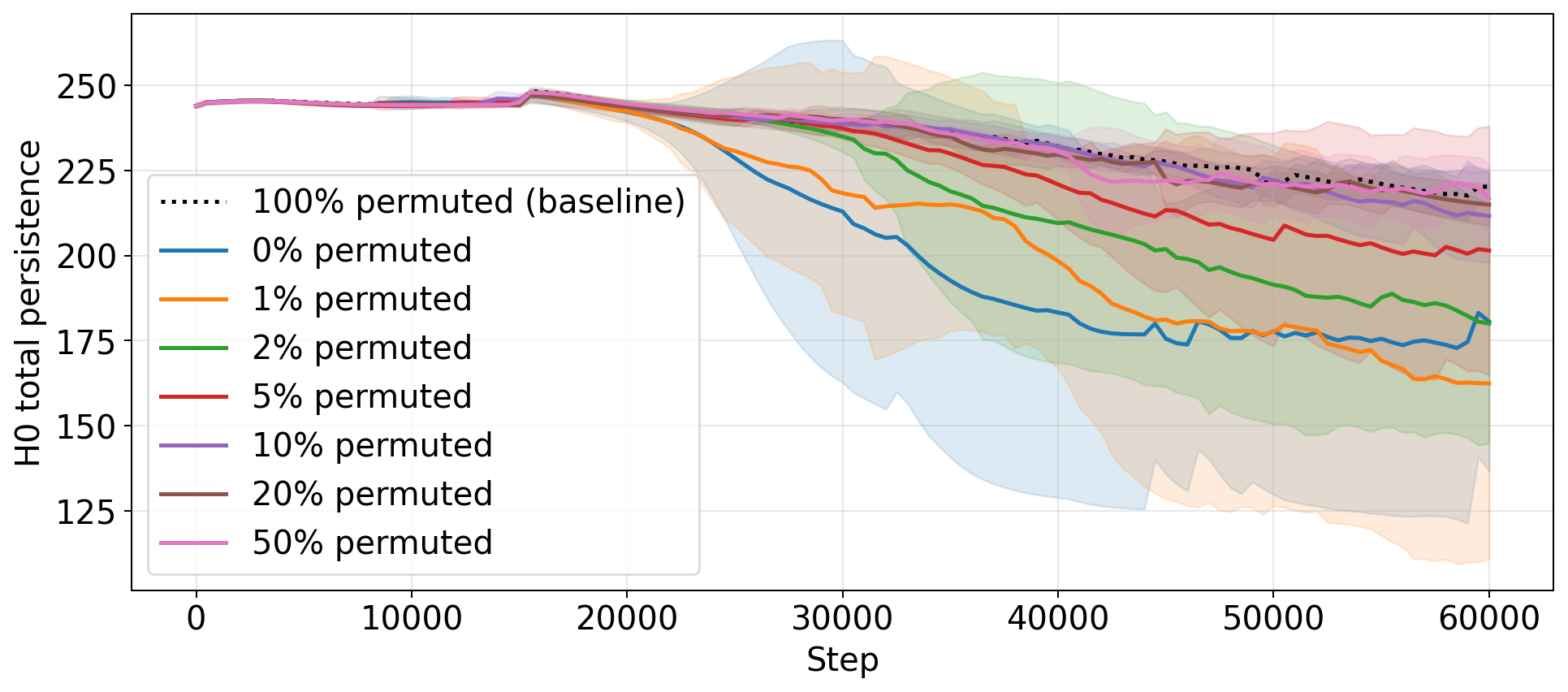}
    \end{subfigure}
    \hfill
    \begin{subfigure}[b]{0.49\textwidth}
        \centering
        \includegraphics[width=\linewidth]{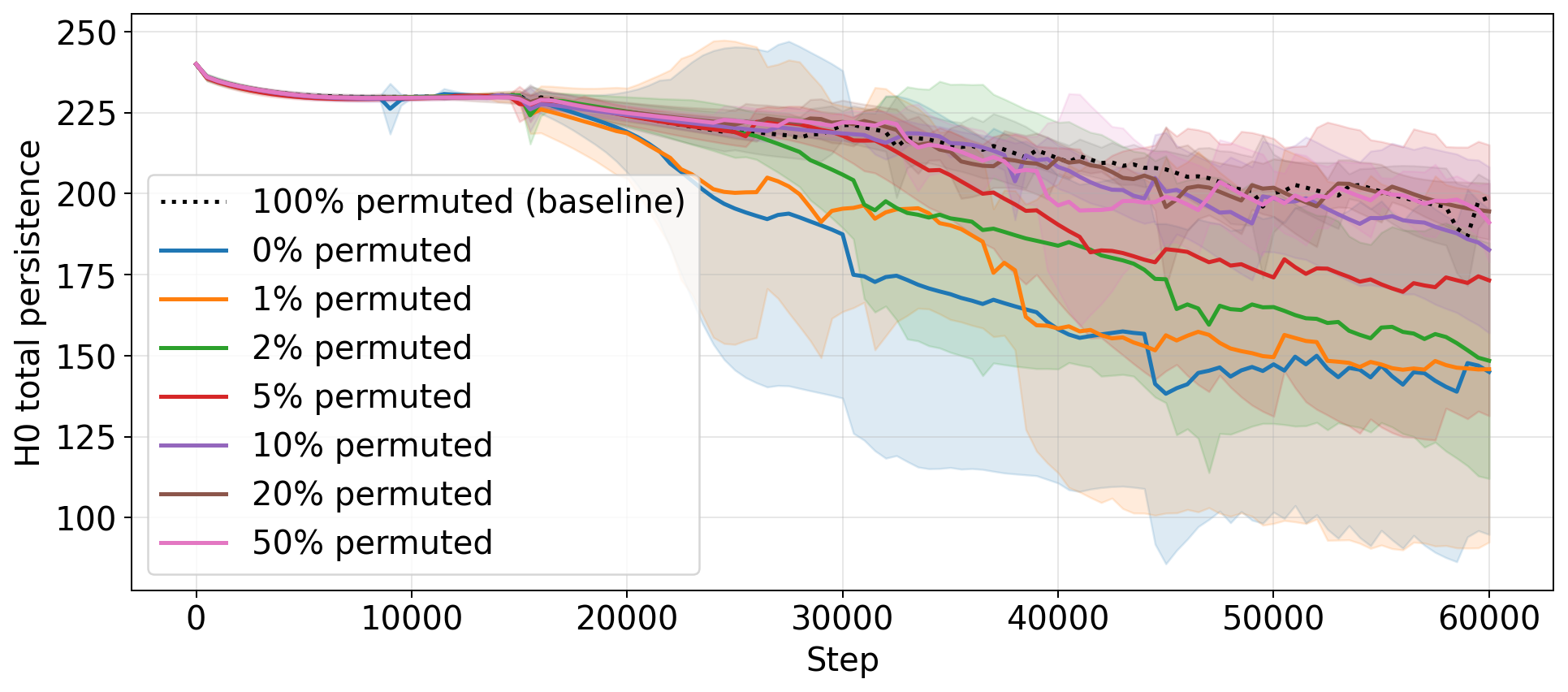}
    \end{subfigure}

    \vspace{0.5em}

    \begin{subfigure}[b]{0.49\textwidth}
        \centering
        \includegraphics[width=\linewidth]{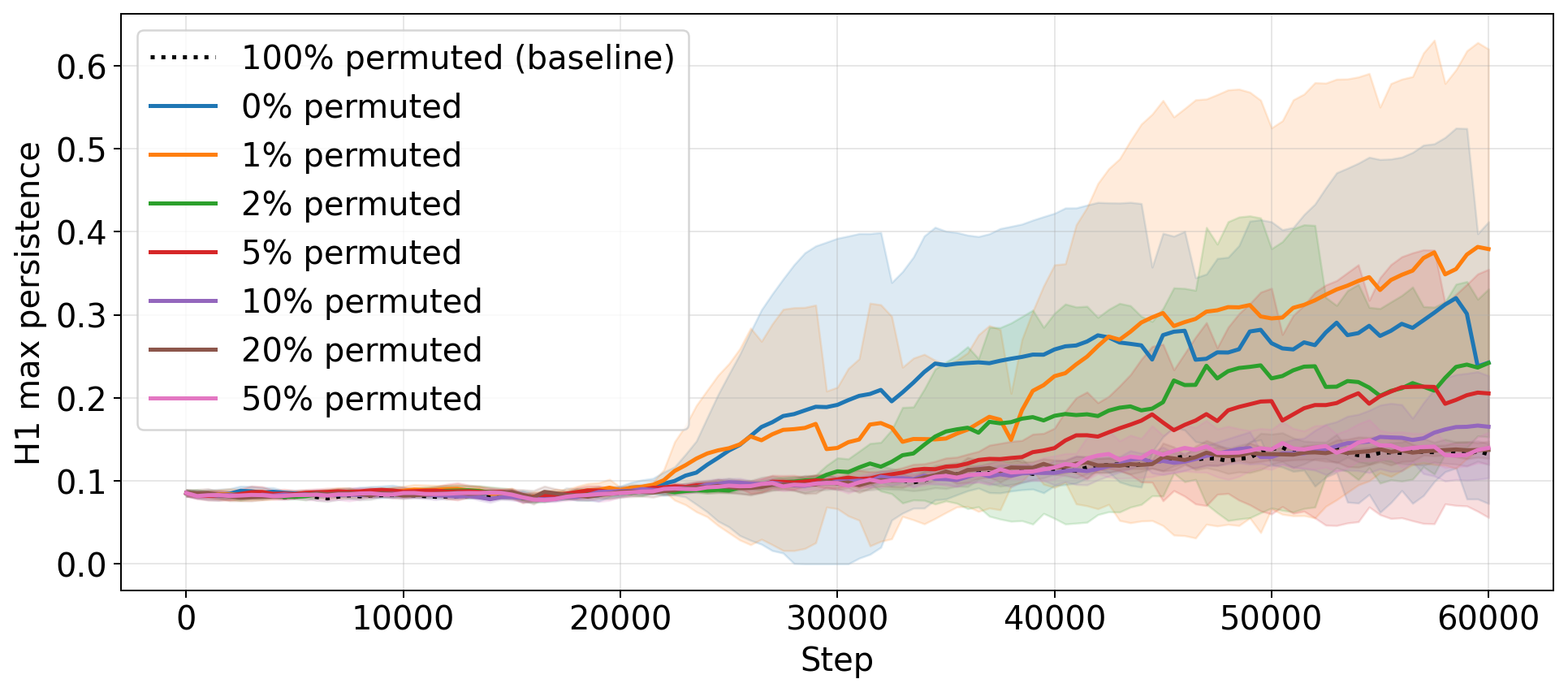}
    \end{subfigure}
    \hfill
    \begin{subfigure}[b]{0.49\textwidth}
        \centering
        \includegraphics[width=\linewidth]{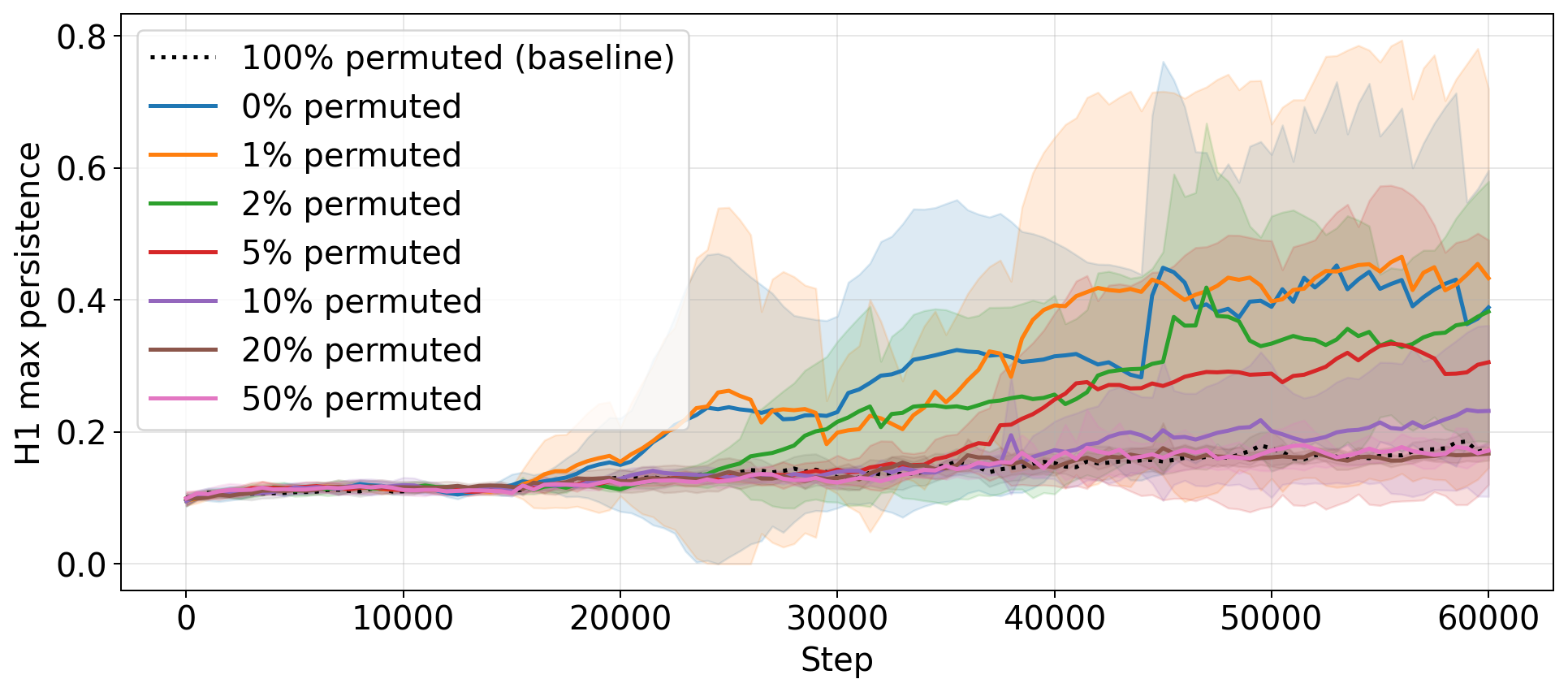}
    \end{subfigure}

    \caption{Training dynamics under label permutation for the MLP model. 
    The top panel shows test accuracy across training for different label permutation fractions $P_{\mathrm{frac}}$. 
    The remaining panels show the corresponding Vietoris--Rips PH statistics computed from the learned representations, with the left and right columns corresponding to the embedding layer (layer 0) and first hidden layer (layer 1), respectively. 
    Rows 2--4 report $H_0$ maximum persistence, $H_0$ total persistence, and $H_1$ maximum persistence. 
    }
\label{fig:mlp_ablation}
\end{figure}

\begin{figure}[htbp]
    \centering

    \begin{subfigure}[b]{0.48\linewidth}
        \centering
        \includegraphics[width=\linewidth]{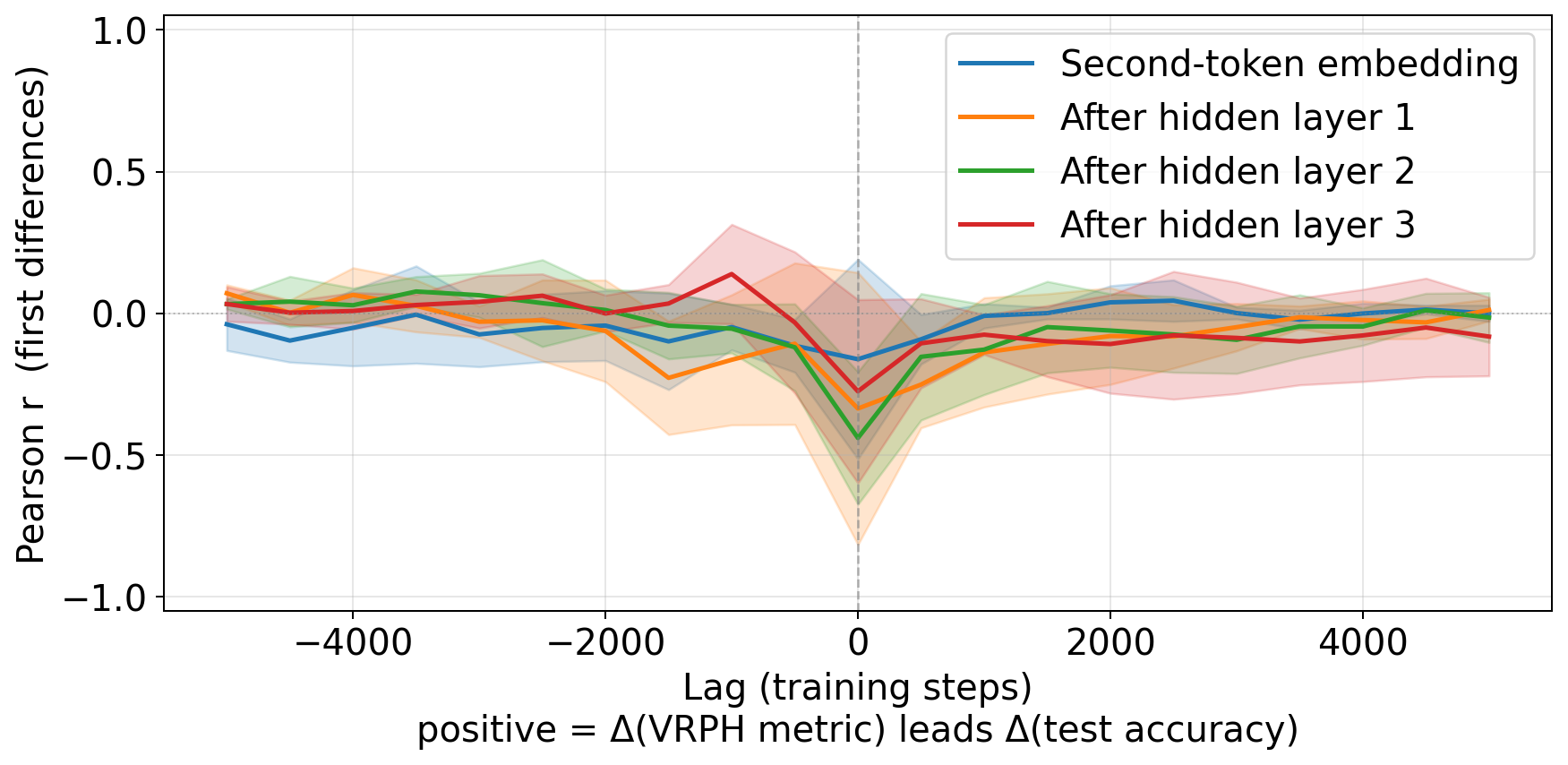}
        \caption{$H_0$ maximum persistence}
    \end{subfigure}
    \hfill
    \begin{subfigure}[b]{0.48\linewidth}
        \centering
        \includegraphics[width=\linewidth]{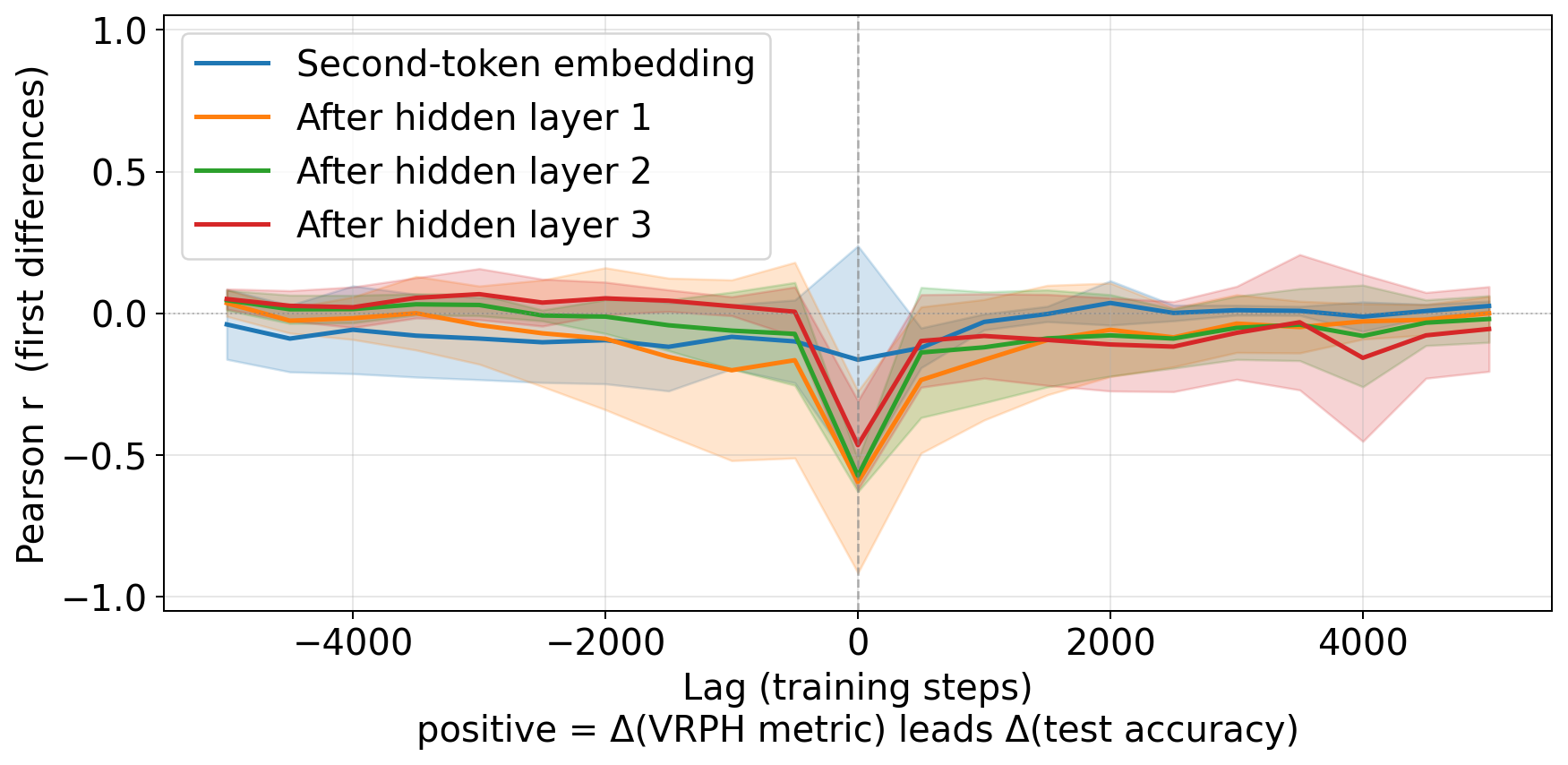}
        \caption{$H_0$ total persistence}
    \end{subfigure}

    \vspace{0.5em}

    \begin{subfigure}[b]{0.48\linewidth}
        \centering
        \includegraphics[width=\linewidth]{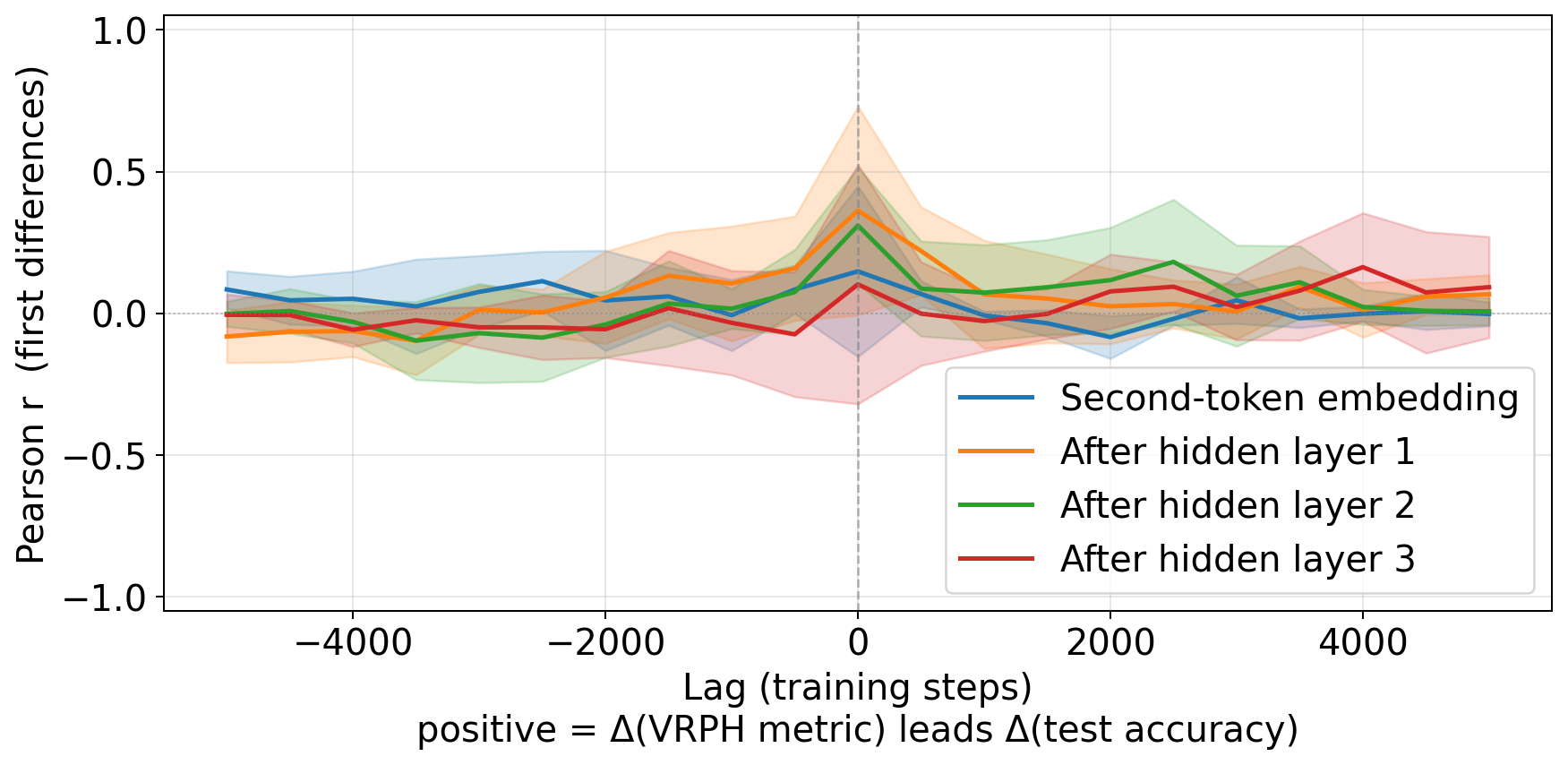}
        \caption{$H_1$ maximum persistence}
    \end{subfigure}
    \hfill
    \begin{subfigure}[b]{0.48\linewidth}
        \centering
        \includegraphics[width=\linewidth]{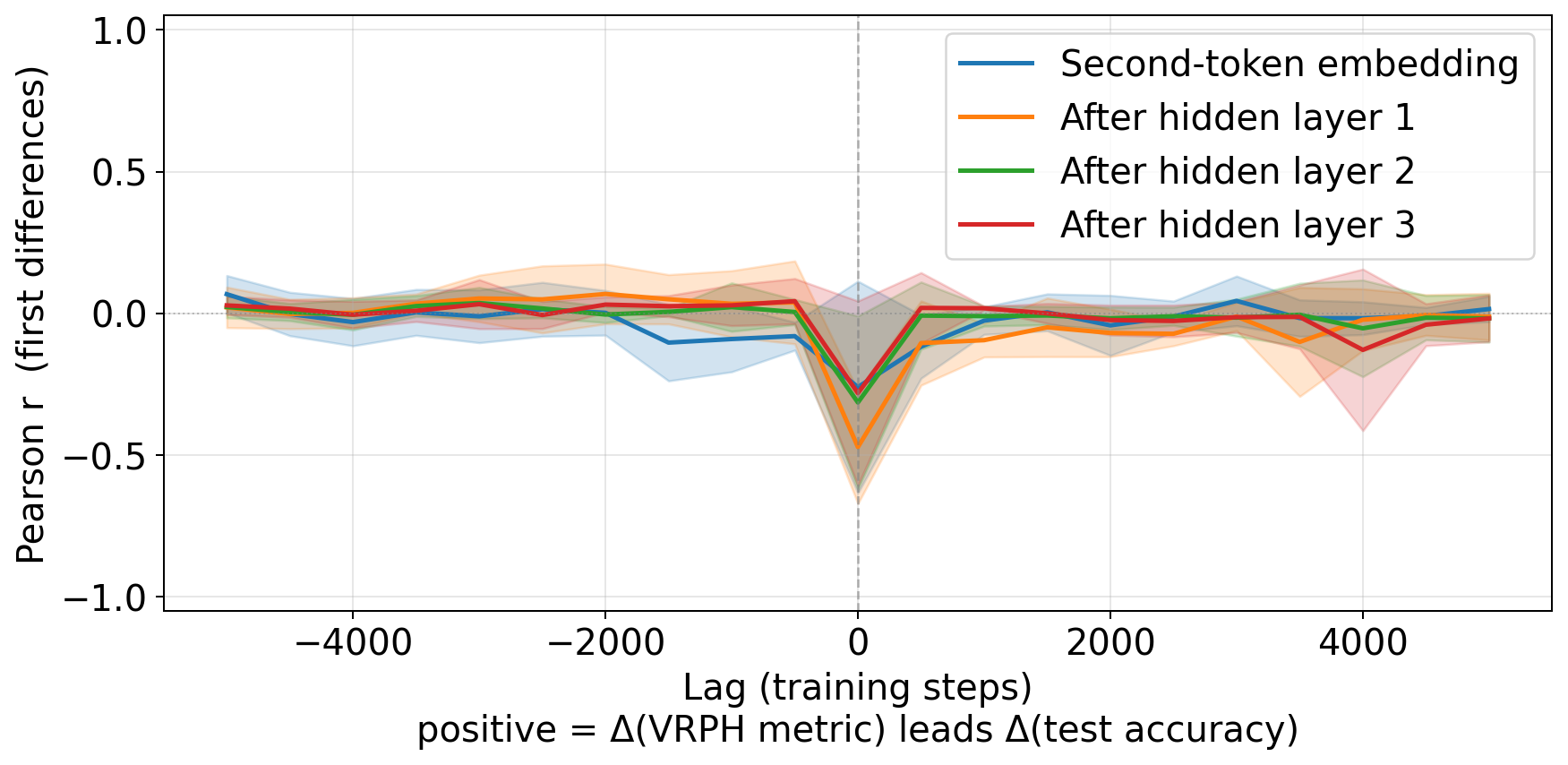}
        \caption{$H_1$ total persistence}
    \end{subfigure}

    \caption{
    Cross-correlation functions (CCFs) between the first-order differences of test accuracy and PH statistics for the MLP model at $P_{\mathrm{frac}}=0$. Across all four topological statistics, the peak correlation occurs near zero lag, indicating that changes in predictive performance and the corresponding topological restructuring of the learned representation space occur approximately simultaneously
    }
    \label{fig:ccf_mlp}
\end{figure}

In this section, we analyse the MLP model under the same label-permutation ablation regime. The corresponding PH–accuracy correlations across all layers and permutation levels are reported in Table~\ref{tab:ph_mlp_all_layers}, with representative training dynamics and CCF analyses shown in Figure~\ref{fig:mlp_ablation} and Figure~\ref{fig:ccf_mlp}.

In contrast to the Transformer model, the MLP exhibits a markedly more distributed structure of correlations across layers. Early layers (embedding and first hidden layer) show weaker and more variable associations with generalization, with large variance and several non-significant pattern under increased permutation. In contrast, deeper layers (Hidden 2 and Hidden 3) consistently exhibit strong and statistically significant correlations across all PH statistics in the low-noise regime ($P_{\mathrm{frac}} \leq 10\%$), with $H_0$ statistics reaching correlations as strong as $\rho=-0.86$ and $H_1$ maximum persistence reaching up to $\rho=0.79$.

As label corruption increases, these correlations weaken and collapse under complete permutation. This is consistent with the model being in an intermediate generalization regime, where PH-aligned cyclic structure is emerging but has not yet fully stabilized.

Overall, this suggests that, unlike the Transformer where PH signatures are strongly localized in early representations, the MLP encodes topological structure more deeply in the network hierarchy, with generalization-related geometric organization emerging predominantly in later hidden layers.

\begin{table}[ht]
\centering
\caption{Spearman Rank Correlation ($\rho \pm \text{SD}$) between PH measures and test accuracy across all MLP layers. Bold values indicate statistical significance ($p < 0.05$). Note that while early layers show inconsistent signals (high SD), deeper layers maintain a robust structural link to generalization even at 50\% noise.}
\label{tab:ph_mlp_all_layers}
\resizebox{\textwidth}{!}{
\begin{tabular}{llccccc}
\toprule
\textbf{Metric} & \textbf{Layer} & \textbf{0\%} & \textbf{10\%} & \textbf{20\%} & \textbf{50\%} & \textbf{100\%} \\ 
\midrule
$H_0$ Max   & Embed (L0) & \textbf{-0.49 $\pm$ 0.45} & -0.26 $\pm$ 0.37          & +0.03 $\pm$ 0.26          & -0.12 $\pm$ 0.14          & -0.08 $\pm$ 0.26 \\
            & Hidden 1   & \textbf{-0.55 $\pm$ 0.39} & -0.30 $\pm$ 0.38          & +0.05 $\pm$ 0.27          & -0.17 $\pm$ 0.15          & -0.11 $\pm$ 0.16 \\
            & Hidden 2   & \textbf{-0.69 $\pm$ 0.27} & \textbf{-0.51 $\pm$ 0.23} & \textbf{-0.39 $\pm$ 0.17} & -0.47 $\pm$ 0.25          & +0.01 $\pm$ 0.22 \\
            & Hidden 3   & \textbf{-0.81 $\pm$ 0.07} & \textbf{-0.69 $\pm$ 0.16} & \textbf{-0.57 $\pm$ 0.08} & \textbf{-0.51 $\pm$ 0.11} & +0.09 $\pm$ 0.16 \\ 
\addlinespace
$H_0$ Total & Embed (L0) & \textbf{-0.54 $\pm$ 0.40} & -0.30 $\pm$ 0.34          & -0.00 $\pm$ 0.21          & -0.16 $\pm$ 0.15          & -0.05 $\pm$ 0.27 \\
            & Hidden 1   & -0.61 $\pm$ 0.36          & -0.42 $\pm$ 0.30          & -0.15 $\pm$ 0.23          & -0.29 $\pm$ 0.18          & -0.04 $\pm$ 0.23 \\
            & Hidden 2   & \textbf{-0.84 $\pm$ 0.10} & \textbf{-0.77 $\pm$ 0.08} & \textbf{-0.64 $\pm$ 0.11} & \textbf{-0.58 $\pm$ 0.24} & +0.04 $\pm$ 0.22 \\
            & Hidden 3   & \textbf{-0.86 $\pm$ 0.06} & \textbf{-0.84 $\pm$ 0.08} & \textbf{-0.80 $\pm$ 0.05} & \textbf{-0.72 $\pm$ 0.13} & +0.11 $\pm$ 0.20 \\ 
\addlinespace
$H_1$ Max   & Embed (L0) & \textbf{+0.49 $\pm$ 0.41} & +0.30 $\pm$ 0.33          & +0.01 $\pm$ 0.20          & +0.18 $\pm$ 0.17          & +0.06 $\pm$ 0.27 \\
            & Hidden 1   & \textbf{+0.56 $\pm$ 0.39} & +0.41 $\pm$ 0.30          & +0.14 $\pm$ 0.21          & +0.24 $\pm$ 0.20          & +0.13 $\pm$ 0.12 \\
            & Hidden 2   & \textbf{+0.79 $\pm$ 0.09} & \textbf{+0.74 $\pm$ 0.06} & \textbf{+0.62 $\pm$ 0.05} & \textbf{+0.50 $\pm$ 0.25} & -0.01 $\pm$ 0.25 \\
            & Hidden 3   & \textbf{+0.76 $\pm$ 0.11} & \textbf{+0.73 $\pm$ 0.12} & \textbf{+0.62 $\pm$ 0.05} & \textbf{+0.56 $\pm$ 0.16} & -0.05 $\pm$ 0.23 \\ 
\addlinespace
$H_1$ Total & Embed (L0) & -0.36 $\pm$ 0.31          & -0.08 $\pm$ 0.27          & \textbf{-0.30 $\pm$ 0.09} & -0.09 $\pm$ 0.20          & +0.11 $\pm$ 0.11 \\
            & Hidden 1   & \textbf{-0.68 $\pm$ 0.16} & \textbf{-0.62 $\pm$ 0.18} & \textbf{-0.58 $\pm$ 0.10} & \textbf{-0.38 $\pm$ 0.27} & +0.06 $\pm$ 0.27 \\
            & Hidden 2   & \textbf{-0.69 $\pm$ 0.18} & \textbf{-0.67 $\pm$ 0.11} & \textbf{-0.58 $\pm$ 0.12} & \textbf{-0.53 $\pm$ 0.23} & +0.07 $\pm$ 0.35 \\
            & Hidden 3   & \textbf{-0.72 $\pm$ 0.16} & \textbf{-0.73 $\pm$ 0.14} & \textbf{-0.65 $\pm$ 0.06} & \textbf{-0.63 $\pm$ 0.19} & -0.01 $\pm$ 0.33 \\ 
\bottomrule
\end{tabular}
}
\end{table}

In contrast to the Transformer model, where changes in predictive performance consistently preceded the associated topological transition, the MLP exhibits peak cross-correlation near zero lag across all PH statistics (Figure~\ref{fig:ccf_mlp}). This behavior is consistent across the other permutation levels exhibiting generalization. Overall, this suggests that, for the MLP, improvements in generalization and the corresponding geometric reorganization of the latent space occur approximately simultaneously.

\end{document}